\documentclass[final,5p,times]{elsarticle}
\usepackage{lipsum}
\usepackage{adjustbox}
\usepackage{makecell}
\usepackage{amssymb}
\usepackage{amsmath}
\usepackage{lineno}
\usepackage{array}
\usepackage{float}
\usepackage[linesnumbered,ruled,vlined]{algorithm2e}
\usepackage{algorithmic}
\usepackage{subcaption}
\usepackage{tablefootnote}
\usepackage{url}
\usepackage[colorlinks = true,
   linkcolor = blue,
   urlcolor = blue,
   citecolor = blue,
   anchorcolor = blue]{hyperref}

\usepackage{etoolbox}
\usepackage{pdflscape}
\usepackage{pifont}
\usepackage{tcolorbox}
\usepackage{tikz}
\usepackage{booktabs}
\usepackage{colortbl}
\usepackage{enumitem}
\usepackage{smartdiagram}
\usepackage{svg}

\usepackage{ifthen}
\definecolor{arrowcolor}{RGB}{5,87,119}
\definecolor{textcolor}{RGB}{3,148,160}
\def\NTsize{\sffamily\large}
\def\CTsize{\sf\small}
\def\body#1#2#3#4#5#6#7#8#9{
        \begin{scope}[shift={(#1)},rotate=0,transform shape]
            \ifthenelse{\equal{#2}{end}}{
                \ifthenelse{\equal{#3}{reverse}}{
                    \filldraw[#6]
                        (-1pt,#5*0.5+#7/2)
                            -- ++(#4*0.5+1pt,0)
                            -- ++ (0,3mm) 
                            -- (#4,#5*0.5)
                            -- (#4*0.5,#5*0.5-#7/2-3mm)
                            -- ++(0,3mm)
                            -- ++(-#4*0.5-1pt,0);
                    }{
                    \filldraw[#6]
                        (#4+1pt,#5*0.5+#7/2)
                            -- ++(-#4*0.5-1pt,0)
                            -- ++ (0,3mm)
                            -- (0,#5*0.5)
                            -- (#4*0.5,#5*0.5-#7/2-3mm)
                            -- ++(0,3mm)
                            -- ++(#4*0.5+1pt,0);
                }
            }{
                \filldraw[#6] (0,#5*0.5+#7/2) rectangle ++(#4,-#7);
                \draw[textcolor](#4*0.5,#5*0.5) node[circle,draw,fill=white, minimum size=#7-3mm,font=\NTsize](temp){#2};
                \draw(temp)+(#8:#9) node[text width=#4*2, align=center,font=\CTsize](t2){#3};
                \draw[#6!40!white, preaction={draw,#6, line width=2pt}](temp)--(t2);
            }               
        \end{scope} 
}

\def\TAIL#1#2#3#4#5{
    \begin{scope}[shift={(#1)}]
        \filldraw[#4] (#2+1pt,#3*0.5+#5/2)-- ++(-#2*0.5-1pt,0) -- (#2*0.75,#3*0.5) -- (#2*0.5,#3*0.5-#5/2) -- ++ (1pt+#2*0.5,0);
    \end{scope} 
}

\def\CURVE#1#2#3#4#5#6{
    \begin{scope}[shift={(#1)},#4,transform shape]
        \filldraw[#5](-1pt,#3*0.5-#6/2) -- ++(1pt,0) arc (90:-90:#3*0.5-#6/2) -- ++(-1pt,0)-- ++(0,-#6) -- ++(1pt,0) arc (-90:90:#3/2+#6/2)--++(-1pt,0);
    \end{scope} 
}

\def\SNAKETEXT#1(#2)[#3][#4][#5]#6#7{
    \begin{scope}[shift={(#2)}]
        \edef\Shiftx{0}
        \edef\Shifty{0}
        \pgfmathparse{int(#6*2)}
        \xdef\tpl{\pgfmathresult}
        \foreach \cod/\direc/\dist/\desc [count=\ctr from 0] in {#1}{
            \ifnum\ctr=0
            \TAIL{\Shiftx*#3,-\Shifty*#4-#4}{#3}{#4}{#5}{#7}
            \fi
            \ifnum\ctr<#6
            \pgfmathparse{int(\Shiftx+1)}
            \xdef\Shiftx{\pgfmathresult}
            \body{\Shiftx*#3,-\Shifty*#4-#4}{\cod}{\desc}{#3}{#4}{#5}{#7}{\direc}{\dist}
            \fi
            \ifnum\ctr=#6
            \pgfmathparse{int(\Shiftx+1)}
            \xdef\Shiftx{\pgfmathresult}
            \CURVE{\Shiftx*#3,-\Shifty*#4-#4}{#3}{#4}{}{#5}{#7}
            \pgfmathparse{int(\Shiftx-1)}
            \xdef\Shiftx{\pgfmathresult}
            \pgfmathparse{int(\Shifty+1)}
            \xdef\Shifty{\pgfmathresult}
            \body{\Shiftx*#3,-\Shifty*#4-#4}{\cod}{\desc}{#3}{#4}{#5}{#7}{\direc}{\dist}
            \fi
            \ifnum\ctr>#6
            \ifnum\ctr<\tpl
            \pgfmathparse{int(\Shiftx-1)}
            \xdef\Shiftx{\pgfmathresult} 
            \body{\Shiftx*#3,-\Shifty*#4-#4}{\cod}{\desc}{#3}{#4}{#5}{#7}{\direc}{\dist}
            \fi
            \fi
            \ifnum\ctr=\tpl
            \pgfmathparse{int(\Shiftx-1)}
            \xdef\Shiftx{\pgfmathresult}
            \CURVE{\Shiftx*#3+#3,-\Shifty*#4-#4}{#3}{#4}{xscale=-1}{#5}{#7}
            \pgfmathparse{int(\Shifty+1)}
            \xdef\Shifty{\pgfmathresult}
            \pgfmathparse{int(\Shiftx+1)}
            \xdef\Shiftx{\pgfmathresult}
            \body{\Shiftx*#3,-\Shifty*#4-#4}{\cod}{\desc}{#3}{#4}{#5}{#7}{\direc}{\dist}
            \xdef\ctr{0}
            \fi
        }
    \end{scope} 
}

\usepackage{tocloft}

\usetikzlibrary{arrows.meta,shapes,positioning,shadows,trees}
\usepackage[edges]{forest}

\renewcommand{\arraystretch}{1.1} 

\newcommand{\squaremarker}[1]{%
    \tikz[baseline=-0.75ex]\node[fill=#1, rectangle, minimum width=2.4mm, minimum height=2.4mm] {};%
}

\begin{document}

\begin{frontmatter}

\title{Categorical data clustering: 25 years beyond K-modes}

\author[a]{Tai Dinh\corref{cor1}}
\author[a]{Wong Hauchi}
\author[b]{Philippe Fournier-Viger}
\author[c]{Daniil Lisik\corref{cor1}}
\author[d]{Minh-Quyet Ha}
\author[d]{Hieu-Chi Dam}
\author[d]{Van-Nam Huynh}

\cortext[cor1]{Corresponding author: Tai Dinh (t\_dinh@kcg.edu), Daniil Lisik (daniil.lisik@gu.se)}

\affiliation[a]{organization={The Kyoto College of Graduate Studies for Informatics},
            addressline={7 Tanaka Monzencho, Sakyo Ward}, 
            city={Kyoto City},
            state={Kyoto},
            country={Japan}}
\affiliation[b]{organization={College of Computer Science and Software Engineering, Shenzhen University},
            addressline={Shenzhen City, Guangdong},
            country={China}}
\affiliation[c]{organization={University of Gothenburg},
            addressline={Medicinaregatan 1F, 413 90}, 
            city={Göteborg},
            country={Sweden}}
\affiliation[d]{organization={Japan Advanced Institute of Science and Technology},
            addressline={1-1 Asahidai, Nomi, Ishikawa},
            country={Japan}}

\begin{abstract}
The clustering of categorical data is a common and important task in computer science, offering profound implications across a spectrum of applications. Unlike purely numerical data, categorical data often lack inherent ordering as in nominal data, or have varying levels of order as in ordinal data, thus requiring specialized methodologies for efficient organization and analysis. This review provides a comprehensive synthesis of categorical data clustering in the past twenty-five years, starting from the introduction of \textsc{K-modes}. It elucidates the pivotal role of categorical data clustering in diverse fields such as health sciences, natural sciences, social sciences, education, engineering, and economics. Practical comparisons are conducted for algorithms 
having public implementations, highlighting distinguishing clustering methodologies and revealing the performance of recent algorithms on several benchmark categorical datasets. Finally, challenges and opportunities in the field are discussed.
\end{abstract}

 
\end{frontmatter}

\tableofcontents


\section{Introduction} \label{sec:introduction}
The ever-increasing volume and diversity of data necessitate continual advancements in methodologies for comprehensive understanding, processing, and segmentation. Data analysis techniques are broadly categorized into two main types: exploratory and confirmatory \citep{tukey1977exploratory}. Exploratory techniques discern the general characteristics or structures of high-dimensional data without pre-specified models or hypotheses, while confirmatory approaches seek to validate existing hypotheses or models given available data. 

Among these techniques, clustering is a fundamental component of exploratory data analysis, used to segment datasets into groups based on similarity or dissimilarity metrics \citep{aggarwal2013introduction}. Techniques such as partitional and hierarchical clustering enable researchers to uncover latent patterns, structures, and relationships within data, providing valuable insights into data organization. Furthermore, clustering facilitates classification tasks, such as segmenting customer groups based on purchasing behavior, contributing to a comprehensive understanding of the data. Additionally, clustering can be used for data compression, reducing large datasets into representative cluster prototypes in a smaller subspace, thus aiding in the visualization and interpretation of complex data structures.

Numerical data types, as the name suggests, represent numerical attributes, such as age or salary. A distinction is sometimes made between continuous (measurable) and discrete (with a clear space between values; countable) data, or by whether there is a true meaningful zero (ratio data) or not (interval data). Numerical data are distinct from categorical data, which encapsulates information into categories that lack the aforementioned calculable distance property of numerical data \citep{han2022data}. In data clustering, the majority of algorithms are designed to exclusively handle numerical data types. However, real-world datasets often consist of categorical data or incorporate both continuous and categorical features (variables), known as mixed data. This underscores the importance of addressing categorical data in clustering methodologies.

\begin{figure*}[!htb]
\vspace{-2.5cm}
  \centering
  \includegraphics[width=\linewidth]{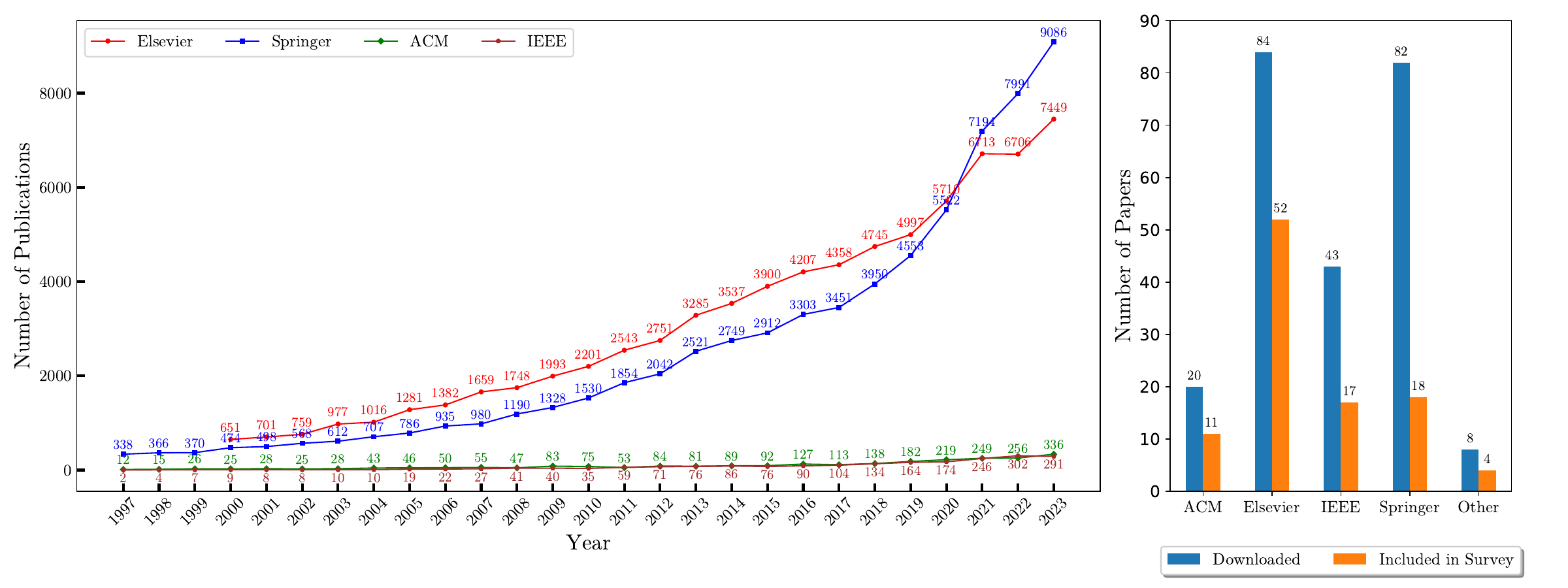}
  \caption{(a) Publications by major publishers over the past 25 years. (b) Papers on categorical data clustering algorithms included in the present survey}
  \label{fig:pub_stats}
\end{figure*}

The evolution of clustering methodology for categorical data represents a comprehensive interdisciplinary effort. Computer scientists, social scientists, psychologists, biologists, statisticians, mathematicians, engineers, medical researchers, and various professionals engaged in data acquisition and analysis have collectively advanced the field of clustering methodology. Figure \ref{fig:pub_stats} (a) illustrates the publication trends associated with the keywords [``categorical data'' OR ``mixed data''] across four publishers: ACM, Elsevier, IEEE, and Springer. The data reveals a consistent upward trajectory in the number of publications concerning these topics across all publishers; particularly noteworthy are the pronounced trends observed in Elsevier and Springer. This suggests a notable surge in research interest and attention towards categorical and mixed data among researchers.

Clustering categorical data presents significant challenges due to the lack of well-established distance metrics for features. Categorical features can be either ordinal, with a natural order, or nominal, with no inherent order (including binary data often considered nominal) \citep{agresti2012categorical}. This variety requires careful consideration of how to assess relationships and distances in such data. \cite{huang1997fast,huang1998extensions} extended the \textsc{K-means} algorithm to handle categorical data by introducing \textsc{K-modes}. Since then, numerous clustering algorithms and methods for categorical data have emerged, making it difficult for researchers to gain an overview or select the most suitable algorithm for their needs. This motivates our comprehensive survey to offer a detailed overview of the field, catering to readers from various disciplines. In general, the major contributions of this paper are as follows: 
\begin{itemize}
    \item [-] We conduct a comprehensive review of the literature addressing categorical data clustering, spanning over the past 25 years (1997-2024), focusing on clustering algorithms designed for categorical data. We provide the background and characteristics of different clustering algorithms. We describe the historical developments of algorithms proposed in this period. We also divide algorithms into three different periods based on the publication years and briefly describe how these algorithms work. The present review aims to enable readers to easily capture the main ideas of each algorithm and attain a general understanding of the evolution of algorithms proposed by the research community thus far. 
    \item [-] We present a taxonomy for clustering algorithms, categorized by clustering type. This taxonomy highlights the key features and characteristics of each algorithm, providing concise overviews of their underlying theoretical frameworks. Additionally, it includes essential information such as parameter settings, datasets, evaluation metrics, and visualization techniques used in the experiments for each algorithm. Furthermore, we explore the use of different types of datasets, visualization methods, and evaluation metrics for categorical data clustering.
    \item [-] We examine how categorical data clustering is applied across various fields and contexts, offering readers a broad overview of its adaptability and utilization.
    \item [-] To further investigate the efficiency of categorical data clustering techniques, we collect available Python source code from various sources, such as GitHub and Python libraries. We evaluate their clustering results on four commonly used categorical datasets using several external validation metrics. The source code and datasets are organized in a notebook for reproducibility.
    \item [-] We investigate major challenges, trends, and opportunities, as well as outline future research directions for categorical data clustering.
\end{itemize}
 
The rest of this paper is structured as follows: Section \ref{sec:datasource} describes the methodology for gathering and curating relevant literature. Section \ref{sec:categorical_data_clustering}  reviews the historical evolution and foundational concepts of clustering algorithms for categorical data. Section \ref{sec:taxonomy} presents a taxonomy categorizing algorithms by various criteria. Section \ref{sec:categorical_clustering_in_other_fields} discusses applications of categorical data clustering in different fields. Section \ref{sec:comparative_experiment} presents a comparative analysis of several algorithms using Python and a performance tracking of algorithms. Section \ref{sec:major_challenges} addresses key challenges in this domain. Section \ref{sec:trends_opportunities} examines trends and future research opportunities. Finally, section \ref{sec:conclusion} summarizes the survey and highlights the key insights.
\section{Data sources of the review} \label{sec:datasource}

We reviewed publications on categorical clustering from four major publishers: ACM, Elsevier, IEEE, and Springer. Specifically, we collected papers from these publishers spanning the period from 1997 to 2024, by searching with keywords [``categorical data clustering" OR ``clustering of categorical data" OR ``categorical clustering"]. Additionally, we incorporated some publications from other publishers found within the first 15 pages of Google Scholar search results using the same keywords. 

The blue columns in Figure \ref{fig:pub_stats} (b) indicate the number of publications dedicated to designing clustering algorithms for categorical data. To refine our selection of significant papers for this survey, we applied the following criteria:
\begin{itemize}
\item [-] Papers from 1997-2007 with at least $100$ citations.
\item [-] Papers from 2008-2015 with at least $50$ citations.
\item [-] Papers from 2016-2019 with at least $20$ citations.
\item [-] Papers from 2020-2021 with at least $10$ citations.
\item [-] Papers from 2022-2023 included regardless of citations.
\end{itemize}
The orange columns in Figure \ref{fig:pub_stats} (b) show the number of publications that meet all specified criteria. In total, 102 peer-reviewed papers from journals and conference proceedings were included in the survey. These papers introduce various methodologies, present novel algorithms and computational techniques, and focus on different aspects of categorical data clustering.

\begin{figure}[!htb]
  \centering
  \includegraphics[width=\linewidth]{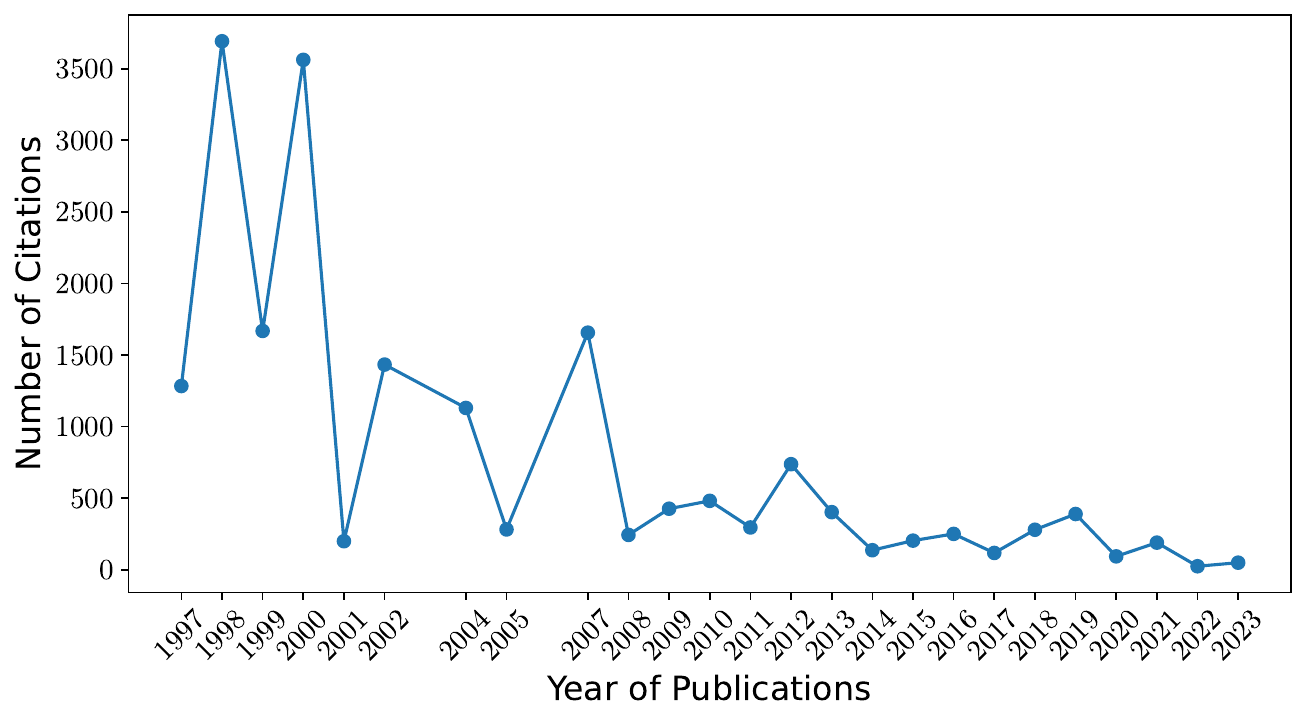}
  \caption{Total citations received by selected papers from their respective publication years up to April 2024}
  \label{fig:pub_citations}
\end{figure}

Figure \ref{fig:pub_citations} illustrates the cumulative total number of citations for the 102 included papers, counted from their publication years up to April 28, 2024. The total number of citations of these papers is 19,612. It can be observed that these papers, most notably the very early works, have garnered significant attention from the research community.

Unlike previous surveys \citep{naouali2020clustering,cendana2024categorical} on categorical data clustering, our review extends to the application of these techniques across various fields and contexts. We examined and compiled approximately 64 papers from diverse domains, published up to 2024, that employ categorical data clustering. Additionally, we reviewed 14 more papers to identify challenges and emerging trends in the field. In total, our survey encompasses around 180 papers from the literature.

\section{Categorical data clustering} \label{sec:categorical_data_clustering}
\subsection{Background}
\begin{table}[!htb]
    \centering
    \caption{Weights of 1996 US Olympic Rowing Team\tablefootnote{\href{http://jse.amstat.org/jse_data_archive.htm}{Source: Data Archive, Journal of Statistics Education}}}
    \begin{adjustbox}{max width=\linewidth}
    \begin{tabular}{|l|l|r||l|l|r|}
      \hline
      \textbf{Name} & \textbf{Event} & \textbf{Weight} & \textbf{Name} & \textbf{Event} & \textbf{Weight} \\
      \hline
      Auth & LW\_double\_sculls & 154 & Klepacki & four & $>200$ \\
      Beasley & single\_sculls & $>200$ & Koven & eight & 175-200 \\
      Brown & eight & $>200$ & Mueller & quad & $>200$ \\
      Burden & eight & 175-200 & Murphy & eight & $>200$ \\
      Carlucci & LW\_four & $<150$ & Murray & four & $>200$ \\
      Collins,D & LW\_four & $<150$ & Peterson,M & pair & $>200$ \\
      Collins,P & eight & 175-200 & Peterson,S & LW\_double\_sculls & 150-175 \\
      Gailes & quad & $>200$ & Pfaendtner & LW\_four & $<150$ \\
      Hall & four & 175-200 & Schnieder & LW\_four & $<150$ \\
      Holland & pair & 175-200 & Scott & four & $>200$ \\
      Honebein & eight & 175-200 & Segaloff & coxswain & $<150$ \\
      Jamieson & quad & $>200$ & Smith & eight & $>200$ \\
      Kaehler & eight & $>200$ & Young & quad & $>200$ \\
      \hline
    \end{tabular}
    \end{adjustbox}
    \label{tab:rowing}
\end{table}
\begin{figure*}[!htb]
\vspace{-2cm}
  \centering
  \includegraphics[width=\linewidth]{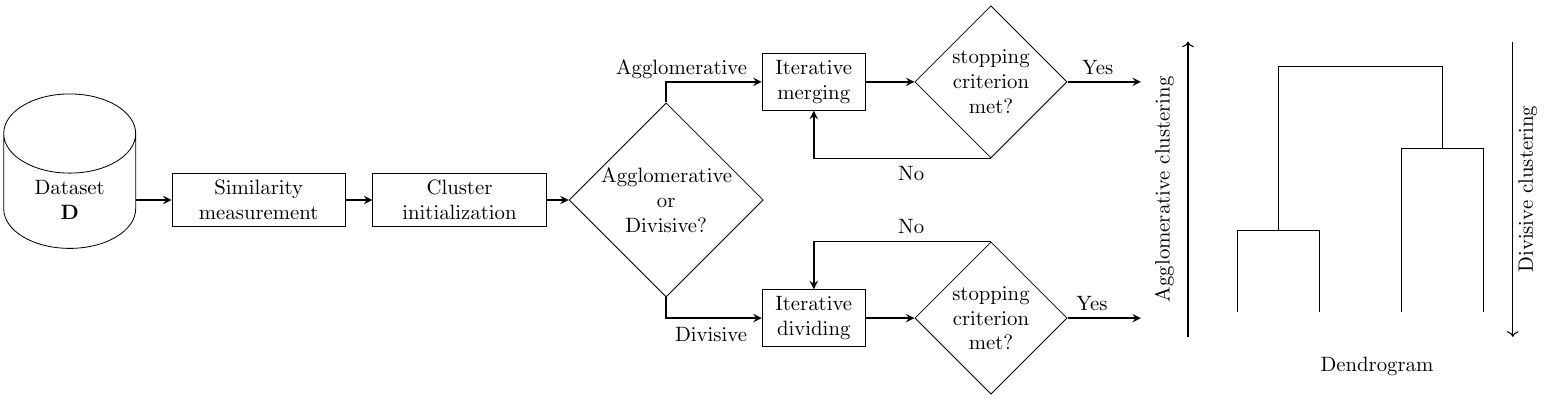}
  \caption{\textsc{Hierarchical Clustering}}
  \label{fig:hierarchical_clustering}
\end{figure*}
Let $\mathbf{A}$ = $\{\mathbf{A}_1, \mathbf{A}_2, \dots, \mathbf{A}_m\}$ denote a set of $m$ distinct categorical attributes (also referred to as features or variables), each associated with a finite set $\mathcal{O}_j$ ($1 \leq j \leq m$) as its domain, where DOM$(\mathbf{A}_j) = \mathcal{O}_j$ ($\geq2$ discrete values). A categorical dataset $\mathbf{D}$ = $\{\mathbf{x}_1, \mathbf{x}_2, \dots, \mathbf{x}_n\}$ comprises $n$ categorical data points (also referred to as data objects, data instances, or observations), where each object $\mathbf{x}_i \in \mathbf{D}$ ($1 \leq i \leq n$) is a tuple $\mathbf{x}_i = (\mathbf{x}_{i1}, \mathbf{x}_{i2}, \dots, \mathbf{x}_{im}) \in \mathcal{O}_1 \times \mathcal{O}_2 \times \dots \times \mathcal{O}_m$. In other words, $\mathbf{D}$ can be represented by a data matrix with $n$ rows and $m$ columns ($n\gg m$), and each element at position $(i, j)$ $(1 \leq i \leq n, 1 \leq j \leq m)$ of the matrix contains the value of the object $\mathbf{x}_i$ for the $j^{th}$ attribute, such that $\mathbf{x}_{ij} \in \mathcal{O}_j$. A categorical dataset can also be considered a transaction dataset, where each instance is a transaction and each category is an item bought by a customer.
As an example, Table \ref{tab:rowing} shows a publicly available categorical dataset named ``Weights of 1996 US Olympic Rowing Team''. In this dataset, the columns describe, from left to right, the rower's: name, event, and weight. There are eight different event categories. Similarly, the weight values are divided into four groups: under 150 lbs, 150-175 lbs, 175-200 lbs, and over 200 lbs, and thus also presented in categorical form.
\subsubsection{\textsc{Similarity} and \textsc{Dissimilarity} measures}
In the context of clustering, the \emph{similarity} measures are used to assess similarities between two objects, whereas \emph{dissimilarity} measures are used to quantify the differences (or \emph{distances}) between two objects or between objects and cluster centers. The relationship between \emph{similarity} and \emph{dissimilarity} can be defined in various ways \citep{deza2009encyclopedia}, with one common approach treating them as complementary measures \citep{han2022data}. Specifically, if the \emph{similarity} and \emph{dissimilarity} of two objects $x_i$ and $x_{i'}$ are denoted as $\text{sim}(x_i,x_{i'})$ and $\text{dis}(x_i,x_{i'})$, respectively, then the \emph{dissimilarity} can be calculated as $\text{dis}(x_i,x_{i'}) = 1 - \text{sim}(x_i,x_{i'})$. 

Numerous \emph{(dis)similarity} measures have been proposed and applied to categorical data clustering \citep{sokal1958statistical,gambaryan1964mathematical,huang1998extensions, san2004alternative, boriah2008similarity, nguyen2023method}. In the literature, the \textsc{Simple Matching Similarity} measure \citep{sokal1958statistical} has become the most
commonly used similarity measure for categorical data. Its popularity likely stems from its simplicity and ease of use. Given two categorical values occurring in two objects $x_1$ and $x_{2}$ at the $j^{th}$ attribute ($\mathbf{A}_j$). The \emph{similarity} between them is defined as:
\begin{equation}
\textnormal{sim}(x_{1j},x_{2j}) = \begin{cases}
 1  & \text{ if } x_{1j}= x_{2j} \\
 0 & ~\textnormal{otherwise}
\end{cases}
\label{eq:simple_matching_sim}
\end{equation}
Equivalently, the \emph{dissimilarity} between them can be defined as:
\begin{equation}
\delta(x_{1j},x_{2j}) = \begin{cases}
 1  & \text{ if } x_{1j} \neq x_{2j} \\
 0 & ~\textnormal{otherwise}
\end{cases}
\label{eq:simple_matching_dis}
\end{equation}
The \emph{dissimilarity} (or \textsc{Hamming Distance}) between two objects $x_1$ and $x_2$ consisting of $m$ attributes can be measured as:

\begin{align}
\textnormal{dis}(x_1, x_2) &= \sum_{j=1}^m \big(1 - \textnormal{sim}(x_{1j}, x_{2j})\big) \\
&= \sum_{j=1}^m \delta(x_{1j}, x_{2j})
\label{eq:distance_simple_matching}
\end{align}

Numerous approaches exist for defining \emph{similarity} measures for categorical data, as discussed in detail by \cite{boriah2008similarity}. Measures such as \textsc{Goodall-1, Goodall-2, Goodall-3, Goodall-4 Similarity} measures \citep{boriah2008similarity} and the \textsc{Gambaryan Similarity} \citep{gambaryan1964mathematical} assign a value of $0$ for mismatched attribute values, while defining different ways for matched values. For example, the \textsc{Goodall-4 Similarity} is computed as follows:
\begin{equation}
\textnormal{sim}(x_{1j},x_{2j}) =\begin{cases}
    1 - P_j^2(x_{1j}) & \text{if } x_{1j} = x_{2j} \\
    0 & \text{otherwise}
  \end{cases}
\label{eq:goodall-1}
\end{equation}
where
\begin{equation}
P_j^2(q) = \frac{f_j(q)(f_j(q) - 1)}{n(n - 1)}
\end{equation}
$f_j(q)$ is the number of times attribute $\mathbf{A}_j$ takes the value $q$ in the dataset $\mathbf{D}$. Note that if $q \notin \mathbf{A}_j$, $f_j(q)$ = 0.
The range of similarity values for matches in the \textsc{Goodall-4 Similarity} is \([\frac{2}{n(n-1)},1]\).
\begin{figure*}[!htb]
\vspace{-2cm}
  \centering
  \includegraphics[width=0.95\linewidth]{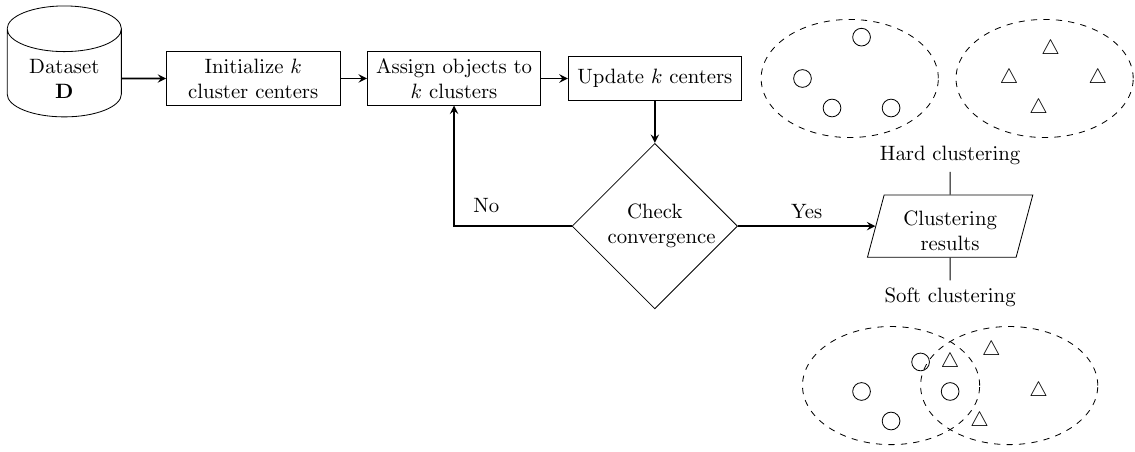}
  \caption{\textsc{Hard} and \textsc{Fuzzy Partitional Clustering}}
  \label{fig:hard-fuzzy}
\end{figure*}

Some other similarity measures such as \textsc{Eskin Similarity} \citep{eskin2002geometric}, \textsc{Occurrence Frequency Similarity}, and \text{Inverse Occurrence Frequency Similarity (IOF)} \citep{boriah2008similarity} assign a value of $1$ for matched attribute values, while defining different ways for mismatched values. The \textsc{Eskin Similarity} measure can be calculated as follows:
\begin{equation}
\textnormal{sim}(x_{1j},x_{2j}) =
\begin{cases}
      1 & \text{if } x_{1j} = x_{2j} \\
      \frac{{n_j}^2}{{n_j}^2+2} & \text{otherwise}
   \end{cases}
\end{equation}
where $n_j$ is the number of values in the attribute $\mathbf{A}_j$. The range of similarity values for mismatches in the \textsc{Eskin Similarity} is \([\frac{2}{3},\frac{n^2}{n^2+2}]\).

Recently, \cite{nguyen2023method} proposed an \textsc{Information Theoretic-Based Similarity} measure, where the similarity between two categorical values is calculated as follows:
\begin{equation}
\textnormal{sim}(x_{1j},x_{2j}) = \frac{2\log P_j({x_{1j},x_{2j}})}{\log P_j(x_{1j}) + \log P_j(x_{2j})}
\end{equation}
\textnormal{
where $P_j(p) = \frac{f_j(p)}{n}$ and $P_j(p,q) = \frac{f_j(p,q)}{n}$, $f_j(p)$ is the number of times attribute $\mathbf{A}_j$ takes the value $p$ in the dataset $\mathbf{D}$, $f_j(p,q)$ is the number of times attribute $\mathbf{A}_j$ takes either the value $p$ or $q$ in the dataset $\mathbf{D}$. The formulation shows that when the two values are identical, their similarity equals $1$.
}
\subsubsection{\textsc{Hierarchical} versus \textsc{Partitional Clustering}}
Hierarchical clustering aims to create a dendrogram, which represents relationships among data points, using either an agglomerative (bottom-up) or divisive (top-down) approach \citep{han2022data}. The workflow of hierarchical clustering for categorical data is shown in Figure \ref{fig:hierarchical_clustering}. In agglomerative clustering, each object starts as a separate cluster. At each step, the two closest clusters (based on a distance metric) are iteratively merged until only a single cluster remains. This merging process is typically guided by a distance or similarity measure, and the dendrogram reflects these pairwise merges. To determine the optimal number of clusters, one can cut the dendrogram at a level that best balances the number of clusters and the homogeneity of the data within each cluster. Conversely, divisive clustering starts with all objects in a single cluster and recursively splits it into smaller clusters. At each step, the cluster that is most dissimilar to the others is divided until each object forms its cluster. The dendrogram in divisive clustering shows how clusters are recursively split, and the optimal number of clusters can be determined by analyzing where the splits yield the most meaningful groupings. 

Partitional clustering (nonhierarchical or flat clustering) divides a dataset into a specified number of clusters $k$ by directly partitioning the data, without creating hierarchical structures. The aim is to optimize a certain objective function, such as minimizing intra-cluster distances or maximizing inter-cluster separation \citep{kaufman2009finding}. One of the most popular and simple partitioning clustering algorithms, \textsc{K-means}, has many variants. \cite{steinhaus1956} was among the first to discuss the concept of partitioning points into clusters. In 1957, \cite{lloyd1982least} developed an algorithm that was officially published in 1982. This algorithm is the standard version of \textsc{K-means} commonly used in data mining and machine learning. The pseudo-code for Lloyd's algorithm is presented in Algorithm \ref{algo:lloyd}. The clustering results of this algorithm are highly sensitive to the initialization of centroids. Specifically, the algorithm is prone to converging to a local optimum depending on the initial configuration of centroids.

\begin{algorithm}
\caption{Lloyd's \textsc{K-means} \citep{lloyd1982least}}
\begin{algorithmic}[1]
    \STATE Initialize by randomly assigning each data point to one of the \(k\) clusters.
    \REPEAT
        \STATE Calculate the centroids of each cluster as the mean of all data points currently assigned to that cluster.
        \STATE Reassign each data point to the cluster with the nearest centroid.
    \UNTIL{no data points change clusters (convergence) or centroids no longer change}
\end{algorithmic}
\label{algo:lloyd}
\end{algorithm}

\cite{macqueen1967some} provides another variant of \textsc{K-means} as shown in Algorithm \ref{algo:macqueen}. Instead of recalculating the centroids after all points have been assigned (as in Lloyd’s algorithm), MacQueen's algorithm updates centroids incrementally as each point is assigned to a cluster. This method may converge faster in some cases due to incremental updates. However, it is not always faster than Lloyd’s, and it might be more sensitive to the order of data points, especially in larger datasets \citep{izenman2008,reddy2013survey}.

\begin{algorithm}
\caption{MacQueen's \textsc{K-means} \citep{macqueen1967some}}
\begin{algorithmic}[1]
    \STATE Randomly select $k$ data points as initial centroids 
    \REPEAT
        \FOR{each data point $x_i$}
            \STATE Assign $x_i$ to the nearest centroid $c_j$.
            \STATE Update the centroid $c_j$ as mean over all data points assigned to it so far, including $x_i$.
        \ENDFOR
    \UNTIL{convergence}
\end{algorithmic}
\label{algo:macqueen}
\end{algorithm}

Generally, \textsc{K-means} is efficient with large datasets and is computationally straightforward, with a time complexity that generally scales linearly with the number of data points \citep{aggarwal2013introduction}. However, it often converges to a local rather than a global optimum, depending on the initial centroid positions \citep{macqueen1967some}. In addition, the algorithm tends to form clusters with convex, spherical shapes, making it less suitable for complex or non-convex structures \citep{anderberg1973cluster}. Despite these limitations, \textsc{K-means} remains popular due to its simplicity and effectiveness in many practical applications \citep{han2022data,wu2008top,torres2020comprehensive,torres2022scalable,khan2024enhanced}.

\cite{huang1997fast,huang1998extensions} introduced \textsc{K-modes} as an extension of \textsc{K-means} for clustering categorical data by incorporating (1) a \textsc{Simple Matching Dissimilarity} measure for categorical objects (Equation \ref{eq:distance_simple_matching}), (2) the use of \emph{modes} rather than \emph{means} as cluster representatives, and (3) a frequency-based method to update the \emph{modes} in a \textsc{K-means}-like iterative process to minimize the clustering cost function. \textsc{K-modes}'s process follows the structure of Lloyd's version of \textsc{K-means} as outlined in Algorithm \ref{algo:kmodes}. Like \textsc{K-means}, \textsc{K-modes} often converge to locally optimal solutions, with results that depend on the initial modes and the order of objects in the dataset \citep{huang1998extensions}.

\begin{algorithm}
\caption{\textsc{K-modes} \citep{huang1997fast,huang1998extensions}}
\begin{algorithmic}[1]
    \STATE Initialize by selecting $k$ initial \emph{modes}, one for each cluster.
    \STATE Allocate each data object to the cluster whose \emph{mode} is the nearest according to the \textsc{Simple Matching Dissimilarity} measure (Equation \ref{eq:distance_simple_matching}).
    \STATE Update the \emph{mode} of the cluster after each allocation.
    \REPEAT
        \FOR{each object in the dataset}
            \STATE Retest the dissimilarity of the object against the current \emph{modes}.
            \IF{the nearest \emph{mode} for the object belongs to a different cluster}
                \STATE Reallocate the object to that cluster.
                \STATE Update the \emph{modes} of both affected clusters.
            \ENDIF
        \ENDFOR
    \UNTIL{no object changes clusters after a full cycle test of the dataset}
\end{algorithmic}
\label{algo:kmodes}
\end{algorithm}

Mathematically, given a dataset $\mathbf{D}$ =$\{\mathbf{x}_1, \dots, \mathbf{x}_n\}$ of $n$ data points, each with $m$ features. \textsc{K-modes} aims to partition $\mathbf{D}$ into $k$ groups by minimizing the following objective (cost) function \citep{huang1998extensions}:\\\\
Minimize
\begin{equation} 
\label{eq:objective_function}
\mathbf{P}(\mathbf{U},\mathbf{C}) = \sum_{l=1}^{k}\sum_{i=1}^{n}\sum_{j=1}^{m}\mathbf{u}_{il}\times \delta(\mathbf{x}_{ij},\mathbf{c}_{lj})
\end{equation}
subject to:
\begin{equation}
\begin{cases}
 \mathbf{u}_{il} \in \{0,1\} \\
 \sum_{l=1}^{k}{\mathbf{u}_{il}=1} (1 \leq i \leq n)
\end{cases}
\end{equation}
where $\mathbf{U}=[\mathbf{u}_{il}]_{n\times k}$ is an $n\times k$ partition matrix, and $\mathbf{C} = \{\mathbf{C}_l, l = 1,\dots,k \}$ is a set of cluster centers. Each cluster center $\mathbf{C}_l$ consists of $m$ values, each representing the \emph{mode} (the most frequent value) of an attribute within that cluster. The function $\delta()$ denotes the \textsc{Simple Matching Dissimilarity} measure defined in Equation \ref{eq:simple_matching_dis}, used to measure the distance between data points and cluster centers. 

In general, the framework of partitional clustering algorithms is illustrated in Figure \ref{fig:hard-fuzzy} and follows the process outlined below \cite{huang1998extensions}:
\begin{enumerate}
    \item Choose an initial set of cluster centers $\mathbf{C}^{(0)}$ = $\{\mathbf{C}^{(0)}_1$, $\mathbf{C}^{(0)}_2$, $\dots$, $\mathbf{C}^{(0)}_k\}$, and set $t = 0$.
    
    \item Determine \(\mathbf{C}^{(t+1)}\) such that \(\mathbf{P}(\mathbf{U}^{(t)}, \mathbf{C}^{(t+1)})\) is minimized. If \(\mathbf{P}(\mathbf{U}^{(t)}, \mathbf{C}^{(t+1)}) = \mathbf{P}(\mathbf{U}^{(t)}, \mathbf{C}^{(t)})\), then stop; otherwise, go to step 3.
    
    \item Determine \(\mathbf{U}^{(t+1)}\) such that \(\mathbf{P}(\mathbf{U}^{(t+1)}, \mathbf{C}^{(t+1)})\) is minimized. If \(\mathbf{P}(\mathbf{U}^{(t+1)}, \mathbf{C}^{(t+1)}) = \mathbf{P}(\mathbf{U}^{(t)}, \mathbf{C}^{(t+1)})\), then stop; otherwise, set \(t = t + 1\) and return to step 2.
\end{enumerate}

With a dissimilarity measure, the partition matrix in step 2 can be determined as follows:
\begin{center}
if $\textnormal{dis}(\mathbf{x}_i, \mathbf{C}_l) \leq \textnormal{dis}(\mathbf{x}_i, \mathbf{C}_{l'})$, then \\
$\mathbf{u}_{i,l}$ = 1, and $\mathbf{u}_{i,l'}$ = 0,	for $1 \leq l' \leq k$, $l \neq l'$	
\end{center}

Partitional clustering algorithms can be further classified into \emph{hard partitional clustering} and \emph{soft partitional clustering}. Hard or crisp clustering partitions the dataset into distinct non-overlapping clusters, assigning each data point to exactly one cluster (a single group). \textsc{K-modes} \citep{huang1997fast, huang1998extensions} and many other algorithms shown in Figure \ref{fig:tree_taxonomy} belong to this group. 

On the other hand, soft (fuzzy) clustering allows data points to belong to multiple clusters simultaneously, assigning each data point a membership value between 0 (indicating the lowest possible membership degree) and 1 (indicating the highest possible membership degree), which reflects its degree of belongingness to each cluster \cite{tan2019introduction}. Likewise, probabilistic clustering methods determine the probability of each point belonging to each cluster, ensuring that these probabilities collectively add up to 1.

\cite{huang1999fuzzy} introduced a fuzzy extension of the \textsc{K-modes} algorithm, named \textsc{Fuzzy K-modes}. In this approach, the objective function in Equation \ref{eq:objective_function} is modified to incorporate fuzzy memberships, and can be expressed as follows for \textsc{Fuzzy K-modes}:\\\\
Minimize
\begin{equation} 
\label{eq:objective_function_fuzzy}
\mathbf{P}(\mathbf{U},\mathbf{C}) = \sum_{l=1}^{k}\sum_{i=1}^{n}\sum_{j=1}^{m}\mathbf{u}_{il}^{\alpha}\times \delta(\mathbf{x}_{ij},\mathbf{c}_{lj})
\end{equation}
subject to:
\begin{equation}
\label{eq:fuzzy_constraints}
\begin{cases}
 \textnormal{for each}~\mathbf{x}_i, \sum_{l=1}^{k} \mathbf{u}_{il} = 1, u_{il} \in [0, 1]\\
 \textnormal{for each}~\mathbf{C}_l, 0 < \sum_{i=1}^{n} \mathbf{u}_{il} < n
\end{cases}
\end{equation}
Here $\alpha > 1$ is the fuzziness parameter that controls the degree of \emph{fuzziness} in the memberships. In \textsc{Fuzzy K-modes}, the membership values $\mathbf{u}_{il}$ are determined by the inverse distance between each data point $\mathbf{x}_i$ and the cluster centers \( \mathbf{C}_l \), weighted according to the fuzziness parameter $\alpha$:
\begin{equation}
\label{eq:membership_values}
\mathbf{u}_{il} = \frac{\left[\text{dis}(\mathbf{x}_i, \mathbf{C}_l)\right]^{-1/(\alpha -1)}}{\sum_{l'=1}^{k} \left[\text{dis}(\mathbf{x}_i, \mathbf{C}_{l'})\right]^{-1/(\alpha -1)}}
\end{equation}
\begin{figure*}[!htb]
\vspace{-2cm}
  \centering
  \includegraphics[width=\linewidth]{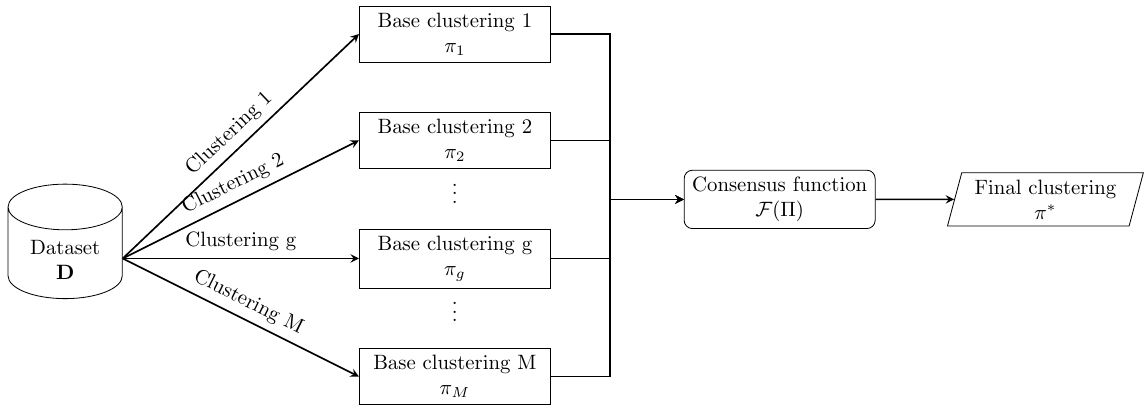}
  \caption{\textsc{Ensemble Clustering}}
  \label{fig:ensemble_clustering}
\end{figure*}
In Equation \ref{eq:fuzzy_constraints}, the first constraint allows each data point $\mathbf{x}_i$ to belong to multiple clusters with varying degrees of membership, and the sum of its membership degrees across all clusters equals $1$. The second constraint ensures that each cluster $\mathbf{C}_l$ includes at least one data point with a non-zero membership value while preventing any cluster from assigning all points a membership value of $1$.

In the \textsc{Fuzzy K-modes} \cite{huang1999fuzzy}, updating the cluster centers $\mathbf{C}_l$ involves computing the \emph{mode} of each attribute across all data points, weighted by their membership degrees. Specifically, for each attribute $\mathbf{A}_j$ and cluster $\mathbf{C}_l$, the updated cluster center value $\mathbf{c}_{lj}$ is determined by selecting the categorical value that maximizes the weighted frequency:

\begin{equation} 
\mathbf{c}_{lj} = \underset{a \in \text{DOM}(A_j)}{\operatorname{argmax}} \sum_{i=1}^{n} \mathbf{u}_{il}^\alpha \, \delta(x_{ij}, a)
\end{equation}

It ensures that the most representative categorical value (the \emph{mode}) is selected for each cluster center attribute, taking into account the degrees of membership of data points to the cluster.

\subsubsection{\textsc{Ensemble Clustering}}
Ensemble clustering involves integrating multiple clustering outputs, possibly from different algorithms or multiple runs, to generate a single consensus partition of the original dataset \citep{strehl2002cluster}. This approach leverages the diversity and strengths of various individual clustering algorithms to improve the overall clustering accuracy and robustness \citep{he2005cluster}. The workflow of ensemble clustering for categorical data is shown in Figure \ref{fig:ensemble_clustering}. A typical clustering ensemble framework generally consists of two key processes: (1) generating a set of base clustering results, and (2) combining these results into a final clustering using a consensus function \citep{zhao2017clustering}. Both the base clusterings and the consensus function significantly affect the performance of the clustering ensemble.

To further improve clustering quality, several methods have been developed to improve clustering quality. These methods evaluate and select a subset of base partitions based on diversity and quality—two important factors to improve the ensemble solution \citep{golalipour2021clustering}. These methods are widely known as \textsc{Clustering Ensemble Selection}. The main objective is to generate clustering results based on a subset of base partitions, achieving performance as good as or better than that obtained by using all clustering solutions \citep{zhao2017clustering}. 

In \textsc{Clustering Ensemble Selection}, evaluation metrics are crucial for selecting a subset of base clusterings, with two main types commonly used: \emph{diversity measures} and \emph{quality measures} \cite{zhao2017clustering}. Diversity measures quantify the disagreement among base clusterings by capturing different data structures. Commonly used metrics are \textsc{Adjusted Rand Index (ARI)} and \textsc{Normalized Mutual Information (NMI)}. Quality measures assess the performance of individual base clusterings using internal validation indices without external references, such as the \textsc{Silhouette Coefficient}, \textsc{Davies-Bouldin Index}, and \textsc{Calinski-Harabasz Index}. High-quality clusterings are likely to contribute positively to the final ensemble result. The formulations of the above metrics can be found in section \ref{sec:evaluation}.

\begin{figure*}[!htb]
\vspace{-2cm}
  \centering
  \includegraphics[width=\linewidth]{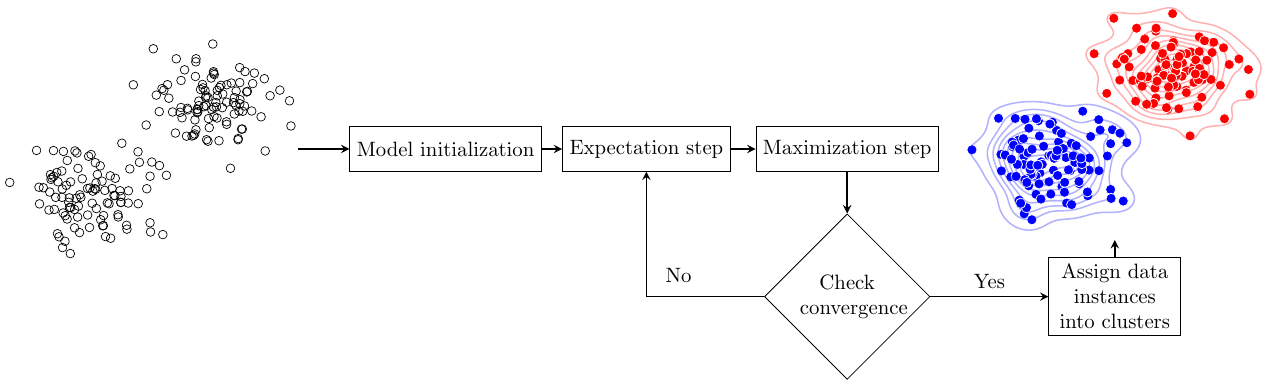}
  \caption{\textsc{Model Based Clustering}}
  \label{fig:model_clustering}
\end{figure*}

Below, we provide the mathematical formulation for ensemble clustering that is widely used in the literature for both numerical and categorical data \citep{golalipour2021clustering}. Given a dataset $\mathbf{D} = \{ x_i \}_{i=1}^n$, where $x_i$ represents the individual data points in the dataset. Let $\mathbf{\Pi} = \{ \mathbf{\pi}_1, \mathbf{\pi}_2, \ldots, \mathbf{\pi}_M \}$ be a cluster ensemble with $M$ base clusterings. Each base clustering $\mathbf{\pi}_g$ (for $1 \leq g \leq M$) consists of a set of clusters:

\begin{equation}
\mathbf{\pi}_g = \{ \mathbf{C}_1^g, \mathbf{C}_2^g, \ldots, \mathbf{C}_{k_g}^g \}
\end{equation}
where $k_g$ is the number of clusters in the $g^\text{th}$ base clustering. These clusters satisfy the conditions:

\begin{equation}
\bigcup_{j=1}^{k_g} \mathbf{C}_j^g = \mathbf{D}
\end{equation}
Finally, a final clustering solution $\mathbf{\pi}^* = \{\mathbf{C}_1, \mathbf{C}_2, \ldots, \mathbf{C}_k \}$ is determined to best encapsulate the information from the cluster ensemble $\mathbf{\Pi}$. This is achieved by applying a consensus function $\mathcal{F}$ to combine the base clusterings into a single clustering \citep{golalipour2021clustering}:

\begin{equation}
\mathbf{\pi}^* = \underset{\mathbf{\pi} \in \mathbf{\Pi}}{\arg \max} \left\{ \frac{1}{M} \sum_{g=1}^{M} \Phi(\mathbf{\pi}_g, \mathbf{\pi}) \right\}
\end{equation}
where $\mathbf{\pi}$ denotes a candidate clustering solution being evaluated as the \emph{potential consensus clustering}, $\mathbf{\pi}^*$ is the \emph{consensus clustering solution} that maximizes the overall similarity with the base clusterings in the ensemble. The function $\Phi(\mathbf{\pi}_g, \mathbf{\pi})$ represents a \emph{metric} such as \textsc{Purity}, \textsc{NMI}, \textsc{ARI} used to measure the similarity between the $g^{th}$ base clustering $\mathbf{\pi}_g$ and the candidate clustering $\mathbf{\pi}$. The operation $\underset{\mathbf{\pi} \in \mathbf{\Pi}}{\arg \max}$ identifies the clustering $\mathbf{\pi}$ within the ensemble $\mathbf{\Pi}$ that maximizes the objective function. Finally, $\frac{1}{M} \sum_{g=1}^{M}$ computes the \textit{average similarity} across all $M$ base clusterings, providing an aggregated similarity score for the candidate clustering.

\subsubsection{\textsc{Model Based Clustering}}
Model-based clustering is a statistical approach that assumes the data are generated by a mixture of underlying probability distributions, where each distribution corresponds to a different cluster \citep{chen2012model}. This approach assumes a generative model for the data, in which a specific parametric distribution represents each cluster. By fitting these distributions to the data, the model identifies distinct clusters that best explain the underlying structure.

\begin{figure*}[!htb]
\vspace{-2cm}
  \centering
  \includegraphics[width=\linewidth]{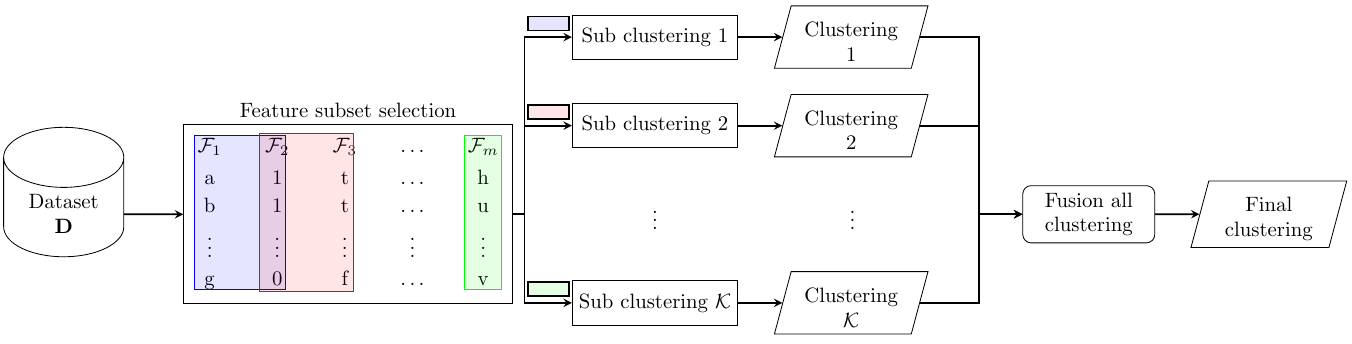}
  \caption{\textsc{Subspace Clustering}}
  \label{fig:subspace_clustering}
\end{figure*}

The general framework of model-based clustering is shown in Figure \ref{fig:model_clustering} with the following process \citep{bouveyron2014model}:
\begin{enumerate}
    \item \textbf{Model initialization}: choose a model suitable for categorical data, such as a Latent Class Analysis \citep{mccutcheon1987latent}, and initialize the parameters of the model (e.g., the number of clusters, initial probabilities).
    \item \textbf{Expectation step (E-step)}: calculate the expected value of the likelihood function, with respect to the conditional distribution of the latent variables given the observed data and current parameter estimates.
    \item \textbf{Maximization step (M-step)}: update the parameters of the model to maximize the likelihood function based on the posterior probabilities computed in the E-step.
    \item \textbf{Check convergence}: repeat the E-step and M-step until convergence is achieved (i.e., the parameters do not change significantly between iterations).
    \item \textbf{Cluster assignment}: assign each data point to the cluster with the highest posterior probability.
\end{enumerate}

Mathematically, assume that the data $\mathbf{D} = \{ \mathbf{x}_i \}_{i=1}^n$ is drawn from a mixture of $k$ components (clusters), each having its probability distribution. Each data point $\mathbf{x}_i$ belongs to a latent cluster denoted by $C_i$, which is not directly observed. The clustering is performed by estimating the posterior probabilities of the latent variables and the parameters of the distributions. Let $\mathbf{x}_i = (\mathbf{x}_{i1}, \mathbf{x}_{i2}, \dots, \mathbf{x}_{im})$ be an observed data point consisting of $m$ categorical variables, and $C_i$ be the latent variable representing the cluster membership for $\mathbf{x}_i$. The model can be formulated as follows:
\begin{equation}
P(\mathbf{x}_i) = \sum_{l=1}^{k} \pi_l \, P(\mathbf{x}_i \mid C_i = l)
\end{equation}
where:
\begin{itemize}
    \item $k$ is the number of clusters.
    \item $\pi_l = P(C_i = l)$ is the mixing proportion for cluster $l$, satisfying $\sum_{l=1}^{k} \pi_l = 1$.
    \item $P(\mathbf{x}_i \mid C_i = l)$ is the conditional probability of observing $\mathbf{x}_i$ given that the latent variable $C_i$ indicates cluster $l$.
\end{itemize}

For categorical data, the conditional probability $P(\mathbf{x}_i | C = l)$ is often modeled using a multinomial distribution \citep{gollini2014mixture}. The parameters of the model, including the mixing proportions $\mathbf{\pi}_l$ and the parameters of the multinomial distributions, are typically estimated using the \textsc{Expectation Maximization} (EM) algorithm. The EM algorithm iteratively estimates the model parameters in the presence of latent variables, which are not directly observed.

In section \ref{sec:taxonomy}, we explore several model-based clustering algorithms \citep{chen2012model,gollini2014mixture,jacques2018model,zhang2019unified} designed for categorical data.
\subsubsection{\textsc{Subspace Clustering}}
\begin{figure*}[!htb]
  \centering
  \includegraphics[width=\linewidth]{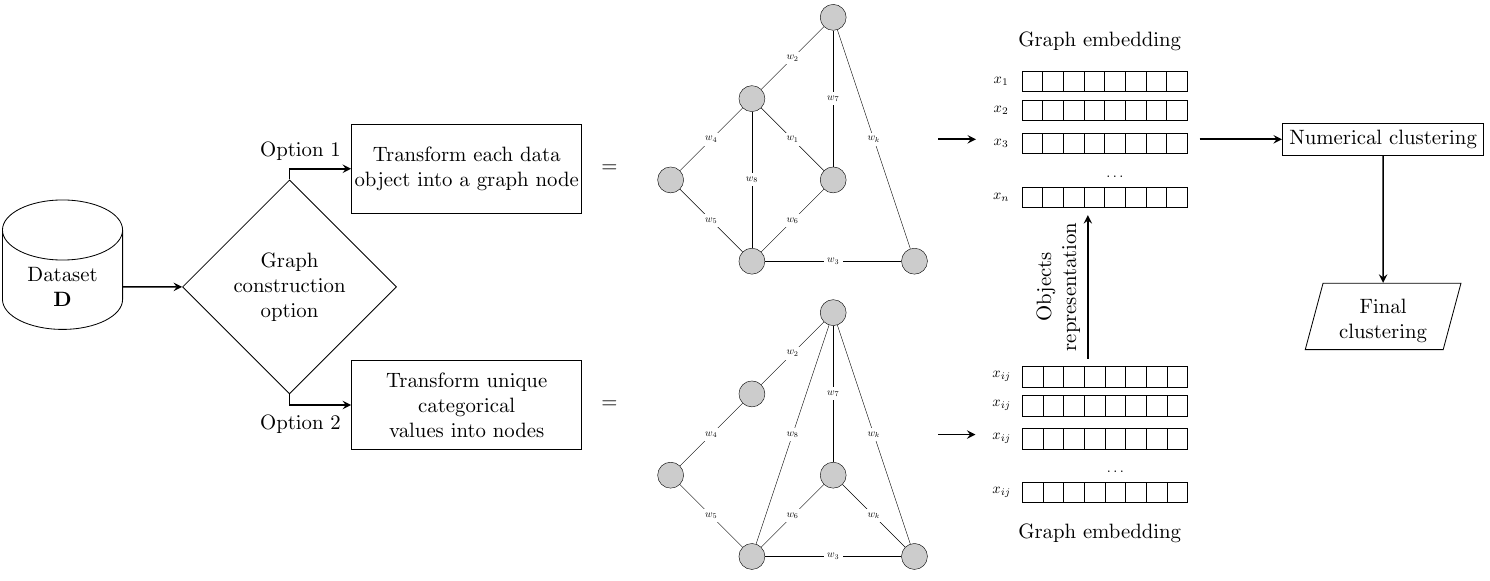}
  \caption{\textsc{Graph Based Clustering}}
  \label{fig:Graph_clustering}
\end{figure*}
Subspace clustering aims to discover clusters within different subspaces of a high-dimensional dataset. The key idea behind subspace clustering is to search for clusters in subsets of the dimensions of the complete data \citep{gan2004subspace,cao2013weighting}. This approach is useful to handle high-dimensional data where the relevant clusters may only be present in a small subset of the dimensions, rather than the full-dimensional space \citep{chen2016soft}. The general framework of subspace clustering for categorical data is shown in Figure \ref{fig:subspace_clustering} and the following process \citep{parsons2004subspace}:
\begin{enumerate}
    \item \textbf{Dimension selection}: identify relevant subspaces by using methods such as grid search, which systematically evaluates different combinations of dimensions; random search, which samples combinations randomly; heuristic methods based on domain knowledge; or dimensionality reduction techniques like Principal component analysis (PCA) or Uniform Manifold Approximation and Projection (UMAP) \citep{mcinnes2018umap}.
    \item \textbf{Clustering within subspaces}: apply clustering algorithms to the selected subspaces. This can involve various methods like partitional clustering, hierarchical clustering, or model-based clustering.
    \item \textbf{Combining subspaces}: integrate clusters found in different subspaces, several approaches can be employed. One common method is to use consensus functions, similar to ensemble clustering, where clusters from different subspaces are aggregated based on their similarity or overlap to produce a unified result. Alternatively, clusters may be merged based on criteria such as overlap or distance measures, or an additional round of clustering can be performed on the combined results to refine and consolidate the clusters into a final comprehensive solution.
\end{enumerate}
\begin{figure*}[!htb]
\vspace{-2cm}
  \centering
  \includegraphics[width=0.9\linewidth]{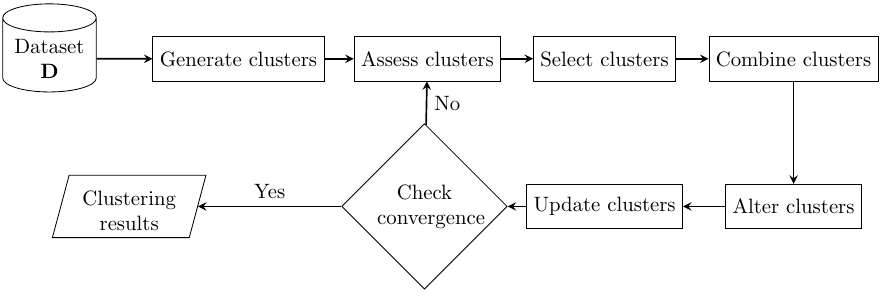}
  \caption{\textsc{Genetic Algorithm Based Clustering}}
  \label{fig:GA_clustering}
\end{figure*}
Mathematically, let $\mathbf{A} = \{\mathbf{A}_1, \mathbf{A}_2, \ldots, \mathbf{A}_m\}$ denote a set of $m$ distinct categorical attributes and $\mathbf{D} = \{\mathbf{x}_1, \mathbf{x}_2, \ldots, \mathbf{x}_n\}$ be a categorical dataset of $n$ data points. A subspace $\mathbf{S}$ is defined as  \citep{zaki2005clicks}:
\begin{equation}
\mathbf{S} = \{\mathbf{A}_{j1}, \mathbf{A}_{j2}, \ldots, \mathbf{A}_{jk} \mid 1 \leq j1 < j2 < \ldots < jk \leq m \}.
\end{equation}
The projection of a dataset $\mathbf{D}$ onto a subspace $\mathbf{S}$ is denoted by $\mathbf{D}_{\mathbf{S}}$ and consists of the tuples from $\mathbf{D}$ restricted to the dimensions in $\mathbf{S}$. The goal of subspace clustering is to identify clusters $\{\mathbf{C}_1$, $\mathbf{C}_2$, $\ldots$, $\mathbf{C}_k\}$ within subspaces such that each cluster $\mathbf{C}_i \subseteq \mathbf{D}$ is a dense region in its corresponding subspace $\mathbf{S}_i$.

A subspace $\mathbf{S}$ is considered dense if the number of data points within this subspace exceeds a certain threshold, which indicates that the subspace is sufficiently populated. Formally, a subspace $\mathbf{S}$ is dense if:
\begin{equation}
\sigma(\mathbf{S}) = \left| \{ \mathbf{x} \in \mathbf{D} \mid \mathbf{x}_{\mathbf{S}} \in \mathbf{S} \} \right| \geq \tau
\end{equation}
where $\tau$ is a density threshold that can vary depending on the specific method used.

In section \ref{sec:taxonomy}, we explore several subspace clustering algorithms \citep{gan2004subspace,zaki2005clicks,parsons2004subspace,cao2013weighting,chen2016soft} designed for categorical data.
\subsubsection{\textsc{Graph Based Clustering}} \label{sec:graph_clustering}
Graph-based clustering organizes data as a graph, where nodes represent data points and weighted edges indicate the degree of similarity between them. Nodes within the same cluster are densely connected, while connections to nodes outside the cluster are sparse or absent. The objective of a graph-based clustering algorithm is to partition the graph into clusters by analyzing the edge structure, maximizing the number or weight of edges within clusters \citep{adewole2017malicious}. Given a dataset $\mathbf{D} = \{\mathbf{x}_1, \mathbf{x}_2, \ldots, \mathbf{x}_n\}$ formed from $m$ distinct categorical attributes $\mathbf{A} = \{\mathbf{A}_1, \mathbf{A}_2, \ldots, \mathbf{A}_m\}$, the general framework of graph-based clustering for categorical data is shown in Figure \ref{fig:Graph_clustering} and the following process \citep{bai2022categorical}:
\begin{enumerate}
    \item \textbf{Graph construction}: choose between two options for graph construction:
    \begin{enumerate}[label=(\alph*), leftmargin=0.1cm]
        \item Transform each data point $\mathbf{x}_i$ into a node in the graph $G = (V, E)$, where $V$ is the set of all data points and $E$ is the set of edges between nodes weighted by similarity measures. For each pair of data points $(x_i, x_j)$, an edge $e_{ij}$ is added with weight $w_{ij}$, computed as:
        \begin{equation}
        w_{ij} = \frac{1}{m} \sum_{k=1}^{m} \text{sim}(A_k(x_i), A_k(x_j)),
        \end{equation}
        where $A_k(x)$ represents the $k$-th attribute value of data object $x$, and $\text{sim}()$ is the similarity measure between attribute values.

        \item Transform the categorical data $\mathbf{D}$ into a graph $G = (V, E)$, where $V$ is the set of nodes corresponding to the unique categorical values across all attributes, and $E$ is the set of edges between nodes weighted by similarity measures. For each pair of nodes $(v_i, v_j)$, an edge $e_{ij}$ is added with weight $w_{ij}$ computed as:
        \begin{equation}
        w_{ij} = \text{sim}(v_i, v_j),
        \end{equation}
        where higher weights indicate higher similarity, the function $\text{sim}()$ represents the similarity measure, such as the \textsc{Jaccard} coefficient.
    \end{enumerate}

    \item \textbf{Graph embedding}: embed the constructed graph into a low-dimensional space where similar nodes are closer together. This embedding aims to preserve the structural information of the original graph while reducing dimensionality. Typically, methods like spectral embedding or neural network-based techniques are employed to capture the local neighborhood structure, ensuring that nodes with strong connections remain close in the embedded space. Other approaches, like t-distributed stochastic neighbor embedding (t-SNE) \citep{van2008visualizing} or Uniform Manifold Approximation and Projection (UMAP) \citep{mcinnes2018umap}, can also capture more global relationships, providing a balance between local and global structure preservation.
    
    \item \textbf{Object representation} (only applicable if (b) in step 1): represent each object $\mathbf{x}_i$ as a numerical vector by integrating the embeddings of its categorical values using one of the following operations:
    \begin{itemize}[leftmargin=10pt]
        \item Joint operation: concatenate the embeddings of all categorical values to form a high-dimensional vector.
        \item Mean operation: average the embeddings of all categorical values to form a lower-dimensional vector.
    \end{itemize}
    
    \item \textbf{Clustering phase}: apply numerical clustering algorithms directly to the graph structure or the embeddings obtained in the previous step. Common choices are partitional, hierarchical, spectral clustering, and community detection, which identifies groups of nodes more densely connected internally than the rest of the graph.
\end{enumerate}

In section \ref{sec:taxonomy}, we explore several graph-based clustering algorithms \citep{david2012spectralcat,bai2022categorical,bandyapadhyay2023parameterized,fomin2023parameterized} designed for categorical data.
\subsubsection{\textsc{Genetic Based Clustering}}
Genetic algorithm-based clustering leverages genetic algorithms (GA) to enhance the clustering results \citep{deng2010g,nguyen2019partition}. GAs are search heuristics that mimic the process of natural selection. They operate on a population of potential solutions, applying selection, crossover, and mutation to evolve better solutions over generations \citep{holland1992adaptation}. In the context of clustering, each potential solution (or chromosome) represents a dataset partitioning into clusters, where each cluster is a subset of the data points. By optimizing cluster assignments through evolutionary processes, these methods provide robust solutions that are less likely to get trapped in local optima \citep{kuo2019genetic}. The general framework of GA-based clustering is shown in Figure \ref{fig:GA_clustering} and the following process \citep{yang2015non}:
\begin{figure*}[!htb]
\vspace{-2cm}
  \centering
  \includegraphics[width=0.9\linewidth]{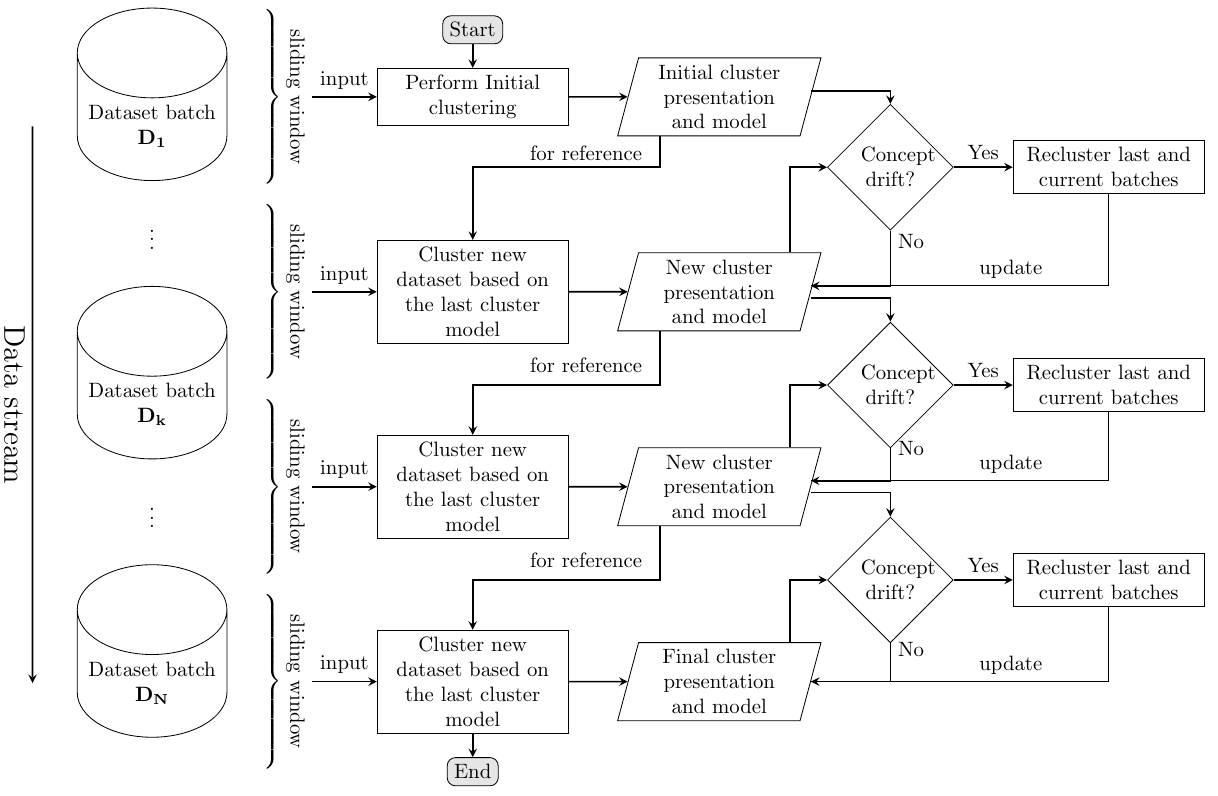}
  \caption{\textsc{Data stream Clustering}}
  \label{fig:stream_clustering}
\end{figure*}

\begin{enumerate}
    \item \textbf{Generate clusters}: generate a random population of chromosomes, each representing a potential clustering of the dataset.
    \item \textbf{Assess clusters}: evaluate each chromosome using a fitness function based on clustering quality, such as intra-cluster homogeneity.
    \item \textbf{Select clusters (selection)}: choose the best chromosomes based on their fitness, favoring those that represent better clustering solutions.
    \item \textbf{Combine clusters (crossover)}: combine parts of two parent chromosomes to create a new offspring by merging cluster assignments from two solutions.
    \item \textbf{Alter clusters (mutation)}: introduce random chromosome changes, altering cluster assignments to maintain diversity and explore new solutions.
    \item \textbf{Update clusters}: form a new population by selecting the best chromosomes from current and new offspring, ensuring evolution towards better clustering.
    \item \textbf{Check convergence}: stop when a predefined number of generations is reached or when convergence occurs, indicated by minimal changes in fitness values. If the condition is not met, return to Assess Clusters.
\end{enumerate}

In section \ref{sec:taxonomy}, we explore several genetic-based clustering algorithms \citep{mukhopadhyay2009multiobjective,deng2010g,yang2015non,kuo2019genetic,kuo2021metaheuristic,jiang2023kernel} designed for categorical data.

\subsubsection{\textsc{Data Stream Clustering}}
Data stream clustering groups similar data points in real time as they continuously flow into a system. This approach is designed to handle the challenges of high-velocity, large-volume, and potentially unbounded data streams by updating clusters incrementally and efficiently \citep{aggarwal2010clustering}. The general framework of data stream clustering is shown in Figure \ref{fig:stream_clustering} and the following process \citep{bai2016optimization,chen2008catching}:
\begin{enumerate}
\item \textbf{Initial clustering}: perform clustering on the first batch of data to identify initial clusters. This step involves computing cluster centers (or representatives) and saving the initial cluster configurations, which will be used as a starting point for processing new data.
\item \textbf{Window model}: use a sliding window model to focus on the most recent data, discarding older data points that may no longer be relevant.
\item \textbf{Cluster representation}: maintain and update statistical summaries or models for each cluster, such as the cluster's center or spread. This helps capture the essential characteristics of the clusters as new data points are processed and ensures that the cluster descriptions remain accurate and up-to-date.
\item \textbf{Online clustering}: incrementally assign new data points to existing clusters or form new clusters based on a similarity measure. In case of detected concept drift, combine recent batches for reclustering to accurately reflect the new data distribution.
\item \textbf{Concept drift detection}: monitor the clusters for significant changes that may indicate concept drift, triggering an update in the clustering model.
\item \textbf{Cluster maintenance}: periodically update the clustering model to reflect the current state of the data stream. This process may involve not only merging, splitting, or discarding clusters but also expanding, repositioning, or otherwise adjusting the characteristics of existing clusters to better accommodate new data and accurately represent the evolving data distribution.
\item \textbf{Return final clustering}: the algorithm outputs the current set of clusters, which can be used for downstream analysis or decision-making processes. This output reflects the current clustering state and doesn’t signify the end of the process, as the algorithm continues to update clusters with new data.
\end{enumerate}
Note that these components are not strictly sequential but interwoven, as data arrives continuously and updates occur dynamically. For instance, concept drift detection and cluster maintenance are performed periodically and as needed, integrating with the ongoing clustering process.

In section \ref{sec:taxonomy}, we explore several algorithms \citep{barbara2002coolcat, he2002squeezer, chen2008catching, aggarwal2010clustering, bai2016optimization} designed for data streams with categorical attributes.
\subsection{Historical developments} \label{sec:historical_dev}

\begin{figure*}[!htb]
\vspace{-2cm}
    \centering
    \begin{adjustbox}{max width=\linewidth}
    \begin{tikzpicture}
        \SNAKETEXT{
            1998/-90/15mm/ \cite{huang1997fast,huang1998extensions}\hspace{4pt}  \squaremarker{blue},
            1999/90/17mm/ \hspace{-4mm} \cite{ganti1999cactus} \squaremarker{orange}\\ \cite{huang1999fuzzy}  \squaremarker{green} \\  \hspace{-2mm} \cite{wang1999clustering}  \squaremarker{green},
            2000/-90/15mm/ \hspace{-3mm} \cite{guha2000rock} \squaremarker{pink}\\ \cite{gibson2000clustering} \squaremarker{brown},
            2001/90/15mm/ \cite{oh2001fuzzy}  \squaremarker{green},
            2002/-90/24mm/ \hspace{4mm} \cite{barbara2002coolcat} \squaremarker{gray} \\ \hspace{0.3mm} \cite{yang2002clope}  \squaremarker{blue} \\ \hspace{-2mm} \cite{he2002squeezer}  \squaremarker{blue} \\ \cite{sun2002iterative}  \squaremarker{blue} \\ \hspace{2mm} \cite{ng2002clustering}  \squaremarker{green},
            2004/-90/20mm/\hspace{7mm}\cite{andritsos2004limbo} \squaremarker{pink}\\ \cite{san2004alternative}  \squaremarker{blue} \\ \hspace{-3mm} \cite{li2004entropy} \squaremarker{pink} \\ \cite{kim2004fuzzy}  \squaremarker{green} \\ \cite{he2004organizing} \squaremarker{pink} \\ \hspace{2mm} \cite{gan2004subspace} \squaremarker{orange} ,
            2005/90/15mm/ \\ \hspace{-2mm} \cite{he2005cluster} \squaremarker{cyan} \\ \cite{zaki2005clicks} \squaremarker{orange},
            2007/-90/17mm/ \hspace{-2mm} \cite{gionis2007clustering} \squaremarker{cyan}\\ \cite{parmar2007mmr}  \squaremarker{blue} \\ \hspace{1mm} \cite{ahmad2007method}  \squaremarker{blue},
            2008/90/15mm/\hspace{-5mm} \cite{he2008k}  \squaremarker{blue} \\ \cite{chen2008data}  \squaremarker{blue},
            2009/-90/15mm/\hspace{-5mm} \cite{cao2009new}  \squaremarker{blue} \\ \cite{mukhopadhyay2009multiobjective}  \squaremarker{green},
            2010/-90/15mm/\cite{iam2010link} \squaremarker{cyan} \\ \hspace{-3mm} \cite{aggarwal2010clustering} \squaremarker{gray},
            2011/90/15mm/\cite{bai2011novel}  \squaremarker{blue} \\ \cite{bai2011initialization}  \squaremarker{blue},
            2012/-90/20mm/\hspace{-4mm} \cite{cao2012dissimilarity}  \squaremarker{blue} \\ \cite{ienco2012context} \squaremarker{pink} \\ \cite{chen2012model} \squaremarker{yellow},
            2013/90/12mm/\hspace{-2mm}  \squaremarker{blue} \cite{khan2013cluster} \\  \squaremarker{blue} \cite{cheung2013categorical},
            2014/-90/15mm/\hspace{11mm} \cite{gollini2014mixture} \squaremarker{yellow}\\ \cite{li2014hierarchical} \squaremarker{pink},
            2015/-90/15mm/\hspace{-3mm} \cite{jia2015new}  \squaremarker{blue} \\ \cite{qian2015space}  \squaremarker{blue} \\ \cite{yang2015non}  \squaremarker{green},
            2016/90/15mm/ \cite{jiang2016initialization}  \squaremarker{blue} \\ \cite{chen2016soft} \squaremarker{orange},
            2017/-90/15mm/\cite{zhao2017clustering} \squaremarker{cyan} \\ \cite{cao2017algorithm}  \squaremarker{blue},
            2018/90/17mm/\cite{salem2018fast}  \squaremarker{blue} \\ \hspace{-4mm} \cite{zhu2018many}  \squaremarker{green},
            2019/-90/15mm/\hspace{-4mm} \cite{dinh2019estimating}  \squaremarker{blue}\\ \cite{vsulc2019comparison} \squaremarker{pink},
            2020/-90/17mm/\hspace{7mm}\cite{dinh2020k}  \squaremarker{blue} \\ \hspace{10mm} \cite{zhang2020new}  \squaremarker{blue},
            2021/90/17mm/\hspace{0.2mm} \cite{dinh2021clustering}  \squaremarker{blue} \\ \cite{kuo2021metaheuristic}  \squaremarker{green},
            2022/-90/15mm/\hspace{2mm} \cite{bai2022categorical} \squaremarker{brown} \\ \hspace{-1mm} \cite{xu2022multi} \squaremarker{orange},
            2023/90/17mm/\hspace{5mm} \cite{nguyen2023method}  \squaremarker{blue} \\ \cite{kar2023efficient}  \squaremarker{blue},
            end/end/end/reverse%
        }(0,0)[30mm][47mm][arrowcolor]{5}{10mm}
    \end{tikzpicture}
    \end{adjustbox}
    \begin{center}
    \begin{tabular}{ccccccccc}
    \squaremarker{blue} Hard partitional &
    \squaremarker{green} Fuzzy partitional &
    \squaremarker{pink} Hierarchical&
    \squaremarker{gray} Data stream &
    \squaremarker{orange} Subspace &
    \squaremarker{cyan} Ensemble &
    \squaremarker{brown} Graph-based &
    \squaremarker{yellow} Model-based\\
    \end{tabular}
    \end{center}
    \caption{The timeline of categorical data clustering algorithms from 1997 to 2023}
    \label{fig:historical_development}
\end{figure*}
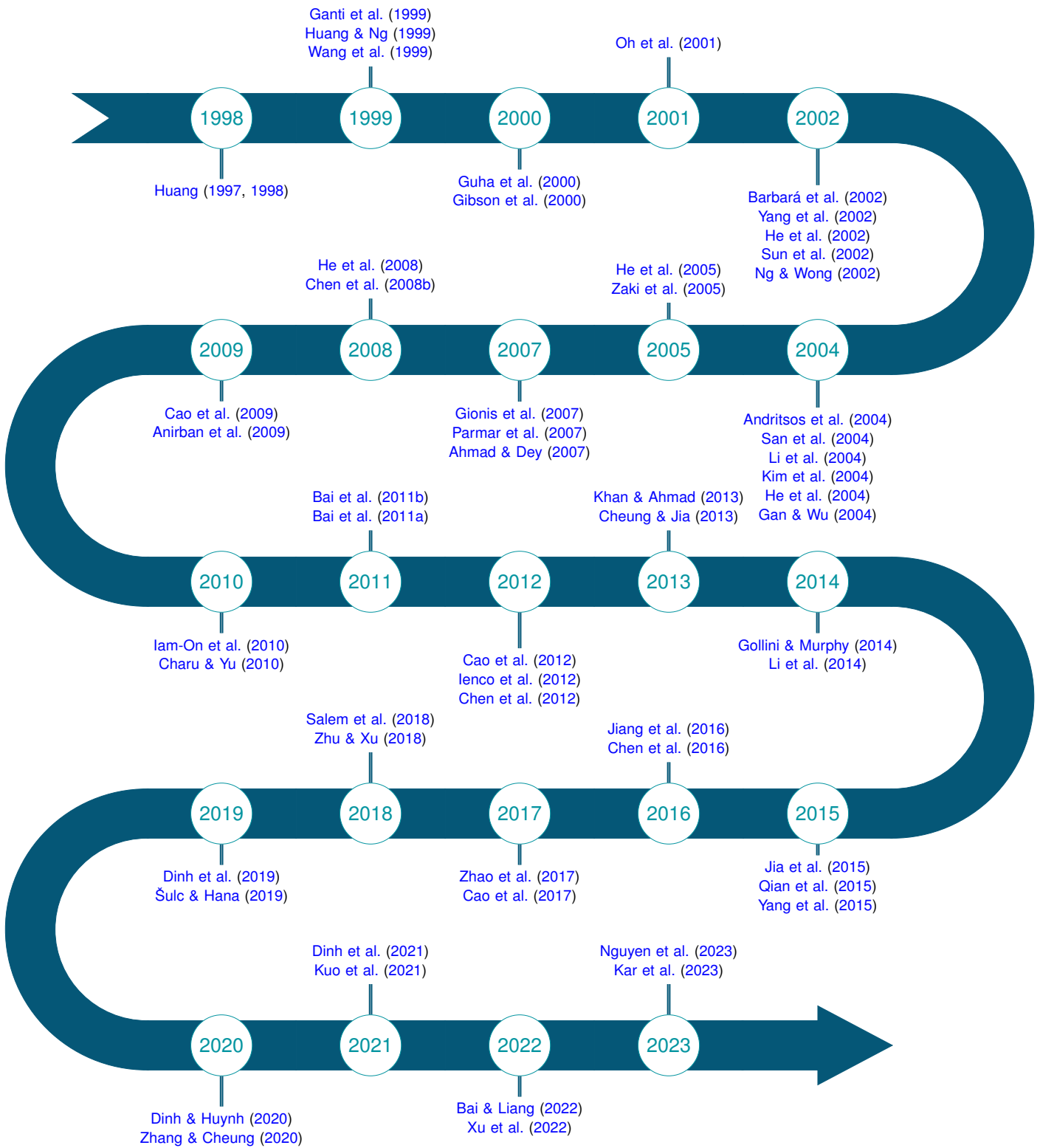
Figure  \ref{fig:historical_development} illustrates the evolution of clustering algorithms for categorical data from 1997 to 2023. For each year, papers with significant impact (having over $100$ citations) are retained. In cases where such papers are not available, the two most cited papers for that year are selected. If there are gaps in the timeline, such as for the years 2003 and 2006, it indicates that no papers met the filter conditions outlined in Section \ref{sec:datasource}. We divide the timeline into three distinct periods.

\subsubsection{From 1997 to 2007}
\textsc{In 1997}, \hypertarget{k-modes}{\cite{huang1997fast, huang1998extensions}} introduced the \textsc{K-modes} algorithm as an extension of \textsc{K-means} for categorical data. \textsc{K-modes} employs the \emph{simple matching} dissimilarity measure for categorical objects, replacing means with \emph{modes} to represent cluster centers. It updates these modes using a \emph{frequency-based approach} throughout the clustering process.

\textsc{In 1999}, \hypertarget{cactus}{\cite{ganti1999cactus}} introduced the \textsc{Cactus} algorithm, which identifies clusters as collections of attribute values exhibiting significant connections. These clusters aim to highlight dense regions in the dataset. The algorithm comprises three main phases: summarization, clustering, and validation. Summarization computes inter-attribute and intra-attribute summaries. The clustering phase uses these summaries to generate candidate clusters, while the validation phase ensures their reliability by examining their support within the dataset. 
\hypertarget{fuzzy_kmodes}{\cite{huang1999fuzzy}} introduced the \textsc{Fuzzy K-modes} algorithm for clustering categorical data, extending the \textsc{K-modes} algorithm by incorporating fuzzy membership values. This extension includes using the \emph{simple matching} dissimilarity measure for categorical objects and finding cluster \emph{modes} instead of \emph{means}. Unlike the original \textsc{K-modes}, \textsc{Fuzzy K-modes} provides a fuzzy partition matrix, which assigns each object a membership grade for each cluster. Experimental results demonstrate that \textsc{Fuzzy K-modes} outperforms \textsc{K-means} and hard \textsc{K-modes} in clustering accuracy for categorical data.
\cite{wang1999clustering} introduced a clustering criterion centered on \emph{large items}, which are individual elements (or attribute values) that appear in at least a user-specified minimum fraction of transactions within a cluster. The cost function comprises two components: the \emph{intra-cluster} cost, which penalizes dissimilarity within clusters by considering the total number of small (non-large) items, and the \emph{inter-cluster} cost, which penalizes similarity across clusters by addressing the duplication of large items. The algorithm iteratively scans transactions, assigning each to an existing or new cluster, and then refines these clusters to minimize the overall cost.

\textsc{In 2000}, \hypertarget{rock}{\cite{guha2000rock}} introduced an agglomerative hierarchical clustering algorithm named \textsc{Rock} that clusters data points based on their \emph{links} rather than distances. Initially, each data point is treated as a separate cluster. The algorithm calculates the number of \emph{links} between every pair of data points, where a link is defined by the number of other data points that have identical feature values with both data points. These shared data points are termed \emph{common neighbors}. The number of links between two clusters is then normalized by the expected number of links, which is computed based on the sizes of the clusters and the total number of data points. This normalization helps account for random chance in the observed number of links. At each step, \textsc{Rock} merges the pair of clusters with the maximum normalized number of links. This process continues until the desired number of clusters is achieved.
In the same year, \hypertarget{stirr}{\cite{gibson2000clustering}} introduced the \textsc{Stirr}, an iterative algorithm that treats categorical data as dynamic systems. It assigns weights to each node (representing data values) and updates them iteratively based on value co-occurrences in tuples, using combining operators such as product, sum, and max. This process leads to weight propagation, allowing indirectly linked nodes—those not directly connected but connected through other nodes—to gain higher weights. The algorithm converges to fixed points or cycles called \emph{basins}, which represent densely correlated value groups across columns. The primary \emph{basin} denotes the main cluster, while non-primary \emph{basin} identify other dense regions and effective data partitions.
\usetikzlibrary{trees,decorations,shadows}

\tikzset{
  level 1/.style={sibling angle=45, level distance=18cm},
  every node/.style={inner sep=7pt, fill=red!80},
  edge from parent/.style={draw, -, out=-90, in=90},
  child_partitional/.style={
    level 2/.style={sibling angle=120},
    every node/.style={fill=olive},
  },
  hard_clustering/.style={
    level 3/.style={sibling angle=9.1, level distance=18cm},
    every node/.style={fill=blue!30, font=\huge},
  },
  fuzzy_clustering/.style={
    level 3/.style={sibling angle=19, level distance=12cm},
    every node/.style={fill=green!50, font=\huge},
  },
  child_hierarchical/.style={
    level 2/.style={sibling angle=25, level distance=9cm},
    every node/.style={fill=pink!50, font=\huge}
  },
  child_data_stream/.style={
    level 2/.style={sibling angle=60, level distance=5cm},
    every node/.style={fill=gray!50, font=\huge},
  },
  child_subspace/.style={
    level 2/.style={sibling angle=45, level distance=6cm},
    every node/.style={fill=orange!50, font=\huge},
  },
  child_ensemble/.style={
    level 2/.style={sibling angle=45, level distance=5cm},
    every node/.style={fill=cyan!50, font=\huge},
  },
  child_genetic/.style={
    level 2/.style={sibling angle=52, level distance=8cm},
    every node/.style={fill=violet!50, font=\huge},
  },
  child_graph/.style={
    level 2/.style={sibling angle=70, level distance=5cm},
    every node/.style={fill=brown!50, font=\huge},
  },
  child_model/.style={
    level 2/.style={sibling angle=90, level distance=5cm},
    every node/.style={fill=yellow!50, font=\huge},
  },
}
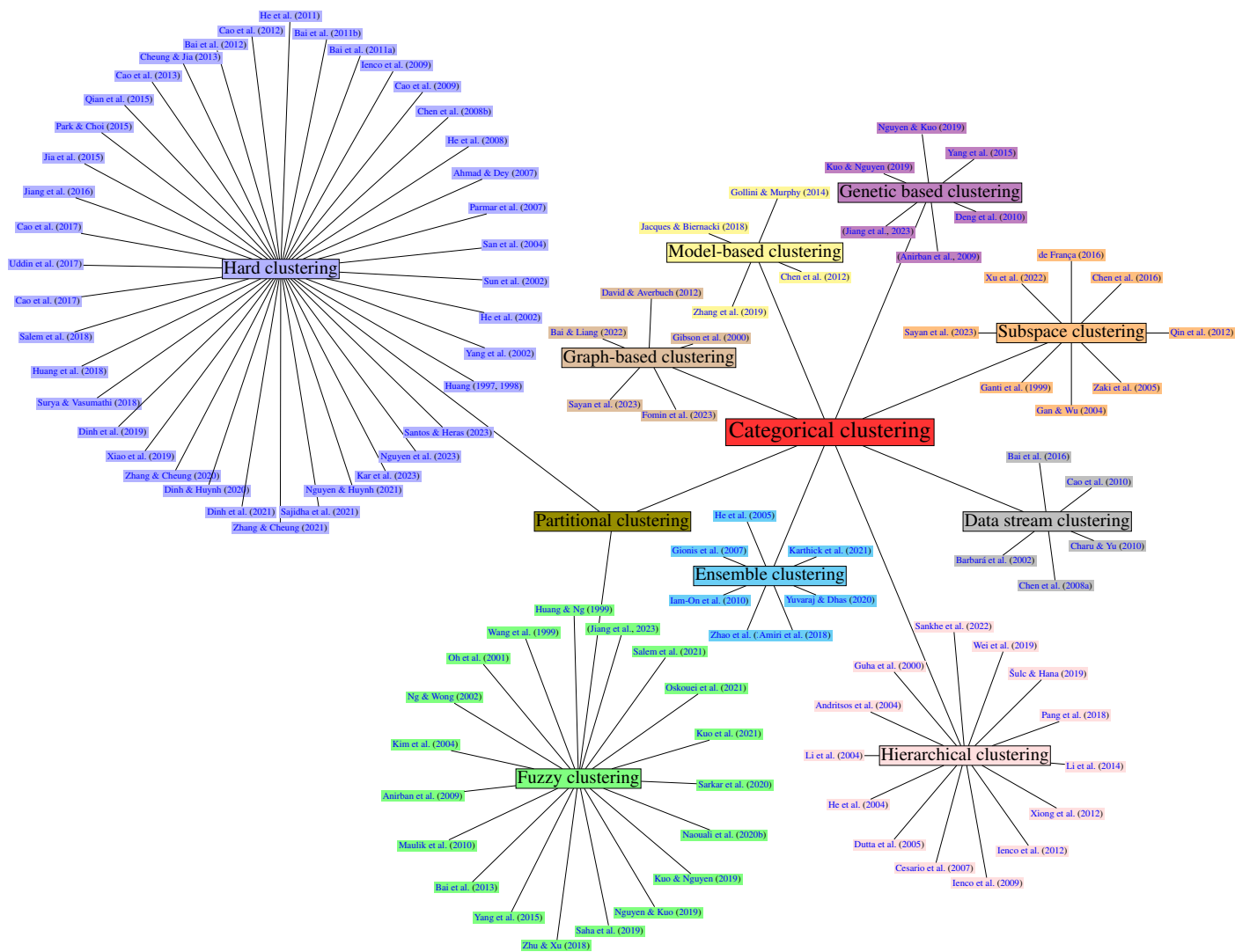
\begin{figure*}[!htb]
\vspace{-2cm}
    \centering
    \begin{adjustbox}{max width=\linewidth}
    \begin{tikzpicture}
    [grow cyclic,shape=rectangle,scale=1]
    \node[draw,minimum size=12pt] {\fontsize{50}{60}\selectfont Categorical clustering} 
     child [child_partitional]  {node[draw,minimum size=8pt] {\fontsize{40}{60}\selectfont Partitional clustering} 
        child [level distance=32cm, style=hard_clustering] {node[draw] {\fontsize{40}{60}\selectfont Hard clustering}
                child {node {\cite{huang1997fast,huang1998extensions}}}
                child {node {\cite{yang2002clope}}}
                child {node {\cite{he2002squeezer}}}
                child {node {\cite{sun2002iterative}}}
                child {node {\cite{san2004alternative}}}
                child {node {\cite{parmar2007mmr}}}
                child {node {\cite{ahmad2007method}}}
                child {node {\cite{he2008k}}}
                child {node {\cite{chen2008data}}}
                child {node {\cite{cao2009new}}}
                child {node {\cite{ienco2009context}}}
                child {node {\cite{bai2011initialization}}}
                child [level distance=18.5cm] {node {\cite{bai2011novel}}}
                child [level distance=19.5cm] {node {\cite{he2011attribute}}}
                child [level distance=18.5cm] {node {\cite{cao2012dissimilarity}}}
                child {node {\cite{bai2012cluster}}}
                child {node {\cite{cheung2013categorical}}}
                child {node {\cite{cao2013weighting}}}
                child {node {\cite{qian2015space}}}
                child {node {\cite{park2015rough}}}
                child {node {\cite{jia2015new}}}
                child {node {\cite{jiang2016initialization}}}
                child {node {\cite{cao2017algorithm}}}
                child {node {\cite{uddin2017empirical}}}
                child {node {\cite{cao2017algorithm}}}
                child {node {\cite{salem2018fast}}}
                child {node {\cite{huang2018new}}}
                child {node {\cite{surya2018attributes}}}
                child {node {\cite{dinh2019estimating}}}
                child {node {\cite{xiao2019optimal}}}
                child {node {\cite{zhang2020new}}}
                child {node {\cite{dinh2020k}}}
                child [level distance=19cm] {node {\cite{dinh2021clustering}}}
                child [level distance=20cm] {node {\cite{zhang2021learnable}}}
                child [level distance=19cm] {node {\cite{sajidha2021initial}}}
                child {node {\cite{mau2021lsh}}}
                child {node {\cite{kar2023efficient}}}
                child {node {\cite{nguyen2023method}}}
                child {node {\cite{santos2023fair}}}
        }
        child [level distance=20cm, style=fuzzy_clustering] {node[draw] {\fontsize{40}{60}\selectfont Fuzzy clustering}
                child [level distance=13cm] {node {\cite{huang1999fuzzy}}}
                child {node {\cite{wang1999clustering}}}
                child {node {\cite{oh2001fuzzy}}}
                child {node {\cite{ng2002clustering}}}
                child {node {\cite{kim2004fuzzy}}}
                child {node {\cite{mukhopadhyay2009multiobjective}}}
                child {node {\cite{maulik2010integrating}}}
                child {node {\cite{bai2013novel}}}
                child {node {\cite{yang2015non}}}
                child [level distance=13cm] {node {\cite{zhu2018many}}}
                child {node {\cite{saha2019integrated}}}
                child {node {\cite{nguyen2019partition}}}
                child {node {\cite{kuo2019genetic}}}
                child {node {\cite{naouali2020uncertainty}}}
                child {node {\cite{sarkar2020machine}}}
                child {node {\cite{kuo2021metaheuristic}}}
                child {node {\cite{oskouei2021fkmawcw}}}
                child {node {\cite{salem2021rough}}}
                child [level distance=12cm] {node {\citep{jiang2023kernel}}}
        }
    }
    child [level distance=12cm, style=child_ensemble]{node[draw,minimum size=8pt] {\fontsize{40}{60}\selectfont Ensemble clustering} 
       child {node {\cite{he2005cluster}}}
       child {node {\cite{gionis2007clustering}}}
       child {node {\cite{iam2010link}}}
       child {node {\cite{zhao2017clustering}}}
       child {node {\cite{amiri2018clustering}}}
       child {node {\cite{yuvaraj2020high}}}
       child {node {\cite{karthick2021ensemble}}}
    }
    child [level distance=27cm, style=child_hierarchical]  {node[draw,minimum size=8pt] {\fontsize{40}{60}\selectfont Hierarchical clustering} 
         child {node {\cite{guha2000rock}}} 
         child {node {\cite{andritsos2004limbo}}}
         child [level distance=10cm] {node {\cite{li2004entropy}}}
         child {node {\cite{he2004organizing}}}
         child {node {\cite{dutta2005qrock}}}
         child {node {\cite{cesario2007top}}}
         child [level distance=10cm] {node {\cite{ienco2009context}}}
         child {node {\cite{ienco2012context}}}
         child {node {\cite{xiong2012dhcc}}}
         child [level distance=10cm] {node {\cite{li2014hierarchical}}}
         child {node {\cite{pang2018parallel}}}
         child {node {\cite{vsulc2019comparison}}}
         child {node {\cite{wei2019hierarchical}}}
         child [level distance=10cm] {node {\cite{sankhe2022mutual}}}
    }
    child [child_data_stream] {node[draw,minimum size=8pt] {\fontsize{40}{60}\selectfont Data stream clustering} 
         child {node {\cite{barbara2002coolcat}}}
         child {node {\cite{chen2008catching}}}
         child {node {\cite{aggarwal2010clustering}}}
         child {node {\cite{cao2010framework}}}
         child {node {\cite{bai2016optimization}}}
    }
    child [level distance=20cm, style=child_subspace] {node[draw,minimum size=8pt] {\fontsize{40}{60}\selectfont Subspace clustering} 
         child {node {\cite{ganti1999cactus}}}
         child {node {\cite{gan2004subspace}}}
         child {node {\cite{zaki2005clicks}}}
         child [level distance=12cm] {node {\cite{qin2012novel}}}
         child {node {\cite{chen2016soft}}}
         child {node {\cite{de2016hash}}}
         child {node {\cite{xu2022multi}}}
         child [level distance=10cm] {node {\cite{bandyapadhyay2023parameterized}}}
    }
    child [level distance=23cm, style=child_genetic] {node[draw,minimum size=8pt] {\fontsize{40}{60}\selectfont Genetic based clustering} 
         child {node {\citep{mukhopadhyay2009multiobjective}}}
         child {node {\cite{deng2010g}}}
         child {node {\cite{yang2015non}}}
         child[level distance=4.3cm] {node {\cite{nguyen2019partition}}}
         child[level distance=6.2cm] {node {\cite{kuo2019genetic}}}
         child[level distance=11cm]  {node {\cite{kuo2021metaheuristic}}}
         child {node {\citep{jiang2023kernel}}}
    }
    child [level distance=15cm, style=child_model] {node[draw,minimum size=8pt] {\fontsize{40}{60}\selectfont Model-based clustering} 
         child {node {\cite{chen2012model}}}
         child {node {\cite{gollini2014mixture}}}
         child {node {\cite{jacques2018model}}}
         child {node {\cite{zhang2019unified}}}
    }
    child [level distance=15cm, style=child_graph] {node[draw,minimum size=8pt] {\fontsize{40}{60}\selectfont Graph-based clustering} 
         child {node {\cite{gibson2000clustering}}}
         child {node {\cite{david2012spectralcat}}}
         child {node {\cite{bai2022categorical}}}
         child {node {\cite{bandyapadhyay2023parameterized}}}
         child {node {\cite{fomin2023parameterized}}}
    }
    ;
    \end{tikzpicture}
    \end{adjustbox}
    \caption{A taxonomy of clustering algorithms }
    \label{fig:tree_taxonomy}
\end{figure*}

\textsc{In 2001}, \cite{oh2001fuzzy} introduced the \textsc{Fccm} algorithm, which maximizes cluster aggregation rather than minimizing distance. It achieves this by defining memberships for individuals described by qualitative variables with multiple categories and optimizing an objective function representing aggregation. The function includes terms for entropy maximization to obtain fuzzy clusters and constraints for memberships. The algorithm iterates, starting with random initial memberships, then updating memberships for categories and individuals based on necessary conditions for local minima. Iteration continues until membership changes fall below a specified threshold, indicating convergence.

\textsc{In 2002}, \hypertarget{coolcat}{\cite{barbara2002coolcat}} introduced an incremental entropy-based algorithm named \textsc{Coolcat}. The process is based on two main steps: initialization and incremental clustering. In the initialization step, an initial set of \emph{k} clusters are found from a small sample of the data by selecting the \emph{k} most dissimilar data points that maximize the minimum pairwise entropy. The remaining data points are then clustered incrementally by computing the expected entropy of placing each data point in the existing clusters and assigning it to the cluster that minimizes the overall expected entropy. This incremental process allows \textsc{Coolcat} to efficiently cluster large datasets and data streams without having to reprocess the entire dataset for each new data point. 
\hypertarget{clope}{\cite{yang2002clope}} introduced the \textsc{Clope} algorithm for clustering transactional data, such as market basket data and web logs. The algorithm operates by iteratively scanning the dataset and assigning each transaction to a cluster that maximizes a global criterion function called \emph{profit}. This \emph{profit} function aims to increase the height-to-width ratio (or gradient) of the cluster histogram, which geometrically represents the density and overlap of items within each cluster. In this context, ``overlap" refers to grouping items that frequently co-occur together in transactions within the same cluster. In the initialization phase, \textsc{Clope} scans the database once and assigns each transaction to an initial cluster that maximizes the \emph{profit} function. In the iteration phase, it repeatedly scans the database and moves transactions to different clusters if it increases the overall profit, until convergence. 
\hypertarget{squeezer}{\cite{he2002squeezer}} introduced the \textsc{Squeezer} algorithm for clustering categorical data. It works by reading tuples from the dataset one by one. The first tuple forms a cluster by itself. For each subsequent tuple, the algorithm computes its similarity to all existing clusters using a \emph{statistics-based similarity} function. If the maximum similarity exceeds a given threshold, the tuple is assigned to the most similar cluster. Otherwise, the tuple forms a new cluster by itself. 
\cite{sun2002iterative} adapted an iterative refinement process to improve the selection of initial points for the \textsc{K-modes} algorithm. This process involves creating multiple small subsamples from the data, clustering each subsample to identify initial cluster modes, and then refining these modes through further clustering. The set of modes that minimizes a distortion measure (which evaluates how well the initial points represent the overall dataset) compared to the original cluster modes is selected as the refined initial points for the final clustering of the full data.
\cite{ng2002clustering} incorporated \emph{tabu search} into the \textsc{Fuzzy K-modes} algorithm to avoid local minima and aim for a global optimal solution. The \emph{tabu search} technique is a meta-heuristic that guides a local heuristic search procedure to explore the solution space beyond local optimality. The \textsc{Tabu Search Based Fuzzy K-modes} uses several parameters to control the search process, such as tabu list size, probability threshold, number of trial solutions, and a reduction factor for the maximum number of non-improving moves.

\textsc{In 2004}, \hypertarget{limbo}{\cite{andritsos2004limbo}} introduced a scalable algorithm named \textsc{Limbo}. The algorithm works by transforming data tuples into \emph{Distributional Cluster Features} (DCF) for summarization purposes. DCFs are then inserted into a height-balanced tree known as the \emph{DCF tree}. In the clustering phase, the \emph{Agglomerative Information Bottleneck} (AIB) algorithm is applied to cluster the DCFs in the leaf nodes into a desired number of \emph{k} clusters. Finally, each data tuple is assigned to the cluster where its DCF is closest to the representative DCF of the selected clusters.
\cite{san2004alternative} introduced the \textsc{K-representatives} that defines \emph{representatives} for each cluster using the relative frequencies of categories within that cluster. It defines the dissimilarity between each object and the representative by the multiplication of relative frequencies of categories within the cluster and the simple matching measure between categories. \textsc{K-representatives} iteratively reassigns objects to the closest representative’s cluster until convergence, aiming to minimize the overall sum of dissimilarities between objects and representatives.
\cite{li2004entropy} applied the \emph{entropy measure} for categorical clustering. Initially, all data points are placed in a single cluster. The algorithm then minimizes the \emph{entropy-based criterion} which is the expected entropy of the clusters. A randomized \emph{Monte Carlo} procedure is used to iteratively refine the cluster assignments until the entropy criterion cannot be further minimized.
\cite{kim2004fuzzy} introduced fuzzy centroids to represent the uncertainty and imprecision in cluster centroids for categorical data, which helps in achieving better clustering performance, particularly for boundary data points. The algorithm is iterative, updating membership degrees and centroids until convergence.
\cite{he2004organizing} introduced the \textsc{MD\_{hac}} algorithm that considers query schemas as categorical data points and uses a multinomial model and statistical hypothesis testing approach tailored for this type of data. It adopts hierarchical agglomerative clustering but defines a \emph{model-differentiation} objective function to maximize the difference between the multinomial models representing each cluster.
\cite{gan2004subspace} introduced the \textsc{SUBCAD} algorithm for clustering high-dimensional categorical datasets, emphasizing the optimization of objective functions to determine both clusters and their associated subspaces.
 
\textsc{In 2005}, \cite{he2005cluster} introduced the \textsc{CCDByEnsemble} algorithm that treats each categorical attribute as a separate clustering of the data, where the values of that attribute represent the cluster labels. It defines an objective function based on \emph{average normalized mutual information} to maximize agreement between the final clustering and the clusterings given by each categorical attribute. It runs three existing cluster ensemble algorithms, namely CSPA, HGPA, and MCLA, and selects the one with the highest mutual information score as the final clustering result.
\cite{zaki2005clicks} introduced the \textsc{Clicks} algorithm that models a categorical dataset as a \emph{k-partite graph}, where vertices represent attribute values and edges connect dense pairs of values from different attributes. It then maps the clustering problem to finding \emph{maximal k-partite cliques} in this graph, where \emph{maximal} refers to cliques that cannot be extended by including any additional vertices without breaking the clique property. These maximal cliques correspond to the largest possible subspace clusters in the data, representing groups of attribute values that frequently co-occur.
\cite{dutta2005qrock} introduces the \textsc{Qrock} algorithm, an efficient version of the \textsc{Rock} algorithm. \textsc{Qrock} bypasses the time-consuming explicit calculation of links and directly merges components based on neighbor lists. Experimental results show that \textsc{Qrock} significantly reduces computation time while preserving the clustering quality of \textsc{Rock}.

\textsc{In 2007}, \hypertarget{aggregation}{\cite{gionis2007clustering}} introduced various algorithms for clustering aggregation, aiming to derive a single clustering that aligns closely with a given set of clusterings. Among these algorithms, \emph{BestClustering} minimizes total disagreement with other input clusterings, \emph{Balls} operates greedily on the principle of correlation clustering (where the objective is to partition data such that the number of disagreements, or edges representing pairs of objects frequently separated in input clusterings, is minimized), \emph{Agglomerative} is a bottom-up approach, \emph{Furthest} starts with all nodes in one cluster and iteratively separates them, \emph{LocalSearch} employs a heuristic to iteratively refine clusters, and \emph{Sampling} clusters a random sample of nodes and assigns the rest accordingly.

\cite{parmar2007mmr} introduced the \textsc{Min-Min-Roughness} algorithm that clusters categorical data by evaluating the \emph{mean roughness} of attributes to determine how effectively they partition the data. It selects the attribute with the lowest \emph{mean roughness} as the splitting attribute and recursively divides the data into two subsets based on its values, forming binary partitions. This process continues until the desired number of clusters is attained. Objects are then assigned to clusters based on their membership degrees in lower and upper approximations, accommodating uncertainty in clustering.
\cite{ahmad2007method} proposed a method to compute the distance between two categorical values of the same attribute
by considering their relationships with other attributes in the dataset. The maximum co-occurrence probability difference over subsets quantifies the distance between two values. Averaging over attributes gives the overall distance between objects. The proposed distance metric was tested on the \textsc{K-modes} algorithm, showing improved clustering performance compared to the original method.

\subsubsection{From 2008 to 2015}
\textsc{In 2008}, \cite{he2008k} introduced a \emph{K-means-like} clustering algorithm  for categorical data named \textsc{K-anmi}. It iteratively optimizes the \emph{average normalized mutual information} objective function with the input parameter \emph{k} specifying the desired number of clusters.
\cite{chen2008data} introduced the \emph{MAximal Resemblance Data Labeling} (\textsc{Mardl}) algorithm. Initially, a sample of data points is taken from the full dataset and labeled through this clustering process. \textsc{Mardl} then labels the remaining unlabeled data points by creating an NNIR \emph{(N-Nodeset Importance Representative)} for each cluster. This NNIR captures key attribute combinations from the labeled sample and uses the maximal combination of \emph{n-nodesets} and their importance values in \emph{NNIR trees} to assign labels.

\textsc{In 2009}, \cite{cao2009new} proposed a new initialization technique to enhance the \textsc{K-modes} and \textsc{Fuzzy K-modes} algorithms. The technique called \textsc{Cao's method} computes the average density for each object by comparing its attribute values to those of other objects in the dataset. The object with the highest average density becomes the first cluster center, representing a dense region in the data space. Subsequent centers are chosen based on both their distance from existing centers and their densities.
\cite{mukhopadhyay2009multiobjective} introduced a \emph{multiobjective genetic clustering} method encoding cluster medoids as integer strings. Each gene represents a data point index as a medoid, and clusters are formed by assigning points to their nearest medoids. The algorithm uses \textsc{Non-dominated Sorting Genetic Algorithm II} (\textsc{Nsga-ii}), a well-known evolutionary algorithm for multiobjective optimization, for selection, crossover, mutation, and elitism, which retains the best solutions across generations. \textsc{Nsga-ii} optimizes both \textsc{K-medoids} error and \textsc{Silhouette} index simultaneously.

\textsc{In 2010}, \cite{iam2010link} introduced a \emph{link-based cluster ensemble} (\textsc{LCE}) algorithm. It creates an ensemble of base clusterings from categorical data using various techniques such as random initializations and subspace projections. It then builds a \emph{refined cluster-association matrix} (RM) to capture associations between data points and clusters. The innovation lies in estimating unknown associations in the RM using a \emph{link-based similarity} measure between clusters. Finally, it applies \emph{spectral graph partitioning} on a \emph{weighted bipartite graph} derived from the RM to produce the final data clustering. \cite{aggarwal2010clustering} introduced \textsc{ConStream} for categorical data streams. The algorithm maintains compact \emph{cluster droplets} that summarize data points in real time. New data points are assigned to the closest existing droplet or create a new one if none are similar enough. Droplets use an additive representation for efficient updates and decay over time to adapt to evolving patterns in the stream. Periodic snapshots allow cluster retrieval over time without reprocessing the entire stream.
 
\textsc{In 2011}, \cite{bai2011novel} introduced the \textsc{Mwkm} algorithm for clustering high-dimensional categorical data. It automatically identifies relevant subspaces through attribute weights for different clusters. The algorithm minimizes a mixed objective function combining within-cluster dispersion and attribute weights. It iteratively updates cluster assignments, centers, and two sets of attribute weights ($L$ for differences from cluster centers and $S$ for matches) using derived rules. It proposes a new weighted dissimilarity measure assigning weights based on object-cluster center matches or differences for each attribute.
\cite{bai2011initialization} proposed an initialization method that selects representative data points (\emph{exemplars}) from each subset of data based on categorical attributes. The algorithm measures each \emph{exemplar}'s density and its cohesiveness with surrounding data points. It then iteratively selects the densest \emph{exemplars} as initial cluster centers, ensuring that each new center is distant from the others. The number of clusters is determined by observing significant changes in possibility values, which indicate the plausibility of cluster numbers by combining each \emph{exemplar}'s density with its distance from previously selected centers.

\textsc{In 2012}, \cite{cao2012dissimilarity} designed a dissimilarity measure based on the idea of biological and genetic taxonomy, where uncommon attribute value matches between objects are considered more significant for determining similarity/dissimilarity. It uses the rough membership function from rough set theory to quantify the degree of overlap between two data points for a given attribute. This allows \textsc{K-modes} to better identify characteristic clusters, which are tight and cohesive groups of objects sharing rare or unique attribute values.
\cite{ienco2012context} introduced the \emph{DIstance Learning for Categorical Attributes} (\textsc{DILCA}) algorithm, which uses symmetric uncertainty, derived from \emph{entropy} and \emph{information gain}, to select \emph{relevant context} attributes for each target attribute, meaning any specific categorical attribute whose value distances need to be computed. The algorithm calculates distances between attribute values by analyzing their \emph{co-occurrence patterns} within this context. To compute distances between data points, these learned distance matrices are then integrated into a standard distance measure, like \emph{Euclidean} distance. These distances are used in \emph{Ward's agglomerative hierarchical clustering} to form the final clusters.
\cite{chen2012model} introduced the \emph{Expansion, Adjustment, Simplification until Termination} {\textsc{EAST}) algorithm that performs multidimensional clustering by learning a \emph{latent tree model} from the data. It begins with an initial model and iteratively refines it by adding, removing, or relocating latent variables and their connections to observed variables. At each step, it uses a local \emph{Expectation-Maximization} algorithm and a model score like \emph{Bayesian Information Criterion} (BIC) to evaluate and improve model fit while controlling complexity. 

\textsc{In 2013}, \cite{khan2013cluster} proposed a cluster center initialization algorithm for \emph{K-modes}. It performs multiple clusterings based on the values of different data attributes, focusing on \emph{Prominent} attributes, which have distinct values less than or equal to the desired number of clusters and higher discriminatory power. It compares this with another method that selects \emph{Significant} attributes based on their distribution and co-occurrence. The algorithm creates multiple clustering views, captures them as cluster strings, and uses these to compute initial cluster centers. 
\cite{cheung2013categorical} designed a unified similarity metric for mixed datasets, but it can work directly on a categorical dataset. The similarity between objects and clusters is determined by a weighted summation considering attribute value distributions within the cluster. This similarity is defined probabilistically, based on the number of objects in the cluster sharing the same attribute value. In purely categorical data, numerical vectors are absent from the similarity metric, and similarity is solely based on categorical attributes. Attribute importance is measured by average entropy over each attribute value, determining attribute weights accordingly.

\textsc{In 2014}, \cite{gollini2014mixture} introduced the \emph{Mixture of Latent Trait Analyzers} (\textsc{Mlta}) model. It enhances traditional \emph{latent class analysis} by using a \emph{categorical latent variable} for group structure and a \emph{continuous latent trait} for within-group dependencies. A variational approximation to the likelihood function is employed to handle the complexity of the model. The authors propose a variational \emph{Expectation-Maximization} algorithm for efficient model fitting, shown to be effective for high-dimensional data and complex latent traits.
\cite{li2014hierarchical} introduced a top-down hierarchical clustering algorithm named \textsc{Mtmdp}. The algorithm starts with the entire dataset as the root node. To determine the splitting attribute, it calculates the \emph{Total Mean Distribution Precision} (TMDP) value for each attribute. This value measures the distribution precision of the equivalence classes--subsets of objects that are indistinguishable from each other based on a given set of attributes--induced by that attribute across all other attributes. The attribute with the highest TMDP value is selected as the splitting attribute, and a binary split is performed, creating two child nodes by dividing the data points based on the values of the selected attribute. The process continues recursively on the child node with the smallest cohesion degree until the desired number of clusters is reached.

\textsc{In 2015}, \cite{jia2015new} introduced \textsc{DM3}, a distance metric for categorical data. It employs a frequency-based measure, dynamic weighting for infrequent attribute pairs, and accounts for attribute interdependence. \textsc{DM3} enhances clustering performance in \textsc{K-modes} relative to other distance metrics such as Hamming and DILCA distance metrics.
\cite{qian2015space} introduced the \emph{Space Structure-Based Clustering} (\textsc{Sbc}) algorithm that transforms categorical data into a \emph{Euclidean} space, where each categorical feature becomes a new dimension. It calculates the probability that any two objects are equal based on their categorical values, resulting in a vector of probabilities for each object. A dissimilarity measure, such as cosine or Euclidean distance, is then used to cluster the objects using a \textsc{K-means} style iterative approach.
\cite{yang2015non} introduced a multi-objective genetic algorithm named \textsc{Nsga-fmc}. It uses fuzzy membership values as chromosomes, represented by a matrix indicating the probability of assigning data points to clusters. The algorithm optimizes two objectives: minimizing intra-cluster distances and maximizing inter-cluster distances. It applies genetic operators like selection, crossover, and mutation to the fuzzy membership chromosomes.

\subsubsection{From 2016 to 2023}
\textsc{In 2016}, \cite{jiang2016initialization} proposed two initialization algorithms for \textsc{K-modes} called \textsc{Ini\_Distance} and \textsc{Ini\_Entropy}. The \textsc{Ini\_Distance} calculates the \emph{distance outlier factor} (DOF) for each data point to measure its outlierness. It selects initial centers based on two factors: lower DOF and greater distances to other cluster centers, iteratively choosing the $k$ data points with the highest possibilities of being good initial centers. \textsc{Ini\_Entropy} works similarly but uses a \emph{partition entropy-based outlier factor} (PEOF) instead of DOF. PEOF measures a data point's contribution to the entropy of data partitions. Initial center selection is based on PEOF and distances to other cluster centers.
\cite{chen2016soft} introduced the \emph{Subspace Clustering of Categories} (\textsc{Scc}) algorithm, which clusters categorical data by minimizing probabilistic distances between data points and clusters. Each object is a probability distribution, with \emph{Euclidean} distance used for measurements. \textsc{Scc} weights attributes based on the kernel-smoothed dispersion of categories, giving higher weights to attributes with lower dispersion. The algorithm iteratively updates cluster assignments, attribute weights, and bandwidth parameters to minimize the weighted sum of distances. It also introduces a new validity index to estimate the optimal number of clusters.

\textsc{In 2017}, \cite{zhao2017clustering} introduced the \emph{Sum of Internal Validity Indices with Diversity} (\textsc{Sivid}), a method to select high-quality, diverse base clusterings from an ensemble in three stages. First, five internal clustering validity indices including \emph{category utility, silhouette, Davies-Bouldin index, cluster cardinality index, \textit{and} information entropy} are calculated for each base clustering to assess quality. Next, diversity between base clusterings is measured using \emph{normalized mutual information}. Finally, an iterative process selects the base clustering with the highest initial quality sum, then adds clusterings that maximize both quality and diversity until the desired number of clusters is reached. 
\cite{cao2017algorithm} introduced the \textsc{SV-K-modes} algorithm, which clusters data by representing each cluster with a \emph{set-valued mode} that minimizes the sum of distances to all objects in that cluster using a \emph{Jaccard coefficient-based distance} measure. The algorithm iteratively assigns objects to clusters based on their distance to these \emph{modes} and updates the \emph{modes} by selecting the most frequent values for each feature. It also employs a heuristic method called HAFSM to update the cluster centers and uses a strategy named GICCA to select initial \emph{modes} by choosing data points with higher density and large mutual distances.

\textsc{In 2018}, \cite{salem2018fast} introduced the \emph{Manhattan Frequency K-means} (\textsc{MFk-M}) algorithm that converts categorical data into numeric values based on category frequency within attributes, enabling the use of numeric clustering algorithms like \textsc{K-means}. Observations are assigned to the nearest cluster centroid based on the \emph{Manhattan} distance, and centroids are updated by averaging the frequencies in each cluster. This process repeats until convergence or a stopping criterion is met.
\cite{zhu2018many} introduced the \emph{Many-objective Optimization based Fuzzy Centroids} (\textsc{MaOFcentroids}) algorithm, which employs \textsc{NSGA-III} (Non-dominated Sorting Genetic Algorithm III), an evolutionary algorithm designed for many-objective optimization that maintains a diverse set of solutions across multiple objectives. The algorithm combines stochastic global search through genetic operations like crossover and mutation with local search via one-step fuzzy centroids clustering to generate new solutions. A clustering ensemble approach integrates selected non-dominated solutions based on their ranking across different validity indices for the final clustering.

\textsc{In 2019}, \cite{dinh2019estimating} introduced the \textsc{K-scc} algorithm to estimate the optimal number of clusters ($k$) for categorical data clustering by utilizing the \textsc{Silhouette Coefficient} as a measure of clustering quality. The algorithm iterates over potential cluster numbers, performing clustering for each candidate $k$ using \emph{kernel density estimation} for cluster centers and an \emph{information theoretic-based dissimilarity} measure for distances. \textsc{K-scc} selects the number of clusters with the highest \emph{average silhouette value} for the best clustering quality.
\cite{vsulc2019comparison} proposed two \emph{variability-based similarity measures} for categorical data in hierarchical clustering. \textsc{Variable Entropy} (\textsc{Ve}) expresses similarity between categories using the entropy formula, giving more weight to matches in variables with high variability (high entropy). \textsc{Variable Mutability} (\textsc{Vm}) expresses similarity using the Gini coefficient/mutability formula, also giving more weight to matches in variables with high variability (high mutability). Both assess object similarity as the average across variables and inform proximity matrices for hierarchical clustering.

\textsc{In 2020}, \cite{dinh2020k} introduced the \textsc{K-pbc} algorithm, which converts categorical datasets into transaction datasets and utilizes \emph{maximal frequent itemset mining} (\textsc{Mfim}) to identify the top $k$ \emph{maximal frequent itemsets} (MFIs). These MFIs represent groups of categorical objects sharing the most frequently co-occurring categories. After merging objects corresponding to the top $k$ MFIs and eliminating overlaps, \textsc{K-pbc} forms $k$ initial clusters. In the clustering phase, objects are iteratively assigned to the nearest cluster center and the centers are updated using a \emph{kernel-based method}. The algorithm employs an \emph{information theoretic-based dissimilarity measure} to compute distances between objects and cluster centers, iterating these steps until convergence.
\cite{zhang2020new} introduced the \emph{Unified Distance Metric} (\textsc{Udm}) for clustering categorical data with both ordinal and nominal attributes. It quantifies category distances using an \emph{entropy-based approach}, measuring interdependence among attributes. The final distance between two data points is computed by weighting the distances measured on each attribute according to the interdependence between that attribute and the others. \textsc{Udm} has been integrated into various clustering algorithms such as \textsc{K-modes} \citep{huang1998extensions}, \textsc{Mwkm} \citep{bai2011novel}, \textsc{Scc} \citep{chen2016soft}, and entropy-based clustering \citep{li2004entropy}.

\textsc{In 2021}, \cite{dinh2021clustering} introduced the \textsc{K-ccm*} algorithm, which integrates imputation for missing categorical values with clustering. For the imputation step, it uses a decision tree approach to find the set of correlated complete objects for each incomplete object with missing values. For the clustering step, it represents cluster centers using the \emph{kernel density estimation method}, and use the \emph{information theoretic-based dissimilarity measure} to compute distances between objects and cluster centers. The imputation and clustering phases are performed iteratively until convergence.
\cite{kuo2021metaheuristic} introduced \textsc{Pfkm}, which combines \textsc{Fkm} with possibility theory to improve categorical data clustering by addressing outliers. It employs a typicality matrix alongside fuzzy membership to handle outliers. Additionally, three meta-heuristic algorithms: \emph{genetic algorithm} (\textsc{Ga}), \emph{particle swarm optimization} (\textsc{Pso}), and \emph{sine cosine} algorithm (\textsc{Sca}) are combined with \textsc{Pfkm} to optimize initial centroids and algorithm parameters. These meta-heuristic-based \textsc{Pfkm} algorithms (\textsc{Ga-Pfkm}, \textsc{Pso-Pfkm} and \textsc{Sca-Pfkm}) iteratively update centroids, fuzzy memberships, and typicality values to minimize an objective function that includes fuzzy membership, typicality, and a frequency probability-based distance measure for categorical data.

\textsc{In 2022}, \cite{bai2022categorical} introduced the \textsc{Cdc\_dr} algorithm that constructs a graph where nodes represent categorical values, with edge weights based on set-similarity measures such as the \emph{Jaccard coefficient}. Graph embedding techniques, such as \emph{spectral embedding} or \emph{non-negative matrix factorization} are used to create \emph{low-dimensional vector} representations for each node, ensuring similar values have similar vectors. For each data instance, these vectors are combined to form a numerical representation. Finally, a clustering algorithm such as  \textsc{K-means} is applied to these numerical representations to cluster the original categorical data.
\cite{xu2022multi} introduced \emph{Multi-view Kernel Clustering framework for Categorical sequences} (\textsc{Mkcc}) that integrates information from multiple views. It constructs \emph{kernel matrices} that represent the relationships between data points in each view without the need to predefine a kernel function. This is achieved through self-representation learning, where data points are expressed in terms of other data points, capturing the inherent structure of the data. \textsc{Mkcc} also uses an approximation method to reduce the storage and computation requirements. Moreover, it includes a sample weighting mechanism to reduce the influence of noise and outliers on the clustering results.

\textsc{In 2023}, \cite{nguyen2023method}\footnote{This paper first appeared online in 2019} introduced an extension of the \textsc{K-means} for categorical data. It incorporates an \emph{information theoretic-based dissimilarity measure} assesses the commonality between categorical values based on the amount of information they share. For representing cluster centers, the algorithm employs a \emph{kernel density estimation method} that provides a statistical interpretation consistent with numerical data’s cluster means. Each cluster center is represented as a \emph{probability distribution} over the categorical values in that cluster. These modifications allow the clustering problem for categorical data to be approached similarly to \textsc{K-means}, iteratively assigning objects to the nearest cluster centers and updating the centers based on new memberships.
\cite{kar2023efficient} introduced an \emph{entropy-based dissimilarity} measure for clustering categorical data, utilizing Boltzmann’s entropy definition to evaluate the disorder within dataset attributes. The algorithm calculates and normalizes the entropy for each categorical attribute, assigning higher weights to those with greater entropy to signify their importance in clustering. The dissimilarity between data points is calculated by comparing their attribute values: different values add the attribute's weight to the dissimilarity score, while identical values contribute nothing. The final dissimilarity score is a weighted sum of these comparisons across all attributes. This proposed dissimilarity measure was tested on \textsc{K-modes}.
\subsection{Top 10 most cited algorithms}
Table \ref{tab:topten_algorithms} shows the 10 most cited algorithms for clustering categorical data from the past 25 years, all of which have significantly influenced the field. These algorithms, as introduced in the section \ref{sec:historical_dev} have collectively shaped the landscape of categorical data clustering, providing novel and robust solutions for various applications and influencing numerous subsequent studies in numerous fields and contexts.

We can observe that partitional and hierarchical clustering remain the two most common clustering techniques for categorical data, as evidenced by the prominence of algorithms such as \textsc{K-modes} and \textsc{Rock}. The main advantages of \textsc{K-modes} are its ease of implementation and extreme efficiency in dealing with large-scale datasets. On the other hand, \textsc{Rock} exemplifies the utility of hierarchical clustering for this type of data analysis, by employing a link-based approach to measure similarity between data points, which is particularly advantageous for handling datasets without specifying the number of clusters in advance. The sustained relevance and high citation counts of these algorithms underscore the robustness and applicability of partitional and hierarchical clustering methods in the domain of categorical exploratory data analysis.

\begin{table}[!htb]
\centering
\caption{Top ten most cited clustering algorithm }
\begin{adjustbox}{max width=\linewidth}
\begin{tabular}{cll}\toprule
\bf{Year}   & \bf{Algorithm}  &\bf{\#Citations} \\ \midrule \rowcolor[gray]{.95}              
{1998} & {\hyperlink{k-modes}{\textsc{\color{red}K-modes}} \citep{huang1997fast,huang1998extensions}}  &  {4697} \\ 
{2000}&{\hyperlink{rock}{\textsc{\color{red}Rock}} \citep{guha2000rock}}&{2954}  \\ \rowcolor[gray]{.95}
{2007}&{\hyperlink{aggregation}{\textsc{\color{red}Aggregation Clustering}} \citep{gionis2007clustering}}&{1213} \\ 
{1999}&{\hyperlink{cactus}{\textsc{\color{red}Cactus}} \citep{ganti1999cactus}}&{736}  \\ \rowcolor[gray]{.95}
{1999}&{\hyperlink{fuzzy_kmodes}{\textsc{\color{red}Fuzzy K-modes}} \citep{huang1999fuzzy}}&{666} \\ 
{2000}&{\hyperlink{stirr}{\textsc{\color{red}Stirr}} \citep{gibson2000clustering}}&{620}  \\ \rowcolor[gray]{.95}
{2002}&{\hyperlink{coolcat}{\textsc{\color{red}Coolcat}} \citep{barbara2002coolcat}}&{561} \\ 
{2004}&{\hyperlink{limbo}{\textsc{\color{red}Limbo}} \citep{andritsos2004limbo}}&{351}  \\ \rowcolor[gray]{.95}
{2002}&{\hyperlink{clope}{\textsc{\color{red}Clope}} \citep{yang2002clope}}&{321} \\ 
{2002}&{\hyperlink{squeezer}{\textsc{\color{red}Squeezer}} \citep{he2002squeezer}}&{302} \\ \bottomrule
\end{tabular}
\label{tab:topten_algorithms}
\end{adjustbox}
\end{table}

\newgeometry{bottom=1.1cm, left=0.5cm}
\begin{table*}[!htb]
\vspace{-3cm}
\caption{Comparison of \textsc{Hard Partitional Clustering Algorithms}}
\label{tab:partitional_clustering_hard_1}
\begin{adjustbox}{max width = 1.1\linewidth}
\renewcommand{\arraystretch}{2.5}
\setlength{\arrayrulewidth}{0.001pt}
\begin{tabular}{|p{1cm}|p{3.6cm}|p{3.6cm}|p{3.1cm}|p{1.9cm}|p{2.3cm}|p{10cm}|}
\hline

\textbf{Ref.} & \textbf{Sim/Dissim measure} & \textbf{Complexity} & \textbf{Evaluation metrics} & \textbf{Data sources} & \textbf{Plots used} & \textbf{Remarks}\\\hline

\cite{huang1997fast, huang1998extensions} & Simple matching & \textbf{$\mathcal{O}(tknm)$}, where $t$: \#iterations, $n$: \#instances, $m$: \#attributes, $k$: \#clusters & Accuracy, percentage of misclassification, runtime, scalability & UCI & Line plot, misclassification matrix, table & \textsc{K-modes} extends \textsc{K-means} for categorical data by using \emph{simple matching}, replacing \emph{means} of clusters with \emph{modes}, and by using a \emph{frequency-based method} to update modes during the clustering process.\ \\\hline

\cite{yang2002clope} & Height-to-width ratio of cluster histograms & \textbf{$\mathcal{O}(NKA)$}, where $N$: \#transactions, $K$: \#clusters, $A$: average length of a transaction  & Purity, runtime, scalability & UCI & Bar plot, line plot, table & \textsc{Clope} scans a transactional database iteratively, assigning transactions to clusters to maximize a global criterion function. This function is based on the height-to-width ratio of cluster histograms, reflecting intra-cluster item similarity. \\\hline

\cite{he2002squeezer} &Similarity between cluster and each transaction tuple $ \text{Sim}(C, \text{tid})$ & Time: $\mathcal{O}(nkpm)$, space: $\mathcal{O}(n+kpm)$  where $n$: \#instances, $k$: \#clusters, $m$: \#attributes, $p$: \#distinct categories per attribute& Percentage of misclassified tuples, runtime, scalability & UCI \newline Synthetic & Line plot, table & The single-pass approach allows \textsc{Squeezer} to achieve high efficiency and scalability for large datasets, while still producing high-quality clustering results. \\\hline
 
\cite{sun2002iterative} & Simple matching & \textbf{$\mathcal{O}\left( JK(\| S_j \|) \right)$}, where $J$: \#sub samples, $K(\| S_j \|)$: cost for clustering $\| S_j \|$ points to $K$ clusters. & Accuracy rate, error rate & UCI & Scatter plot, table & The refinement process aims to derive initial points that better approximate the true underlying cluster modes, leading to more accurate and stable clustering results compared to using random initial points directly. \\\hline

\cite{san2004alternative} & Relative frequencies and simple matching of categorical values & Not discussed  & Accuracy, clustering error & UCI & Misclassification matrix, table & \textsc{K-representatives} uses \emph{fuzzy representatives} based on category frequencies instead of \emph{modes}, and a dissimilarity measure that accounts for all possible modes via the representatives. This avoids the instability issue in \textsc{K-modes}. \\\hline

\cite{parmar2007mmr} & Roughness measure based on rough set theory & \textbf{$\mathcal{O}(knm$+$kn^2l)$}, where $k$: \#clusters, $n$: \#attributes, $m$: \#instances, $l$: maximum \#values in the attribute domains& Purity & UCI & Table & \textsc{Mmr} is a rough set theory-based hard clustering algorithm. Unlike fuzzy clustering, it does not use fuzzy membership functions, but it accommodates varying degrees of cluster membership, making it adept at handling clustering uncertainty. \\\hline

\cite{ahmad2007method} & Co-occurrence probabilities of attribute values with respect to other attributes in the dataset & $\mathcal{O}(m^2n+R^3m^2)$, where $m$: \#attributes, $n$: \#instances, $R$: \#unique categories in the attribute having largest values. & Accuracy, average inter-cluster distance, average intra-cluster distance & UCI & Table & The proposed distance measure improves the performance of \textsc{Knn} classification and \textsc{K-modes} clustering on various categorical datasets compared to traditional distance measures. \\\hline

\cite{he2008k} & Simple matching  & \textbf{$\mathcal{O}(I n k^2 r p^2)$} where $I$: \#iterations, $n$:\#instances, $k$: \#clusters, $r$: \#attributes, $p$: \#distinct categories & Accuracy, clustering error, runtime, scalability & UCI \newline Synthetic & Histogram, bar plot, line plot, table & \textsc{K-anmi} employs multiple hash tables to compactly represent clusterings, allowing for the computation of average normalized mutual information without accessing the original data, facilitating analysis of large datasets. \\\hline

\cite{chen2008data} & Similarity by using the NNIR tree, representing clusters by attribute value combination significance & \textbf{$\mathcal{O}(q^2 |C|) + \mathcal{O}(|C||CN| + k^2 |CN|) + \mathcal{O}(|CN|^2)$}, where $|C|$: \#instances in $C$, $q$: \#attributes, $|CN|$: total candidate node sets, $k$: \#clusters
 & Accuracy, runtime, scalability & UCI \newline Datgen.com & Bar plot, line plot, tree plot, table & \textsc{Mardl} is a data labeling mechanism that is used to allocate unlabeled data points to the appropriate clusters after an initial clustering has been performed on a sample of the data using any categorical clustering algorithm. \\\hline

\cite{cao2009new} & Simple matching & \textbf{$\mathcal{O}(|U|$ $\times$ $|A|$ $\times$ $k^2)$}, where $|U|$: \#instances, $|A|$: \#attributes, $k$: \#clusters
 & Precision, recall, accuracy, iteration time & UCI & Table & The proposed initialization method incorporates both object density and inter-object distance when selecting initial cluster centers, enhancing the performance of the \textsc{K-modes} and \textsc{Fuzzy K-modes}. \\\hline

\cite{ienco2009context} &Euclidean distance between conditional probability distributions of categorical values for correlated attributes &$O(nm^2)$, where $n$: \#instances, $m$: \#attributes. & Classification error, accuracy, NMI, runtime, scalability & UCI \newline Datgen.com & Line plot, table & \textsc{Dilca} (\emph{DIstance Learning in Categorical Attributes}) computes \emph{Symmetric Uncertainty} (\textsc{Su}) with other attributes, selects those with \textsc{Su} above a specified threshold, and then measures the \textsc{Euclidean} distance between the conditional probability distributions of these context attributes. \\\hline

\cite{bai2011initialization} &Simple matching & $\mathcal{O}(2nm|V| + |V| + mk^{2}|V|)$, where $n$: \#instances, $m$: \#attributes, $|V|$: \#categories, $k$: \#clusters  & Precision, recall, accuracy, runtime, scalability & UCI & Line plot, flowchart, table & The proposed method provides an initialization technique to find good initial cluster centers and estimate the number of clusters for \textsc{K-modes} and \textsc{Fuzzy K-modes}.\\\hline

\cite{bai2011novel} &Weighted distance measure adjusting based on individual attribute contributions
 &$\mathcal{O}(tmnk)$, where $t$: \#iterations, $n$: \#instances, $m$: \#attributes, $k$: \#clusters  & Category utility, ARI, error rate, runtime, scalability & UCI \newline Synthetic & Bar plot, matrix plot, line plot, table & \textsc{Mwkm} clusters high-dimensional categorical data by integrating multiple attribute weighting schemes and a weighted dissimilarity measure, all while maintaining scalability and efficiency.\\\hline

\cite{he2011attribute} &Simple matching & $\mathcal{O}(tmnk)$, where $t$: \#iterations, $n$: \#instances, $m$: \#attributes, $k$: \#clusters& Accuracy, runtime, scalability & UCI & Line plot, table & Extends \textsc{K-modes} by incorporating attribute value weighting and new distance and weighting functions to enhance intra-cluster similarity and overall clustering performance. The improved algorithm consistently outperforms \textsc{K-modes} in accuracy while maintaining scalability. \\\hline

\cite{bai2012cluster} &Simple matching  & $\mathcal{O}(|U||V|+|U|k^2+|U||V|k+|A|k^2)$, where $|U|$: \#instances, $|V|$: total \#attribute values, $|A|$: \#attributes, $k$: \#cluster & Precision, recall, accuracy, runtime, scalability & UCI & Line plot, table & A novel initialization method for \textsc{K-modes} considering both density and distance from the mode of the dataset, improving scalability. \\\hline

\end{tabular}
\end{adjustbox}
\end{table*}

\begin{table*}[!htb]

\vspace{-3cm}

\caption{Comparison of \textsc{Hard Partitional Clustering Algorithms} (continue Table \ref{tab:partitional_clustering_hard_1})}
\label{tab:partitional_clustering_hard_2}
\begin{adjustbox}{max width = 1.1\linewidth}
\renewcommand{\arraystretch}{2.5} 
\setlength{\arrayrulewidth}{0.001pt} 
\begin{tabular}{|p{1cm}|p{3.6cm}|p{3.6cm}|p{3.4cm}|p{2.3cm}|p{2.2cm}|p{9.9cm}|}
\hline

\textbf{Ref.} & \textbf{Sim/Dissim measure} & \textbf{Complexity} & \textbf{Evaluation metrics} & \textbf{Data sources} & \textbf{Plots used} & \textbf{Remarks}\\\hline

\cite{cao2012dissimilarity} & Distance metric that considers attribute distributions and integrates rough membership functions & $\mathcal{O}(t|U||P|k)$, where $t$: \#iteration, $|U|$: \#instances, $|P|$: \#attributes, $k$: \#clusters   & Accuracy, runtime, scalability  & UCI & Line plot, table, runtime, scalability & This dissimilarity measure aims to create tighter, more cohesive clusters by emphasizing uncommon matching attribute values across the whole dataset. \\\hline

\cite{khan2013cluster} & Simple matching & $\mathcal{O}(nm + rKm^2 n + n \log n)$, where $n$: \#instances, $m$: \#attributes, $r$: \#prominent attributes, $K$: \#clusters & Precision, recall, accuracy, matchMetric, runtime, scalability & UCI & Line plot, table & The proposed algorithm creates multiple data partitions using different attribute subsets to extract comprehensive information about the cluster structure, which is then used to determine high-quality initial cluster centers for the \textsc{K-modes} algorithm. \\\hline

\cite{cheung2013categorical} &Entropy-based similarity metric capturing categorical attribute distribution and significance in clusters &$\mathcal{O}(dN)$, where $d$: \#attributes, $|N|$: \#instances & Accuracy, clustering error, runtime & UCI & Scatter plot, line plot, table & The algorithm iteratively assigns objects to clusters based on average attribute similarity. It automates cluster number selection using penalized competitive learning, gradually removing redundant clusters based on assigned weights. \\\hline

\cite{cao2013weighting} &Dissimilarity between categorical object pairs increasing with the number of mismatched attributes &  $\mathcal{O}(tm|U|k)$, where $t$: \#iterations, $|U|$: \#instances, $m$: \#attributes, $k$: \#clusters  & ARI, accuracy, runtime, scalability & UCI & Line plot, table & \textsc{WK-modes} enhances the \textsc{K-modes} algorithm by assigning weights to each attribute using complement entropy. It identifies important subspaces by updating cluster centers and recalculating weights iteratively, improving clustering accuracy and efficiency for high-dimensional data. \\\hline

\cite{qian2015space} &Transformation of the categorical data into Euclidean space with subsequent use of Euclidean and Cosine distances & $\mathcal{O}(n^2)$, where $n$: \#instances& ARI, accuracy, runtime, scalability & UCI & Scatter plot, line plot, bar plot, table & \textsc{Sbc} algorithm maps the categorical objects into a \emph{Euclidean space}, where each original categorical object becomes a new dimension. This provides the categorical data with an explicit spatial structure representation. \\\hline

\cite{park2015rough} & Computation of the entropy-based roughness of attributes and selection of those with the highest dependency degrees & Not discussed & Purity & UCI & Table & \textsc{Itdr} uses rough set theory for categorical data by computing equivalence classes and selecting clustering attributes based on an entropy-based roughness measure. It iteratively partitions the data, aiming to minimize complexity and enhance cluster purity. \\\hline

\cite{jia2015new} & Dissimilarity between categorical objects based on frequency probabilities, attribute weights, and attribute interdependencies   & $\mathcal{O}(d^2n)$, where $d$: \#attributes, $n$: \#instances  & Davies-Bouldin index, accuracy, error rate, Rand index, NMI, cluster-discrimination index & UCI   & Table, line plot, bar plot  & \textsc{DM3}, the proposed distance metric, considers frequency probabilities and co-occurrence with highly correlated attributes. It dynamically weights attributes and evaluates their dependencies. When incorporated into the \textsc{K-modes} clustering algorithm, DM3 outperforms other distance metrics such as \emph{Hamming} distance. \\\hline

\cite{jiang2016initialization} & Weighted matching distance metric assigning weights to attributes based on their significance levels & Ini\_Distance: $\mathcal{O}(|U|^2|A|)$ for time, $\mathcal{O}(|U|+|A|)$ for space; Ini\_Entropy: $\mathcal{O}(|A||U|(K+|A|))$ for time, $\mathcal{O}(|U|(K+|A|))$ for space, where $|U|$: \#instances, $|A|$: \#attributes, $K$: \#clusters   & Precision, recall, accuracy, runtime, scalability & UCI & Line plot, confusion matrix, table & Two proposed initialization algorithms, \textsc{Ini\_Entropy} and \textsc{Ini\_Distance}, which use outlier detection techniques to avoid selecting outliers as initial centers, while ensuring the selected centers are well-separated from each other. \\\hline

\cite{cao2017algorithm} & Jaccard coefficient & $\mathcal{O}(nmtk \times 2^{|V'|})$, where $n$: \#instances, $m$: \#attributes, $k$: \#clusters, $t$: \#iterations, $|V'|$: maximum number of unique values in any feature  & ARI, NMI, accuracy, precision, recall, runtime, scalability & UCI \newline Alibaba \newline GroupLens Research & Scatter plot, line plot, table & \textsc{SV-K-modes} extends \textsc{K-modes} to handle categorical data with set-valued features, using set-valued modes as cluster representatives, a \emph{Jaccard} coefficient-based distance measure. It proposed a method named HAFSM for updating these \emph{modes}, and GICCA for cluster \emph{modes} initialization. \\\hline

\cite{uddin2017empirical}   & Does not use distance metrics, but rather rely on the indiscernibility relations among the attributes to form clusters   &$\mathcal{O}(nm)$ where $n$: \#instances, $m$: \#attributes  & Purity, entropy, accuracy, number of iterations, rough accuracy, runtime & UCI & Bar plot, table & \textsc{Mia} (\emph{Maximum Indiscernible Attribute}) utilizes the \emph{rough indiscernibility relation} of attributes to identify the best clustering attribute. It calculates the indiscernibility relations and their cardinalities for each attribute, selecting the one with the highest cardinality for clustering. \\ \hline

\cite{salem2018fast} & Manhattan distance & $\mathcal{O}(NKTd)$ where $N$: \#instances, $d$: \#attributes, $K$: \#clusters, $T$: \#iterations  & Purity, entropy, NMI, runtime, scalability & Maryland Univ \newline UCI \newline Baltimore Police Dept & Tree plot, line plot, table & \textsc{MFK-M} converts categorical values into numeric measures using the relative frequency of each category (modality) within the attributes, and uses a stop criterion to avoid trivial clusterings such as cluster death cases where all observations belong to only one cluster. \\\hline

\cite{huang2018new} & Frequency-based distance measure and a normalized distance measure &  $\mathcal{O}(3tnm+3tkn(mlog(m)+m)+tknm)$  where $n$: \#instances, $m$: \#attributes, $k$: \#clusters, $t$: \#iterations & Accuracy, Rand index, F-score, NMI, runtime, scalability& UCI & Line plot, bar plot, table, critical difference diagram & $L^2$-NOF and $L^2$-NDM employ $L^2$-norm regularization on feature weights and use frequency-based vectors to represent cluster centroids. The two algorithms aim to minimize intra-cluster scatter and the $L^2$-norm of feature weights in their objective functions. \\\hline

\cite{surya2018attributes} &Euclidean distance & Not discussed & Clustering error, convergence time, accuracy, ARI, dimensionality variation, convergence rate, runtime & UCI & Flowchart, bar plot, table & The algorithm preprocesses the dataset to remove noise and irrelevant attributes, computes inter- and intra-attribute similarities, filters significant attributes, and uses association rule mining to estimate support and confidence. It then performs \textsc{K-Medoids} clustering with Euclidean distance and uses bee colony optimization to select optimal features, enhancing clustering performance. \\\hline

\end{tabular}
\end{adjustbox}
\end{table*}


\begin{table*}[!htb]

\vspace{-3cm}
\caption{Comparison of \textsc{Hard Partitional Clustering Algorithms} (continue Table \ref{tab:partitional_clustering_hard_2})}
\label{tab:partitional_clustering_hard_3}
\begin{adjustbox}{max width = 1.1\linewidth}
\renewcommand{\arraystretch}{2.5} 
\setlength{\arrayrulewidth}{0.001pt} 
\begin{tabular}{|p{1cm}|p{3.6cm}|p{3.6cm}|p{3.4cm}|p{2.3cm}|p{2.2cm}|p{9.9cm}|}
\hline

\textbf{Ref.} & \textbf{Sim/Dissim measure} & \textbf{Complexity} & \textbf{Evaluation metrics} & \textbf{Data sources} & \textbf{Plots used} & \textbf{Remarks}\\
\hline

\cite{dinh2019estimating} & Information theoretic-based dissimilarity measure & Not discussed & Silhouette score & UCI \newline Datgen.com & Flowchart, table, dendrogram, line plot, silhouette plot &\textsc{K-scc} uses \emph{kernel density estimation} to form cluster centers and an \emph{information theoretic dissimilarity} measure to calculate distances between objects and cluster centers. It employs the \emph{silhouette coefficient} to evaluate clustering quality and estimate the optimal $k$. \\\hline

\cite{xiao2019optimal} & Distance metric indicating the number of data points classified into different clusters in any two solutions of the ILP model& $\mathcal{O}(K_{\text{max}} L_{\text{max}}T)$, where: $K_{\text{max}}$: farthest neighborhood, $L_{\text{max}}$: the maximum tries between two improvements, $T$: the average CPU time& Total inner-distance, runtime, scalability & UCI & Bar plot, line plot, pie chart, table & \textsc{IPO-ILP-VNS} formulates \textsc{K-modes} as an \emph{integer linear programming} (ILP) model for optimal solutions in small datasets. For larger datasets, it combines ILP with \emph{variable neighborhood search} to find near-optimal solutions efficiently. \\\hline

\cite{dinh2020k} & Information theoretic-based dissimilarity measure & $\mathcal{O}(\alpha knmt$) where $n$: \#instances, $m$: \#attributes, $k$: \#clusters, $\alpha$: \#iterations, $t$: average \#categories per attribute & Silhouette score, accuracy, precision, recall, purity, NMI, ARI, runtime, scalability & UCI & Bar plot, line plot, tree plot, flowchart, table & \textsc{K-pbc} leverages \emph{maximal frequent itemset mining} to form initial clusters, which enhances the overall clustering quality. It then uses \emph{kernel density estimation} and \emph{information theoretic dissimilarity} to form cluster centers and assign data into clusters.\\\hline

\cite{zhang2020new} & Distance between categories by exploiting the information provided by different attributes that correlate with a specific attribute when calculating distances& $\mathcal{O}(d^2N)$, where $d$: \#attributes, $N$: \#instances & Accuracy, ARI, NMI, label-order consistency, runtime, scalability & UCI \newline Weka \newline Shenzhen Univ \newline Education Univ of Hong Kong & Bar plot, line plot, gray scale map, table & \textsc{UDM} quantifies distances between mixed \emph{ordinal} and \emph{nominal} attributes by using order information and attribute interdependence. It calculates distances based on the entropy of joint distributions with correlated attributes, preserving ordinal relationships. UDM weights the contributions of different attributes by their interdependence, resulting in a unified measure for clustering. \\\hline

\cite{zhang2021learnable} & Distances between attribute values using conditional probability distributions for both nominal and ordinal attributes & $\mathcal{O}(E(kdVnI+dn+dV^2k)$, where $E$: \#outer loop iterations, $k$: \#clusters, $d$:\#attributes, $V$: \#maximum categories per attribute, $n$: \#instances, $I$: \#inner loop iterations. & Bonferroni-Dunn test, Wilcoxon signed-rank test, ARI, NMI, accuracy, intra- and inter-cluster distance, error rate, runtime, scalability & UCI \newline Weka \newline Shenzhen Univ & Gray scale map, scatter plot, line plot, bar plot, critical difference diagram, table & \textsc{HD-NDW} integrates \emph{Homogeneous Distance} (HD) and \emph{Novel Distance Weighting} (NDW) mechanisms to cluster data with nominal and ordinal attributes. \textsc{HD} defines \emph{intra-attribute distances} uniformly for both attribute types, while \textsc{NDW} iteratively updates the weights of these distances based on current data partitions. This iterative process refines the distance weights and data partitions until convergence.\\\hline

\cite{mau2021lsh} & Distance between two categorical attribute values determined by how their distributions among other attributes are observed across the dataset & Neighborhood updating phase: $\mathcal{O}(D|\Bar{A}|k(k-1)/2)$, where $\Bar{A}$: average size of attribute domain, $D$: \#attributes, $k$: \#clusters  & Accuracy, purity, NMI, ARI, AMI, homogeneity, silhouette score, runtime, scalability, memory usage& UCI \newline Datgen.com & Line plot, bar plot, flowchart, table & \textsc{LSH-k-representatives} combines \emph{Locality-Sensitive Hashing} (LSH) with \emph{K-representatives} clustering for large-scale categorical data. It employs data-driven dissimilarity measures to create \emph{binary hash functions} for the initial clustering by grouping similar objects into \emph{buckets}. Iteratively, it optimizes object reassignment using a \emph{nearest neighbor search} within a subset of clusters. \\\hline

\cite{sajidha2021initial} &Simple matching & $\mathcal{O}(n^2)$, where $n$: \#instances, & Precision, recall, accuracy & UCI & Table & The algorithm selects initial seeds based on density and distance from the mode. It starts with the densest and farthest artifact from the mode, then selects subsequent seeds to ensure they are well-separated. \\\hline

\cite{salem2021rough} &Simple matching & $\mathcal{O}(NK|C_k|ptdn_{c_{kj}})$, where $N$: \#instances, $K$:\#clusters, $d$: \#attributes, $t$: \#iterations, $|C_k|$: cardinality of cluster $C_k$, $p$: \#candidate modes, $n_{c_{kj}}$: \#instances having the category $c_{kj}$  & Accuracy, silhouette score, entropy, NMI, precision, recall, F-score, runtime, scalability & UCI \newline Twitter \newline Maryland Univ & Line plot, box plot, table & \textsc{DRK-M} selects optimal modes using rough set theory in three steps: initializing \emph{k} cluster modes, assigning observations with a matching dissimilarity measure, and updating modes by considering attribute density and frequency to find central centroids. \\\hline

\cite{dinh2021clustering} &Information theoretic-based dissimilarity measure & $\mathcal{O}(nm+nm_{objs}m^2+\lambda knmt$), where $n$: \#instances, $m$: \#attributes, $m_{objs}$: \#instances with missing values, $k$: \#clusters, $\lambda$: \#iterations, $t$: average \#categories per attribute & Purity, NMI, homogeneity, completeness, memory usage, runtime, scalability & UCI & Line plot, bar plot, heatmap, tree plot, flowchart, table & \textsc{K-ccm*} integrates imputation and clustering into a unified process that can be used for clustering categorical datasets with missing values, eliminating the need for separate preprocessing steps. \\\hline

\cite{nguyen2023method} &Information theoretic-based dissimilarity measure & Not discussed & Purity, NMI, ARI & UCI & Table & The proposed \textsc{K-means-like} algorithm maximizes the intra-cluster similarity and minimize inter-cluster similarity by using an information theoretic-based dissimilarity measure and kernel density-based cluster centers.\\\hline

\cite{kar2023efficient} & Entropy to capture intra-attribute statistical information to assess attribute significance & $\mathcal{O}(dn^2)$ where $n$: \#instances, $d$: \#attributes & Rand index, ARI, accuracy, cluster discrimination matrix, runtime, scalability & Kaggle \newline UCI \newline NHTSA & Flowchart, gray scale map, line plot, table & Proposes a non-parametric distance metric for \textsc{K-modes}, which ensures that attributes with more information content have a greater influence on the dissimilarity measure. \\\hline

\cite{santos2023fair} & Combination of coincidence matrix, Fleiss' kappa coefficient, and Euclidean distance & Not discussed & Fowlkes-Mallows index, maximum-match measure, normalized variation of information, overlap coefficient, fairness ratio & Kaggle \newline UCI \newline Macquarie Univ& Table & \textsc{Fair-Multicluster} partitions categorical data into clusters that are both \emph{homogeneous} and \emph{fair}. It starts by forming initial clusters based on single attributes, then iteratively merges these into multiclusters based on attribute similarities. Clusters are adjusted to match a desired distribution of a protected attribute, ensuring fairness. It continues until the desired number of clusters is attained.\\\hline

\end{tabular}
\end{adjustbox}
\end{table*}


\begin{table*}[!htb]

\vspace{-3.2cm}

\caption{Comparison of \textsc{Fuzzy Partitional Clustering Algorithms}}

\label{tab:partitional_clustering_fuzzy}

\begin{adjustbox}{max width = 1.1\linewidth}
\renewcommand{\arraystretch}{1.15} 
\setlength{\arrayrulewidth}{0.001pt} 
\begin{tabular}{|p{1.2cm}|p{1.9cm}|p{6cm}|p{2.7cm}|p{1.9cm}|p{2.3cm}|p{9.5cm}|}
\hline

\textbf{Ref.} & \textbf{Fuzzification parameter} & \textbf{Complexity} & \textbf{Evaluation metrics} & \textbf{Data sources} & \textbf{Plots used} & \textbf{Remarks}\\\hline

\cite{huang1999fuzzy} & $m$ = 1.5  & Time: $\mathcal{O}(kmn+kn(m+M))$, space: $\mathcal{O}(m(n+k)+km)$, where $t$: \#iterations, $k$: \#clusters, $n$: \#instances, $m$: \#attributes, $M$: total \#categories of all attributes & Accuracy, runtime, scalability & UCI & Bar plot, table & \textsc{Fuzzy K-modes} extends \textsc{K-modes} by assigning \emph{fuzzy membership values} to objects in different clusters, providing more information about cluster boundaries and uncertainty.    \\ \hline

\cite{wang1999clustering} & Not used & Not discussed & Purity & UCI & Table & The algorithm uses \emph{large items} (frequent items in a cluster) to measure cluster similarity. It updates the \emph{supports} of items when transactions are added or removed, using \emph{Hash} and \emph{Btree} data structures.     \\ \hline

\cite{oh2001fuzzy} & $T_u$=0.1, $T_\omega$=1.5  & Not discussed & Comparison of results with those from Correspondence analysis & Literature retrieval dataset  & Scatter plot, table & \textsc{Fccm} maximizes a degree of aggregation objective function to assign fuzzy memberships, using entropy maximization for regularization. It iteratively updates memberships until a stopping condition is met, optimizing for local maxima.     \\ \hline

\cite{ng2002clustering}   &$m$ = 1.5    & Not discussed  & Accuracy, runtime & UCI & Scatter plot, line plot, table & The algorithm combines the \textsc{Fuzzy K-modes} approach with \emph{tabu search techniques} to avoid converging to a local optimum.     \\ \hline

\cite{kim2004fuzzy}  & $m$ = 1.8    &  Time:  $\mathcal{O}(kpn(N+1)s)$, space: $\mathcal{O}(p(kN+n)+kn)$, where $n$: \#instances, $p$: \#attributes, $k$: \#clusters, $N$: \#maximum categories for any attribute, $t$: \#iterations    & Accuracy, runtime, scalability & UCI & Bar plot, line plot, table &  This work enhances \textsc{Fuzzy K-modes} for categorical data by using fuzzy centroids instead of hard-type centroids. This approach better represents uncertainty and improves clustering performance.   \\ \hline

\cite{mukhopadhyay2009multiobjective} & $m = 2.0$ & Not discussed & ARI, t-test, runtime & UCI \newline Datgen.com & VAT, Pareto front plots, table & This multi-objective genetic clustering around medoids algorithm can simultaneously optimize the often conflicting criteria of compactness and separation of clusters. \\ \hline

\cite{maulik2010integrating}   & $m = 2.0$   & $\mathcal{O}(KNP_{opsize}M_{axgen})$  where  $N$: \#instances, $P_{opsize}$: population size, $M_{axgen}$: maximum \#generations, $K$: \#clusters   & ARI, Minkowski score, Xie–Beni index, I index & UCI \newline Datgen.com & VAT, box plot, line plot, table & \textsc{MoDEFCMdd} is a differential evolution-based \textsc{Fuzzy C-Medoids} algorithm utilizing \emph{medoids} as cluster representatives. It combines local and global information with adaptive weighting during mutation for improved optimization.  \\ \hline

\cite{bai2013novel}  & $m \in (1, 1.5]$    &  $\mathcal{O}\left(n \sum_{j=1}^{m} n_j + mnk \sum_{e=1}^{o} t_e \right)$ where $n$: \#instances, $m$: \#attributes, $n_j$: \#categories for attribute $a_j$, $k$: \#clusters, $t_e$: \#iterations to obtain local minima & Accuracy, recall, precision, partition coefficient, partition entropy & UCI & Bar plot, line plot, table & \textsc{Nfkm} improves \textsc{Fuzzy K-modes} by incorporating between-cluster information to minimize within-cluster dispersion and enhance between-cluster separation. It iteratively updates the membership matrix and cluster prototypes until convergence.   \\ \hline

\cite{yang2015non}  &  $m = 1.2$  & Not discussed & ARI, compactness, fuzzy separation, runtime & UCI & Scatter plot, flowchart, table & \textsc{Nsga-fmc} optimizes fuzzy membership assignments and cluster compactness/separation objectives simultaneously using a multi-objective genetic algorithm approach. \\ \hline

\cite{yanto2016modified}   &   $m \in [1.1, 2.0]$   & $\mathcal{O}(KM(I+1)t+IJ)$ where $t$: \#iterations, $K$: \#clusters, $I$: \#instances, $J$: \#attributes, $M$: total \#categories of all attributes     & Purity, ARI, Davies-Bouldin index, Dunn index, MSE, runtime & UCI & Line plot, table &  \textsc{FID} preprocesses categorical data by forming equivalence classes using the \emph{indiscernibility relation}. The algorithm iteratively optimizes cluster membership and parameters to reduce complexity and enhance accuracy.  \\ \hline
 
\cite{zhu2018many}  & $m = 1.2$  & $\mathcal{O}(GNKnd(T+1)+nK^2r^2)$ where $G$: \#generations, $N$: population size, $K$: \#clusters, $n$: \#instances, $d$: \#attributes, $T$: \#maximum categories for any attribute, $r$: \#solutions to be integrated in the cluster ensemble method & Accuracy, ARI, runtime, scalability & UCI \newline Datgen.com & Box plot, line plot, bar plot, flowchart, table & \textsc{MaOFcentroids} applies evolutionary many-objective optimization, optimizing multiple cluster validity indices simultaneously to achieve a more nuanced and effective clustering solution. \\ \hline

\cite{saha2019integrated}  & $m = 2.0$   & $\mathcal{O}(2Knm + n + Kn + KMn + \alpha \times \max I_{\text{tr}} \times (2Km + 2Knm + n + Kn + KMn) + \max \text{Gen} \times \text{PS} (Kn + KMn) + T nm \log n)$, where $n$: \#instances, $m$: \#attributes, $K$: \#clusters, $T$: \#trees in the Random Forest, $PS$: population size, $M$: total \#categories of all attributes, $\alpha$: \#repetitions, $maxI_{tr}$: max iterations, $maxGen$: max generations  & Minkowski score, percentage of correct pair, Davies–Bouldin index, Dunn index, Xie–Beni index, runtime & UCI \newline Datgen.com   & VAT, box plot, flowchart, table & \textsc{IRFKMd-RF} (\textsc{Integrated Rough Fuzzy K-Modes using Random Forest}) combines three clustering methods—RFKMd, SARFKMd, and GARFKMd—to generate sets of central and peripheral points. A \emph{roughness measure} is calculated for each method, and the one with the lowest value is chosen to determine the \emph{best central points}. Points found in one or two other sets but not in the best set are considered \emph{semi-best central points}, while those not in any set are classified as \emph{pure peripheral points}. Random Forest is then used to classify the \emph{semi-best central points} and p\emph{pure peripheral points}. \\ \hline

\cite{nguyen2019partition}  & $m = 2.0$ &  $\mathcal{O}(2n^2 + \sqrt{n}-k+n)$ where $n$: \#instances, $k$: \#clusters & ARI, NMI, Davies–Bouldin index, runtime & UCI & Bar plot, flowchart, table & \textsc{PM-FGCA} partitions the dataset into clusters using a fuzzy genetic approach. It then merges clusters based on inter-cluster distances until reaching a specified cluster count. Finally, it improves clustering quality by reassigning instances based on intra-cluster distances. \\ \hline

\cite{kuo2019genetic}  & Not specified  & $\mathcal{O}(k^2n^2(2m+M))$ where $k$: \#clusters, $n$: \#instances, $m$: \#attributes, $M$: total \#categories in all attributes   & Accuracy, ARI, cluster discrimination index, runtime & UCI & Bar plot, flowchart, table & \textsc{GIWFKM} combines \emph{intuitionistic fuzzy sets} with \textsc{K-modes}, using a \emph{frequency probability-based distance} metric and \emph{weighted attributes}. It employs a \emph{genetic algorithm} (GA) for global optimization and includes feature selection to reduce redundancy and computation time.\\ \hline

\cite{naouali2020uncertainty}  & $m \in [1, +\infty]$ & $\mathcal{O}(NK|C_k|ptdn_{ck_j})$ where $N$: \#observations, $K$: \#clusters, $|C_k|$: cardinality of cluster, $p$: \#candidate modes in a cluster, $t$: \#iterations, $d$: \#attributes, $n_{ck_j}$: \#instances having the category $ck_j$& Accuracy, entropy, NMI, precision, recall, runtime, scalability & UCI \newline Twitter \newline University of Maryland & Scatter plot, line plot, table & \textsc{DRK-M} enhances \textsc{K-modes} by minimizing mode selection randomness. It first selects random centroids and assigns observations based on the simple matching. Modes are derived from frequent attribute values, prioritizing those with higher densities to iteratively refine centroids for more stable clustering. \\ \hline

\cite{sarkar2020machine}  & $m = 2.0$    & Time: $\mathcal{O}(nm+KMn+K^2n)$, Space: $\mathcal{O}(2nm+Km+3Kn)$ \newline where $K$: \#clusters, $n$: \#instances, $m$: \#attributes, $M$: total \#categories in all attributes   & Minkowski score, percentage of correct pair, ARI, Xie–Beni index, average rank, chi-square & UCI \newline Datgen.com & Bar plot, box plot, VAT, table & \textsc{MLCrKMd} enhances the \emph{Credibilistic K-modes} (CrKMd) by integrating various machine learning techniques to handle uncertain data points. It computes the \emph{credibility matrix} to identify certain and uncertain data. Certain data are used to train machine learning models such as \emph{K-NN}, \emph{SVM}, \emph{ANN}, \emph{Decision Tree}, and \emph{Random Forest}. These models then classify the uncertain data to identify the best clustering. \\ \hline

\cite{kuo2021metaheuristic}  & $m = 2.0$  & Not discussed  & Sum of squared errors, accuracy, runtime & UCI & Line plot, flowchart, table & \textsc{Pfkm} mitigates outlier and noise effects in clustering. It integrates meta-heuristic algorithms like genetic algorithms, particle swarm optimization, and sine cosine algorithms to improve initial centroid and parameter selection.
 \\ \hline

\cite{oskouei2021fkmawcw}  & $m = 2.0$   & $\mathcal{O}(KN(3M + \hat{M}))$ where $K$: \#clusters, $N$: \#instances, $M$: \#attributes, $\hat{M}$: total \#categories in all attributes  & ARI, accuracy, precision, recall & UCI \newline Synthetic & Heatmap, line plot, bar plot, table & \textsc{FKMAWCW} enhances \textsc{Fuzzy K-modes} by incorporating local attribute and cluster weighting. It assigns varying weights to attributes within each cluster and adjusts cluster weights iteratively to improve clustering results. \\ \hline

\cite{jiang2023kernel}  & Not specified & $\mathcal{O}(k(2n)(2m+M))$ where $k$: \#clusters, $n$: \#instances, $m$: \#attributes, $M$: total \#categories in all attributes  & ARI, NMI, accuracy, F-score & UCI & Bar plot, flowchart, table & \textsc{KIWFKM-DCP} combines \emph{kernel tricks} and \emph{intuitionistic fuzzy sets} in the \textsc{Weighted Fuzzy K-modes} algorithm. The coupled DCP system uses \emph{DNA genetic rules} (selection, crossover, mutation) to enhance clustering, avoid local optima, and improve robustness to noise and attribute importance.\\ \hline

\end{tabular}
\end{adjustbox}
\end{table*}

\begin{table*}[!htb]
\vspace{-3.4cm}
\caption{Comparison of \textsc{Hierarchical Clustering Algorithms}}
\label{tab:partitional_hierarchical_clustering}
\begin{adjustbox}{max width = 1.1\linewidth}
\renewcommand{\arraystretch}{1.5} 
\setlength{\arrayrulewidth}{0.001pt} 
\begin{tabular}{|p{1.3cm}|p{2.2cm}|p{6cm}|p{2.7cm}|p{1.9cm}|p{2.3cm}|p{9.5cm}|}
\hline

\textbf{Ref.} & \textbf{Type} & \textbf{Complexity} & \textbf{Evaluation metrics} & \textbf{Data sources} & \textbf{Plots used} & \textbf{Remarks}\\\hline

\cite{guha2000rock} & Agglomerative & Time: $\mathcal{O}$$(n^2$+$n m_{max} m_{avg}$ +$n^2logn)$, Space: $\mathcal{O}(min\{n^2,nm_{avg}m_{max}\})$ where $n$: \#instances, $m_{avg}$ and $m_{max}$: average and maximum \#neighbors for a point, respectively & Misclassification rate, runtime, scalability & UCI \newline MIT AI laboratories \newline Synthetic & Line plot, table & \textsc{Rock} starts with each data point as its own cluster and merges the most similar clusters based on a criterion function until reaching the desired number of clusters. It measures cluster similarity through \emph{links} between data points instead of traditional distance metrics.\\
\hline

\cite{andritsos2004limbo}   & Agglomerative & $\mathcal{O}(nhdB+dUB^2+L^2d^2\log L+kdn)$ where $n$: \#instances, $h$: height of tree, $d$: \#categories, $B$: branching factor of tree, $U$: \#non-leaf nodes in tree, $L$: \#entries at the leaves of the tree, $k$: \#clusters & Information loss, category utility, classification error, precision, recall, runtime, scalability & UCI \newline Datgen.com & Tree plot, bar plot, line plot, scatter plot, table & \textsc{Limbo} can produce high-quality clusterings comparable to the \emph{Agglomerative Information Bottleneck} algorithm but in a scalable manner by operating on the \emph{DCF Tree summary} instead of the full data.      \\ \hline

\cite{li2004entropy}   & Divisive   &  Not discussed     & Purity, recovery rate& UCI \newline UC Berkeley & Line plot, confusion matrix, bar plot, table & The algorithm uses a probabilistic framework to minimize an entropy-based criterion, equivalent to maximizing the mutual information between the data distribution and cluster assignments.      \\ \hline

\cite{he2004organizing}   & Agglomerative  &   $\mathcal{O}(n^2m)$ where $n$: \#instances, $m$: \#attributes  & Conditional entropy & Deep web collected from search engines& Line plot, bar plot, tree plot, table & Considers each query schema (e.g. {author, title, publisher} for Amazon) as a transaction or categorical data point. The motivation is to accurately organize a large number of structured web sources/databases based on their query schemas. \\ \hline

\cite{dutta2005qrock}   & Agglomerative  &   $\mathcal{O}(n^2)$ where $n$: \#instances & Runtime, scalability  & UCI & Table & \textsc{Qrock} uses connected components in a graph to identify clusters, cutting computational time significantly compared to \textsc{Rock}. \\ \hline

\cite{cesario2007top} & Divisive & Not discussed & Rand, Jaccard, Fowlkes-Mallows index, error rate, runtime, scalability & Cornell Univ \newline UCI \newline Synthetic & Bitmap representation, matrix, bar plot, line plot, table & \textsc{At-dc} begins with the entire dataset as one cluster and recursively splits it to enhance clustering quality. It uses a frequency-based measure to favor clusters with a high overlap of frequent items or attribute values in transactions, greedily reassigning transactions between candidate clusters to maximize quality gain. \\ \hline

\cite{ienco2012context}   & Agglomerative  & DILCA M: $\mathcal{O}(nm^2)$, DILCA RR: $\mathcal{O}(nm^2\log m)$ where $n$: \#instances, $m$: \#attributes   & Purity, NMI, ARI, runtime, scalability & UCI, Kaggle, synthetic & Line plot, table & While DILCA generally outperforms competitors, its performance can degrade on small datasets with unrepresentative attribute value distributions; furthermore, context selection may exclude relevant attributes. \\ \hline

\cite{xiong2012dhcc}   & Divisive   & Time: $\mathcal{O}(nJr + \psi nJ\log K)$, space: $\mathcal{O}(nJ)$ where $n$: \#instances, $J$: \#attributes, $r$: rank of the indicator matrix, $\psi$: average \#refinement iterations, $K$: \#clusters  & NMI, category utility, runtime, scalability & UCI \newline Synthetic & Scatter plot, incidence matrix, box plot, line plot, table & \textsc{Dhcc} converts data into an indicator matrix, performs an initial split using \emph{multiple correspondence analysis} (MCA), and iteratively refines clusters by relocating objects to optimize quality. This process continues until no further improvements are possible, producing a hierarchical structure that identifies clusters within data subspaces.     \\ \hline

\cite{li2014hierarchical}   & Divisive    & $\mathcal{O}(knm^2)$ where $n$: \#instances, $m$: \#attributes, $k$: \#clusters & Purity, ARI, NMI, runtime, scalability & UCI & Tree plot, table & \textsc{Mtmdp} is based on probabilistic rough set theory and introduces the concept of distribution approximation precision as a more effective uncertainty measure compared to traditional roughness measures.     \\ \hline

\cite{pang2018parallel}  & Agglomerative   &  Not discussed & ARI, Jaccard index, purity, runtime, scalability & UCI, China National Astronomical Observatory, Datgen.com & Line plot, bar plot, table & \textsc{Papu} uses \emph{Locality Sensitive Hashing} (LSH) to partition large-scale categorical data into smaller subdatasets. Subspace clustering is performed on these partitions in parallel using \textsc{Hadoop}. Local subclusters are merged with hierarchical agglomerative clustering to form a global dendrogram. \\ \hline

\cite{vsulc2019comparison}  & Agglomerative & Not discussed    & Pseudo-F coefficient based on the mutability, Pseudo-F coefficient based on the entropy & Synthetic &  Table & Two similarity measures are introduced. \textsc{Variable Entropy} (VE) uses entropy to weigh matches in the categories of variables, while \textsc{Variable Mutability} (VM) uses nominal variance (Gini coefficient). These measures enhance clustering quality without binary transformation, employing a proximity matrix and standard linkage methods such as single, complete, or average linkage.\\ \hline

\cite{wei2019hierarchical}  &  Divisive   & Not discussed   & Accuracy, ARI  & UCI & Flowchart, table & The algorithm selects informative attributes for splitting a cluster by evaluating measures like \emph{Total Roughness} (TR) or \emph{Mean Information Gain Ratio} (MGR). Using these attributes, it generates bipartitions to create two sub-clusters and then determines which sub-clusters should be further split in subsequent iterations.\\ \hline

\cite{sankhe2022mutual}  &  Divisive    & Not discussed  & Purity & UCI, National Data Archive on Child Abuse and Neglect & Table & \textsc{MIS} partitions data into clusters using mutual information scores. It recursively identifies the most informative attribute and uses it to create data partitions. These partitions are evaluated for entropy, and the one with the least entropy forms a new cluster. This iterative process is repeated until the specified number of clusters are formed. \\ \hline

\end{tabular}
\end{adjustbox}
\end{table*}
\begin{table*}[!htb]
\caption{Comparison of \textsc{Model-based Clustering Algorithms}}
\label{tab:model-based_clustering}
\begin{adjustbox}{max width = 1.1\linewidth}
\renewcommand{\arraystretch}{1.5} 
\setlength{\arrayrulewidth}{0.001pt} 
\begin{tabular}{|p{1.5cm}|p{3.6cm}|p{4cm}|p{3.1cm}|p{1.9cm}|p{2cm}|p{9.5cm}|}
\hline

\textbf{Ref.} & \textbf{Estimation method} & \textbf{Complexity} & \textbf{Evaluation metrics} & \textbf{Data sources} & \textbf{Plots used} & \textbf{Remarks}\\\hline

\cite{chen2012model}   & Expectation-Maximization algorithm & $\mathcal{O}(\max\{lr^2,ln\})$ where $l$: \#latent variables, $r$: maximum \#neighbors a node can have, $n$: \#manifest variables & BIC, MI, NMI, runtime & LCA literature, ICAC of Hong Kong, UCI & Tree plot, line plot, table &\textsc{East} employs heuristic strategies like local \emph{Expectation-Maximization} and a \emph{grow-restructure-thin} approach to iteratively improve the model, making the latent tree analysis scalable for datasets with over 100 attributes.      \\ \hline

\cite{gollini2014mixture}   & Variational Expectation-Maximization algorithm, with Gauss-Hermite quadrature in final step & Not discussed & Chi-squared test, truncated sum of squares Pearson residual criterion, BIC, adjusted BIC & National Institute on Aging, UCI & Line plot, table & The \textsc{Mixture of Latent Trait Analyzers (MLTA)} model is designed for model-based clustering of categorical data. It extends \emph{latent class analysis} by introducing a continuous latent trait variable to capture dependencies within groups.\\ \hline

\cite{jacques2018model}  & SEM-Gibbs algorithm  & Not discussed & Integrated completed likelihood-Bayesian information criterion (ICL-BIC), ARI & EORTC, Kaggle & Line plot, heatmap, bar plot, box plot, table & The algorithm uses a \emph{latent block model} based on the \emph{Binary Ordinal Search} (BOS) distribution for handling ordinal data. It employs a \emph{Stochastic EM} (SEM) algorithm coupled with a \emph{Gibbs sampler} for model inference and the \emph{ICL-BIC criterion} to select the number of co-clusters. It can also handle missing data.\\ \hline

\cite{zhang2019unified}   & Entropy-based measures to estimate the distance between ordinal and nominal attributes & $\mathcal{O}(kdNI)$ where $N$: \#instances, $d$: \#attributes, $k$: \#clusters, $I$: \#iterations     & Accuracy, ARI, NMI & Education University of Hong Kong, Shenzhen University, Weka, UCI & Pie chart, bar plot, table & \textsc{Ebdm} measures distances in categorical data by using \emph{entropy} to account for \emph{ordinal} and \emph{nominal} attributes. It calculates the entropy values for ordered categories in ordinal data and uses statistical information for nominal data, offering a unified, non-parametric method for clustering.     \\ \hline

\end{tabular}
\end{adjustbox}
\end{table*}


\begin{table*}[!htb]
\vspace{-3.2cm}
\caption{Comparison of \textsc{Ensemble Clustering Algorithms}}
\label{tab:ensemble_clustering}
\begin{adjustbox}{max width = 1.1\linewidth}
\renewcommand{\arraystretch}{1.8} 
\setlength{\arrayrulewidth}{0.001pt} 
\begin{tabular}{|p{1.3cm}|p{4cm}|p{5cm}|p{3.2cm}|p{2.3cm}|p{2.3cm}|p{9.5cm}|}
\hline

\textbf{Ref.} & \textbf{Integration strategy} & \textbf{Complexity} & \textbf{Evaluation metrics} & \textbf{Data sources} & \textbf{Plots used} & \textbf{Remarks}\\\hline

\cite{he2005cluster}  &  Three hypergraph-based algorithms from Strehl and Ghosh's work for integrating the clusters   & Not discussed   & Accuracy, clustering error, NMI & UCI & Bar plot, table & \textsc{CcdByEnsemble} (Clustering Categorical Data By Cluster Ensemble) combines multiple clustering results to achieve more accurate and robust final clustering. It uses three hypergraph-based methods--CSPA, HGPA, and MCLA--toselect the one with the highest \emph{average normalized mutual information} as the final output.\\
\hline

\cite{gionis2007clustering}   & Clustering aggregation formulated as an optimization problem to minimize disagreements with the input clusterings  & BestClustering: $\mathcal{O}(mn)$, Balls: $\mathcal{O}(mn^2+n^2)$, Agglomerative: $\mathcal{O}(mn^2+n^2\log n)$, Furthest: $\mathcal{O}(mn^2+k^2n)$, LocalSearch: $\mathcal{O}(In^2)$, Sampling: $\mathcal{O}(\log n)$, where $m$: \#input clusterings, $n$: \#instances, $k$: \#clusters  & Classification error, disagreement error, runtime, scalability & UCI & Scatter plot, line plot, table & Clustering aggregation combines multiple clusterings, handles various data complexities, and leverages the strengths of different clustering algorithms in a unified manner.      \\ \hline

\cite{iam2010link}   &  Spectral graph partitioning on a weighted bipartite graph from the refined cluster-association matrix to obtain the final clustering  & $\mathcal{O}(P^2+NP)$, where $N$: \#instances, $P$: \#clusters in the ensemble & Accuracy, NMI, ARI, significant better performance, significant worse performance, scalability & UCI \newline 20 Newsgroups & Bar plot, line plot, flowchart, table & \textsc{Lce} uses a link-based approach that effectively captures cluster-cluster relationships to refine the cluster association information.     \\ \hline

\cite{zhao2017clustering}  & Three consensus functions (CSPA, HGPA, MCLA) to generate the final clustering results   & $\mathcal{O}(vMI_{max}$+$M^2)$ where $M$: \#base clusterings and $v$: \#internal validity indices & Accuracy, ARI, NMI & UCI \newline 20 Newsgroups & Flowchart, line plot, table & \textsc{Sivid} is an ensemble selection algorithm that uses five internal validity indices and \emph{normalized mutual information} to choose a subset of high-quality, diverse base clusterings for the final clustering.\\ \hline

\cite{amiri2018clustering}  & Ensemble at the level of dissimilarity matrices with a subsequent final clustering step  & $\mathcal{O}(0.6J × n^3)$ where $J$: \#attributes, $n$:\#instances   & Classification rate, ARI, Jaccard index & UCI & Line plot, dendrogram, phylogenetic tree, table & \textsc{EN-AL (Ensembled Average Linkage)} and \textsc{EN-CL (Ensembled Complete Linkage)} perform \emph{hierarchical clustering} on bootstrap samples, using \emph{Hamming distance} as the initial dissimilarity measure. They then ensemble the clusterings by aligning them with the Hungarian algorithm and averaging the dissimilarity matrices.
\\ \hline

\cite{yuvaraj2020high}    &  Link-based approach utilizing a weighted triple quality measure   &  Not discussed  & Accuracy, NMI, ARI, precision, recall, cohesion, variance, runtime, scalability & UCI & Flowchart, line plot, table & \textsc{HFALCEF} uses the \emph{Firefly algorithm} to generate initial clusters, partitions data with a bipartite spectral algorithm and multi-viewpoint-based similarity measures, and creates a refined cluster association matrix. An \emph{artificial neural network} then classifies the optimized clusters to produce the final ensemble of data points.  \\ \hline

\cite{karthick2021ensemble}  & Generation of a consensus similarity matrix to produce the final clustering result  & Not discussed  & Accuracy, NMI, ARI& UCI & Line plot, table & \textsc{WSE} creates similarity matrices from different clustering algorithms and merges them into a weighted \emph{consensus matrix}. It uses \emph{swarm-based optimization} to handle empty or duplicate clusters during centroid selection, and \emph{Division Reminder Hashing}to assign cluster weights based on similarity, eliminating null and duplicate datasets.\\ \hline

\end{tabular}
\end{adjustbox}
\end{table*}

\begin{table*}[!htb]
\caption{Comparison of \textsc{Subspace Clustering Algorithms}}
\label{tab:subspace_clustering}
\begin{adjustbox}{max width = 1.1\linewidth}
\renewcommand{\arraystretch}{1.8} 
\setlength{\arrayrulewidth}{0.001pt} 
\begin{tabular}{|p{1.4cm}|p{3.6cm}|p{3.6cm}|p{3.1cm}|p{1.9cm}|p{2.3cm}|p{10cm}|}
\hline

\textbf{Ref.} & \textbf{Cluster representation} & \textbf{Complexity} & \textbf{Evaluation metrics} & \textbf{Data sources} & \textbf{Plots used} & \textbf{Remarks}\\\hline

\cite{ganti1999cactus} & Defines the concept of ``strongly connected" to identify clusters   & Not discussed   & Runtime, scalability & UCI \newline Synthetic & Line plot, table & \textsc{CACTUS} defines clusters as \emph{strongly connected sets} of attribute values that meet specific conditions. It exploits attributes with a small number of values and uses \emph{summary information} (both \emph{inter-attribute} and \emph{intra-attribute} summaries) to identify clusters. \\ \hline

\cite{gan2004subspace}   &  Similar data points described by a specific subset of features & Not discussed   & Accuracy  & UCI  & Table  & \textsc{SUBCAD} starts by choosing the $k$ most dissimilar data points as seeds and assigning other points to the nearest seed. It then iteratively reassigns points to clusters to minimize the first objective function and optimizes the second objective function to determine each cluster's subspace. This approach allows \textsc{SUBCAD} to identify clusters within various subspaces of the original high-dimensional space. \\ \hline

\cite{zaki2005clicks}   & Defines clusters by density and strong connectedness   & Not discussed  & Purity, runtime, scalability & UCI & Confusion matrix, tree plot, line plot, table &  \textsc{Clicks} models the data as a \emph{k}-partite graph, maps clustering to maximal clique enumeration, and uses connectivity ordering and pruning for efficient search. It guarantees the finding of all maximally dense and strongly connected subspace clusters.      \\ \hline

\cite{chen2016soft}  & Probabilistic distribution of categories within each cluster & $\mathcal{O}(KNDT)$ where $K$: \#clusters, $N$: \#instances, $D$: \#attributes, $T$: \#iterations  & MI, accuracy, NMI, runtime, scalability & UCI & Line plot, bar plot, table & \textsc{Scc} uses a \emph{kernel density-based probabilistic distance function} to measure dissimilarity between categorical data objects. It features a \emph{soft feature-selection scheme} that assigns weights to categorical attributes based on \emph{smoothed category dispersion}, optimizing these weights to minimize within-cluster scatter. \textsc{Scc} also uses a cluster validity index to determine the optimal number of clusters.      \\ \hline

\cite{de2016hash}   & Clusters are represented as co-clusters, which are defined as pairs of subsets of objects and features  & Not discussed   & Purity, NMI, pointwise mutual information, runtime, scalability & UCI, 20 Newsgroups, IMDB & Line plot, table & \textsc{HBLCoClust} preprocesses the feature set to eliminate rare features, then uses \emph{Locality Sensitive Hashing} (\textsc{Lsh}) to find promising regions for co-clusters. These regions are explored with the \textsc{InClose} algorithm to extract all co-clusters, which are then merged if they share elements.       \\ \hline

\cite{qin2012novel}   &  Soft set model over equivalence classes  & Not discussed     & Runtime, scalability  & UCI & Bar plot, line plot, table & \textsc{Nss} defines a \emph{soft set model} over equivalence classes in an information system and constructs soft sets to approximate rough sets. It uses these to develop a heuristic for selecting the best clustering attribute by computing dependency degrees among attributes.  \\ \hline

\cite{xu2022multi}   & Kernel methods and self-representation learning to capture the inherent structure of clusters    & Not discussed      &  NMI, accuracy, F-score, Rand index   & Synthetic \newline UCI \newline BBC    & Line plot, scatter plot, density plot, heatmap, table, bar plot   & \textsc{MKCC} constructs kernel matrices for each view using self-representation learning, eliminating the need for explicit kernel functions. It incorporates sample weighting to reduce noise impact, uses an approximation method to lower computational costs, and integrates multiple views for better clustering performance. \\ \hline

\cite{bandyapadhyay2023parameterized}   & Partition of data points and their respective centers in the selected feature space, minimizing the total Hamming distance within each cluster   & $(kB)^{\mathcal{O}(kB)}|\sum|^{kB}n^{\mathcal{O}(k)}m^2$ where $B$: budget parameter, $n$: \#instances, $m$: \#attributes, $|\sum|$: a finite set of non-negative integers   & Not use & Not use & Graph, matrix &  Several objectives are addressed: handling outliers, optimizing feature selection, and improving clustering efficiency. The algorithm uses a parameterized complexity approach to evaluate and select the most relevant features, iteratively removing the least significant ones until an optimal set is obtained. Additionally, it incorporates \emph{hypergraph-based techniques} to manage outliers and ensure robust clustering performance by focusing on the best $k$ clusters.    \\ \hline

\end{tabular}
\end{adjustbox}
\end{table*}

\begin{table*}[!htb]
\vspace{-3.4cm}
\caption{Comparison of \textsc{Data Stream Clustering Algorithms}}
\label{tab:data_stream_clustering}
\begin{adjustbox}{max width = 1.1\linewidth}
\renewcommand{\arraystretch}{1.3} 
\setlength{\arrayrulewidth}{0.001pt} 
\begin{tabular}{|p{1cm}|p{3.6cm}|p{6.1cm}|p{3.1cm}|p{1.9cm}|p{2cm}|p{10cm}|}
\hline

\textbf{Ref.} & \textbf{Sim/Dissim measure} & \textbf{Complexity} & \textbf{Evaluation metrics} & \textbf{Data sources} & \textbf{Plots used} & \textbf{Remarks}\\\hline

\cite{barbara2002coolcat}  & Entropy-based measure   & Not discussed    & Expected entropy, external entropy, category utility, runtime, scalability & UCI \newline Synthetic & Line plot, table &  \textsc{Coolcat} clusters data points by minimizing the \emph{expected entropy} of the clusters, thereby reducing the entropy of the clusters and ``cooling" them. It operates incrementally, allowing new points to be clustered without re-processing the existing set.  \\ \hline

\cite{chen2008catching} & Cosine similarity measure & $\mathcal{O}(qkN+N^2)$, where $q$: \#attributes, $k$: \#clusters, $N$: \#instances  & Precision, recall, category utility, confusion matrix accuracy, runtime, scalability & UCI \newline Datgen.com & Line plot, flowchart, table & \textsc{DCD} identifies when the underlying concept of the data stream has changed, which is referred to as concept drift. It uses the \emph{Node Importance Representative} (\textsc{NIR}) to measure the similarity of new data to existing clusters, compares cluster distributions to detect drift, and decides whether to update the current model or recluster the data.    \\ \hline

\cite{aggarwal2010clustering} & Cosine similarity measure   & Not discussed       & Purity, fraction of new cluster, fraction of true outliers, precision, recall, memory consumption, runtime, scalability & Synthetic & Line plot, table & \textsc{ConStream} summarizes and analyzes large, evolving text and categorical datastreams online, with flexibility for offline analysis using additive droplet summaries over user-defined parameters.     \\ \hline

\cite{cao2010framework} & Simple matching   & DCDA: $\mathcal{O}(|U||A|)$, where $U$: the universe, $|A|$: \#attributes, DLA: $\mathcal{O}(|S^{T_i}||A||S^{T_j}||k^{T_{i}})$, CRAA: $\mathcal{O}(k^{T_{i}} k^{T_{j}} |S^{T_{i}} \cup S^{T_{j}}||A|)$, where $S^{T_i}$: \#instances in the sliding window $T_i$, $A$: \#attributes, $S^{T_j}$: \#instances in the sliding window $T_j$, $k^{T_i}$ and $k^{T_j}$: \#clusters in the sliding window $T_i$ and $T_j$, respectively & Precision, recall, accuracy & UCI & Line plot, table & This framework utilizes a sliding window technique and three main algorithms: \textsc{Dcda} for drifting-concept detection, \textsc{Dla} for data labeling, and \textsc{Craa} for cluster-relationship analysis. \textsc{Dcda} triggers reclustering if the distance between two sliding windows exceeds a specified threshold, capturing recent concepts. \textsc{Dla} assigns objects in the current window to clusters based on similarity to previous results. \textsc{Craa} analyzes clustering over time to identify evolving trends.   \\ \hline

\cite{bai2016optimization}  & Dissimilarity measure that takes into account the representability of each categorical attribute value in the clusters & Time: $\mathcal{O}\left(\sum_{p=1}^{T} k^p t^p |S^p| \left( \sum_{j=1}^{m} n_j \right) \right)$, Space: $O\left(\frac{n}{T} + \sum_{p=1}^{T} k^p \sum_{j=1}^{m} n_j \right)$ where $T$: \#sliding windows, $k^p$: \#clusters in the $p^{th}$ window, $t^p$: \#iterations to cluster data in $p^{th}$ window, $n$: \#instances, $m$: \#attributes, $n_j$: \#categories in $j^{th}$ attribute, $S^p$: $p^{}th$ data subset & Accuracy, precision, recall, Euclidean distance, runtime, scalability & UCI & Line plot, flowchart, table & The proposed model uses a cluster validity function to evaluate and optimize clustering models as new data arrives. It ensures the clustering model's certainty and continuity with previous models. Additionally, it includes a detection index to identify evolving cluster structures, improving the reliability of clustering results over time.      \\ \hline

\end{tabular}
\end{adjustbox}
\end{table*}

\begin{table*}[!htb]
\vspace{-0.4cm}
\caption{Comparison of \textsc{Graph-based Clustering Algorithms}}
\label{tab:graph-based_clustering}
\begin{adjustbox}{max width = 1.1\linewidth}
\renewcommand{\arraystretch}{1.3} 
\setlength{\arrayrulewidth}{0.001pt} 
\begin{tabular}{|p{1cm}|p{6cm}|p{5cm}|p{3cm}|p{1cm}|p{2.3cm}|p{10cm}|}
\hline

\textbf{Ref.} & \textbf{Graph technique} & \textbf{Complexity} & \textbf{Evaluation metrics} & \textbf{Data sources} & \textbf{Plots used} & \textbf{Remarks}\\\hline

\cite{gibson2000clustering}   & Spectral graph partitioning    & Not discussed & Purity, runtime, scalability & IBM & Line plot, table &  \textsc{Stirr} focuses on the patterns of co-occurrence of categorical values within the data and uses an iterative method for assigning and propagating weights on these values to facilitate a type of similarity measure. The technique is inspired by spectral graph partitioning but extends it to handle relational data more complex than graphs, such as hypergraphs, without suffering from combinatorial explosion.     \\ \hline

\cite{david2012spectralcat}  & Uses the eigenstructure of a matrix to cluster points into disjoint groups based on their components in the top singular vectors  &  Not discussed & Calinski-Harabasz index, accuracy & UCI & 3D scatter plot, scatter plot, line plot, flowchart, table  & \textsc{SpectralCAT} transforms high-dimensional numerical and nominal data into categorical values using the \emph{Calinski–Harabasz} index. It then applies \emph{spectral clustering} with dimensionality reduction, identifying geometric patterns and projecting them onto low-dimensional spaces.     \\ \hline

\cite{bai2022categorical} & Constructs a similarity-relational graph $G = \langle V, W \rangle$, where $V$ represents all categorical values and $W$ is the weight matrix of node similarities. Graph embedding techniques are then used to learn node representations & $\mathcal{O}(|V|^2n+nmp+\text{embedding method})$ where $|V|$: \#nodes in the graph, $n$: \#instances, $m$: \#attributes, $p$: embedding dimension. For non embedding: $\mathcal{O}(1)$, spectral embedding: $\mathcal{O}(|V|^2p)$, Non-negative matrix factorization: $\mathcal{O}(|V|^2p+|V|p^2)$, Autoencoder: $\mathcal{O}(|V|^3p)$ & ARI, NMI, runtime & UCI & Scatter plot, table & \textsc{Cdc\_dr} leverages the graph structure of the categorical values to derive suitable numerical representations that capture value similarities, before applying \textsc{K-means} on these integrated numerical representations to cluster the original categorical data points. \\ \hline

\cite{fomin2023parameterized}  &  Leverages properties of bipartite graphs, perfect matching, and the structural simplification of enforcing a forest-like structure in the intersection graph   & $2^{\mathcal{O}(B\log B)}|\sum|^B(nm)^{\mathcal{O}(1)}$ where $B$: budget parameter, $n$: \#instances, $m$: \#attributes, $|\sum|$: a finite alphabet   & None used & None used & Tree plot & The \textsc{Fixed-Parameter Tractable (FPT)} algorithm is a \textsc{Capacitated Clustering} problem that enumerates all potential medians in FPT time, guesses the number of composite clusters and intersecting initial clusters, and uses \emph{color coding} and \emph{dynamic programming} on their intersection graph to find a feasible clustering.
\\ \hline

\end{tabular}
\end{adjustbox}
\end{table*}

\begin{table*}[!htb]
\vspace{-0.4cm}
\caption{Comparison of \textsc{Genetic-based Clustering Algorithms}}
\label{tab:GA_clustering}
\begin{adjustbox}{max width = 1.1\linewidth}
\renewcommand{\arraystretch}{1.3} 
\setlength{\arrayrulewidth}{0.001pt} 
\begin{tabular}{|p{1.2cm}|p{3.8cm}|p{3.6cm}|p{3.1cm}|p{1.9cm}|p{2.3cm}|p{10cm}|}
\hline

\textbf{Ref.} & \textbf{Objective function} & \textbf{Complexity} & \textbf{Evaluation metrics} & \textbf{Data sources} & \textbf{Plots used} & \textbf{Remarks}\\\hline

\cite{mukhopadhyay2009multiobjective} & Global fuzzy compactness of the clusters and fuzzy separation, optimized simultaneously & Not discussed & ARI, t-test, runtime & UCI \newline Datgen.com & VAT, Pareto front plot, table & This multi-objective genetic clustering around medoids algorithm can simultaneously optimize the often conflicting criteria of compactness and separation of clusters. \\ \hline

\cite{deng2010g} &Average normalized mutual information   & Not discussed & Accuracy, clustering error, NMI, runtime, scalability & UCI & Line plot, table & \textsc{G-anmi} uses a \emph{genetic algorithm} to optimize clustering solutions based on the \emph{Average Normalized Mutual Information} (ANMI) objective function. It iteratively improves data partitions through selection, crossover, and mutation.\\     \hline

\cite{yang2015non}  & Fuzzy compactness within a cluster and separation among clusters optimize clustering quality   & Not discussed  & ARI, compactness, fuzzy separation, runtime & UCI & Scatter plot, flowchart, table & \textsc{Nsga-fmc} optimizes fuzzy membership assignments and cluster compactness/separation objectives simultaneously using a multi-objective genetic algorithm approach. \\ \hline

\cite{kuo2019genetic}  & Incorporates intuitionistic fuzzy sets and attribute weighting  & $\mathcal{O}(k^2n^2(2m+M))$ where $k$: \#clusters, $n$: \#instances, $m$: \#attributes, $M$: total \#categories in all attributes   & ARI, accuracy, cluster discrimination index, runtime & UCI & Bar plot, flowchart, table & \textsc{GIWFKM} combines \emph{intuitionistic fuzzy sets} with \textsc{K-modes}, using a \emph{frequency probability-based distance} metric and \emph{weighted attributes}. It employs a \emph{genetic algorithm} (GA) for global optimization and includes unsupervised feature selection based on the correlation coefficient to reduce redundancy and computation time.\\ \hline

\cite{nguyen2019partition}  & Two objective functions: $\pi$ (cluster compactness) and $sep$ (cluster separation) &  $\mathcal{O}(2n^2 + \sqrt{n}-k+n)$ where $n$: \#instances, $k$: \#clusters & ARI, NMI, Davies–Bouldin index, runtime & UCI & Bar plot, flowchart, table & \textsc{PM-FGCA} partitions the dataset into clusters using a fuzzy genetic approach. It then merges clusters based on inter-cluster distances until reaching a specified cluster count. Finally, it improves clustering quality by reassigning instances based on intra-cluster distances. \\ \hline

\cite{jiang2023kernel}  & Intuitionistic fuzzy sets to handle uncertainty, \emph{kernel trick} to estimate distances, and incorporate attribute weights to address varying attribute importances  & $\mathcal{O}(k(2n)(2m+M))$ where $k$: \#clusters, $n$: \#instances, $m$: \#attributes, $M$: total \#categories in all attributes  & ARI, NMI, accuracy, F-score & UCI & Bar plot, flowchart, table & \textsc{KIWFKM-DCP} combines \emph{kernel tricks} and \emph{intuitionistic fuzzy sets} in the \textsc{Weighted Fuzzy K-modes} algorithm. The coupled DCP system employs selection, crossover, and mutation rules from DNA genetic rules to enhance clustering, avoid local optima and improve robustness to noise and attribute importance. \\ \hline

\end{tabular}
\end{adjustbox}
\end{table*}
\restoregeometry
\section{A Taxonomy of categorical data clustering algorithms} \label{sec:taxonomy}
\subsection{Based on clustering methods}
In this subsection, we provide a detailed taxonomy of categorical data clustering algorithms, classifying them based on their underlying approaches. This classification spans various methodologies developed over the past 25 years, showcasing the diverse techniques and their characteristics. Algorithms are organized by the year of publication within each clustering type. Tables \ref{tab:partitional_clustering_hard_1}, \ref{tab:partitional_clustering_hard_2}, and \ref{tab:partitional_clustering_hard_3} detail hard partitional clustering algorithms. They highlight key characteristics, including the similarity or dissimilarity measures, computational complexity, evaluation metrics, data sources, visualization types, and notable remarks. This information is essential for understanding each algorithm's operation and performance.

\begin{figure*}[!htb]
\vspace{-2cm}
  \centering
  \includegraphics[width=0.8\linewidth]{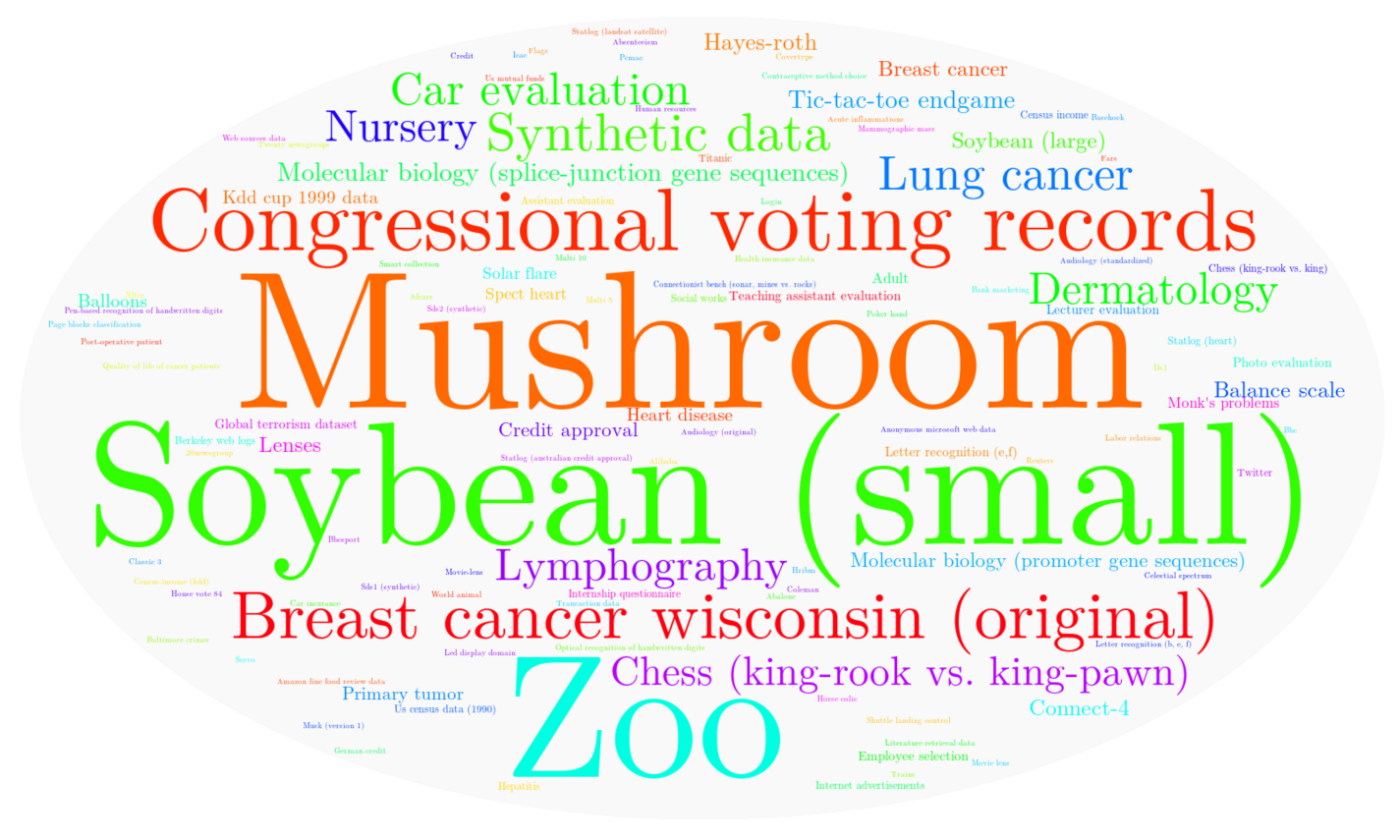}
  \caption{Common categorical datasets used for experiments in the literature}
  \label{fig:wordcloud}
\end{figure*}

Table \ref{tab:partitional_clustering_fuzzy} compares various fuzzy partitional clustering algorithms. It highlights each algorithm's fuzzification parameter, which controls the degree of cluster membership, and provides information on their computational complexity. The table also lists the evaluation metrics used to assess performance, along with details about the data types and sources. Visualization techniques employed to illustrate the clustering results are included, and the remarks column provides additional insights and observations specific to each algorithm.

Table \ref{tab:partitional_hierarchical_clustering} compares hierarchical clustering algorithms, detailing whether they are agglomerative or divisive and their computational complexity. It also includes evaluation metrics for performance assessment, information on data sources and types, visualization methods used, and notable comments or findings in the remarks section.

Table \ref{tab:model-based_clustering} offers a comprehensive comparison of model-based clustering algorithms. It details the methods used for estimating model parameters, the computational complexity of each algorithm, and the evaluation metrics for performance assessment. The table also includes information on data types and sources, visualization methods, and significant observations in the remarks section.

Table \ref{tab:ensemble_clustering} compares ensemble clustering algorithms, detailing the integration strategies used to combine clustering results, their computational complexity, and the evaluation metrics for performance assessment. It also provides information on data types and sources, visualization techniques employed, and remarks that offer insights into each algorithm's strengths and limitations.

Table \ref{tab:subspace_clustering} compares subspace clustering algorithms, highlighting the cluster representation used, computational complexity, and evaluation metrics for performance assessment. It also includes information on data types and sources, visualization techniques, and key findings or observations in the remarks section.

Table \ref{tab:data_stream_clustering} compares data stream clustering algorithms, specifying the similarity or dissimilarity measures used, their computational complexity, and the evaluation metrics. It also includes information on data types and sources, visualization plots, and additional notes or findings relevant to each algorithm.

Table \ref{tab:graph-based_clustering} compares graph-based clustering algorithms, detailing the graph techniques used, computational complexity, and evaluation metrics for performance assessment. It also includes information on data types and sources, visualization methods, and key comments or insights in the remarks section.

Table \ref{tab:GA_clustering} compares genetic-based clustering algorithms, highlighting the objective functions optimized by each algorithm, their computational complexity, and the evaluation metrics used for performance assessment. It also provides information on data types and sources, visualization techniques, and notable observations or findings in the remarks section.
\subsection{Based on categorical datasets and  data sources} \label{sec:datasets}

\begin{figure*}[!htb]
  \centering
  \includegraphics[width=0.8\linewidth]{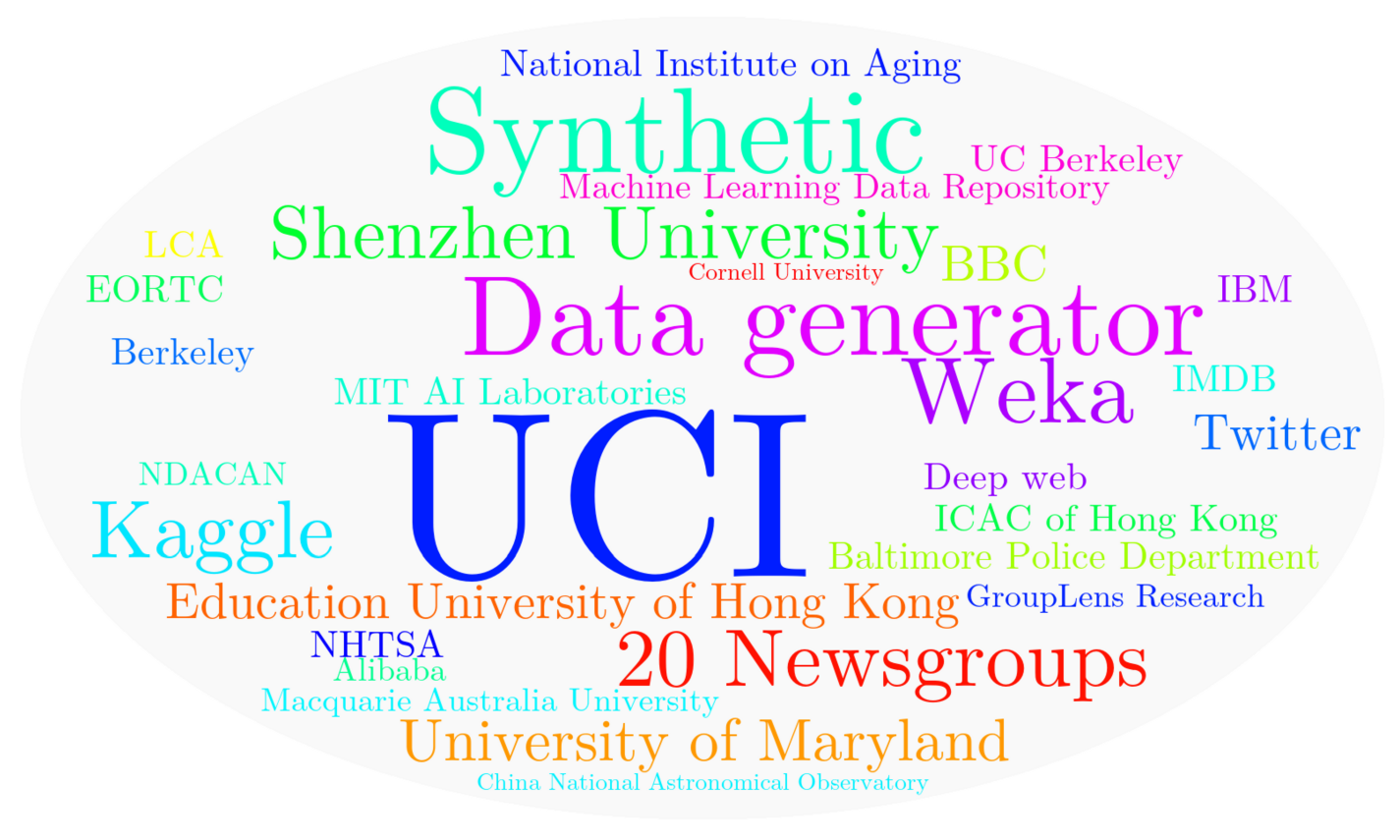}
  \caption{Sources of categorical datasets used in the experiments}
  \label{fig:wordcloud_sources}
\end{figure*}

\begin{table*} [!htb]
\centering
\caption {Characteristics of common categorical datasets from UCI (sorted by number of objects)}
\begin{adjustbox}{max width = \linewidth}
\renewcommand{\arraystretch}{1} 
\begin{tabular}{|p{4cm}|c|c|c||p{4cm}|c|c|c||p{4.5cm}|c|c|c|} \hline
\textbf{Dataset} & \textbf{\#Objs} & \textbf{\#Attrs} & \textbf{\#Classes} & \textbf{Dataset} & \textbf{\#Objs} & \textbf{\#Attrs} & \textbf{\#Classes} &\textbf{Dataset} & \textbf{\#Objs} & \textbf{\#Attrs} & \textbf{\#Classes} 
\\ \hline \hline
Balloons [\href{https://archive.ics.uci.edu/dataset/13/balloons}{Link}] & 16 & 4 & 2  & Primary Tumor [\href{https://archive.ics.uci.edu/dataset/83/primary+tumor}{Link}] & 339& 17 & 21  & Chess (KRKPA7) [\href{https://archive.ics.uci.edu/dataset/22/chess+king+rook+vs+king+pawn}{Link}] & 3,196 & 36 & 2\\ \hline
Lenses  [\href{https://archive.ics.uci.edu/dataset/58/lenses}{Link}]  & 24 & 4 & 3  & Dermatology [\href{https://archive.ics.uci.edu/dataset/33/dermatology}{Link}] & 366 & 34 & 6 & Mushroom [\href{https://archive.ics.uci.edu/dataset/73/mushroom}{Link}]  & 8,124 & 22 & 2 \\ \hline
Lung cancer [\href{https://archive.ics.uci.edu/dataset/62/lung+cancer}{Link}] & 32 & 56 & 3  & Voting records [\href{https://archive.ics.uci.edu/dataset/105/congressional+voting+records}{Link}] & 435 & 16 & 2 & Nursery [\href{https://archive.ics.uci.edu/dataset/76/nursery}{Link}] & 12,960 & 8 & 5\\ \hline
Soybean small [\href{https://archive.ics.uci.edu/dataset/91/soybean+small}{Link}] & 47 & 35 & 4 & Balance scale & 625 & 4 & 3  & Chess (KRK) [\href{https://archive.ics.uci.edu/dataset/23/chess+king+rook+vs+king}{Link}] & 28,056   & 6   & 18   \\ \hline
Zoo [\href{https://archive.ics.uci.edu/dataset/111/zoo}{Link}] & 101 & 16 & 7 & Soybean large [\href{https://archive.ics.uci.edu/dataset/90/soybean+large}{Link}] & 683 & 35 & 19 & Secondary mushroom [\href{https://archive.ics.uci.edu/dataset/848/secondary+mushroom+dataset}{Link}] & 61,068 & 17 & 2\\ \hline
Promoter gene sequences [\href{https://archive.ics.uci.edu/dataset/67/molecular+biology+promoter+gene+sequences}{Link}] & 106 & 57 & 2  & Breast cancer Wisconsin (original) [\href{https://archive.ics.uci.edu/dataset/15/breast+cancer+wisconsin+original}{Link}]& 699 & 9 & 2  & Connect-4 [\href{https://archive.ics.uci.edu/dataset/26/connect+4}{Link}] & 67,557 & 42 & 3\\ \hline 
Lymphography  [\href{https://archive.ics.uci.edu/dataset/63/lymphography}{Link}] & 148 & 18 & 4 & NPHA [\href{https://archive.ics.uci.edu/dataset/936/national+poll+on+healthy+aging+(npha)}{Link}]& 714 & 14 & 3  & CDC diabetes [\href{https://archive.ics.uci.edu/dataset/891/cdc+diabetes+health+indicators}{Link}] &253,680  & 21  & 2  \\ \hline
Hayes-Roth [\href{https://archive.ics.uci.edu/dataset/44/hayes+roth}{Link}] & 160 & 4 & 3 & Tic-tac-toe Endgame [\href{https://archive.ics.uci.edu/dataset/101/tic+tac+toe+endgame}{Link}] & 958 & 9 & 2 & Census-Income (KDD) [\href{https://archive.ics.uci.edu/dataset/117/census+income+kdd}{Link}] & 299,285 &41 & 2 \\ \hline
SPECT Heart [\href{https://archive.ics.uci.edu/dataset/95/spect+heart}{Link}] & 267 &22 & 2  & Car evaluation [\href{https://archive.ics.uci.edu/dataset/19/car+evaluation}{Link}] & 1,728 & 6 & 4 & Covertype [\href{https://archive.ics.uci.edu/dataset/31/covertype}{Link}] & 581,012 & 54 & 7\\ \hline
Breast cancer [\href{https://archive.ics.uci.edu/dataset/14/breast+cancer}{Link}] & 286& 9 & 2  & Splice-junction DNA [\href{https://archive.ics.uci.edu/dataset/69/molecular+biology+splice+junction+gene+sequences}{Link}] & 3,190 & 60 & 3 &  Poker Hand [\href{https://archive.ics.uci.edu/dataset/158/poker+hand}{Link}] & 1,025,010 & 10 & 10\\ \hline
\end{tabular}
\end{adjustbox}
\label{Table:dataset_stats} 
\end{table*}

In this subsection, we summarize the categorical datasets frequently used in experiments and their sources. This classification aids in understanding the variety of datasets employed in evaluating these algorithms. Figure \ref{fig:wordcloud} displays a word cloud of the datasets used in 102 papers surveyed. Datasets utilized more than five times are highlighted to underscore their importance in categorical data clustering research.

Among the most frequently used datasets is the \textsc{Mushroom} dataset, which has been used $56$ times. It contains data points of mushroom species with categorical features describing physical characteristics. The \textsc{Soybean (Small)} dataset (utilized $51$ times), categorizes different types of soybean diseases based on various attributes. The \textsc{Zoo} dataset (utilized $51$ times) classifies animals in a zoo into categories based on their characteristics. The \textsc{Congressional Voting Records} dataset (utilized $45$ times) contains voting records of U.S. Congress members categorized by different votes. The \textsc{Breast Cancer Wisconsin (Original)} dataset (utilized $36$ times)  includes attributes related to breast cancer cases.

Other notable datasets include the \textsc{Synthetic} dataset (utilized $22$ times), which is artificially generated to test the performance of clustering algorithms; \textsc{Lung Cancer} (utilized $17$ times), containing features related to lung cancer patients; and the \textsc{Car Evaluation} dataset (utilized $16$ times), which categorizes cars based on various evaluation criteria. The \textsc{Dermatology} dataset (utilized $15$ times) includes data points of dermatological diseases, while the \textsc{Nursery} dataset (utilized $15$ times) evaluates clustering algorithms based on nursery application attributes. The \textsc{Lymphography} dataset (utilized $15$ times) contains lymphography exam results, which help in diagnosing different lymph node conditions.

The \textsc{Chess (King-Rook vs. King-Pawn)} dataset (utilized $14$ times) includes endgame situations in chess. The \textsc{Molecular Biology (Splice-junction Gene Sequences)} dataset (utilized $13$ times) contains gene sequence data for biological clustering experiments. The \textsc{Tic-Tac-Toe Endgame} dataset (utilized $12$ times) includes endgame positions to test clustering algorithms in recognizing tic-tac-toe outcomes. The \textsc{Hayes-Roth} dataset (utilized $10$ times) includes family relationship attributes. The \textsc{Connect-4} dataset (utilized $9$ times) consists of game states for testing pattern recognition. The \textsc{Balance Scale} dataset (utilized $9$ times) categorizes different balance scale configurations, while the \textsc{Soybean (Large)} dataset (utilized $9$ times) is a larger version of the soybean dataset, providing more data for clustering experiments.

Further, the \textsc{Lenses} dataset (utilized $8$ times) includes attributes related to lens prescription, while the \textsc{Breast Cancer (Ljubljana)} dataset (utilized $8$ times) predicts the recurrence of breast cancer based on various patient and tumor attributes. The \textsc{Credit Approval} dataset (utilized $7$ times), contains data points of credit approval decisions, while the \textsc{Molecular Biology (Promoter Gene Sequences)} dataset (utilized $7$ times) is another gene sequence dataset. Finally, the \textsc{KDD Cup 1999 Data} and \textsc{Primary Tumor} datasets, along with the \textsc{Heart Disease}, \textsc{Balloons}, and \textsc{SPECT Heart} datasets, have each been used $6$ times. These datasets provide varied sources of data for clustering algorithm evaluation.

Figure \ref{fig:wordcloud_sources} shows the data sources where categorical datasets come from. The most common source for categorical datasets is the \textsc{UCI Machine Learning Repository} \citep{kelly2023uci}, with a significant frequency of $370$ times. \textsc{Synthetic} data sources are also notable, used $9$ times, providing controlled conditions for algorithm testing. \textsc{Data Generators} \citep{gabor1999datgen} are used $7$ times, offering customized dataset creation. The \textsc{20 Newsgroups dataset} and \textsc{Weka repository} \citep{h2017data} have both been utilized $5$ times, while \textsc{Shenzhen University} and \textsc{Kaggle} each contribute $4$ datasets. These sources provide a broad spectrum of categorical data, facilitating diverse and comprehensive evaluations of clustering algorithms.

Table \ref{Table:dataset_stats} presents a list of categorical datasets from the UCI repository. These datasets vary in characteristics, encompassing combinations such as \emph{small size + low dimensionality}, \emph{small size + high dimensionality}, \emph{large size + low dimensionality}, and \emph{large size + high dimensionality}. By focusing on these widely used datasets, we emphasize their importance in the field of categorical data clustering and provide a benchmark for evaluating the performance of various clustering algorithms.
\subsection{Based on data visualization techniques}
\begin{figure*}[!htb]
 \vspace{-2.4cm}
  \centering
  \includegraphics[width=\linewidth]{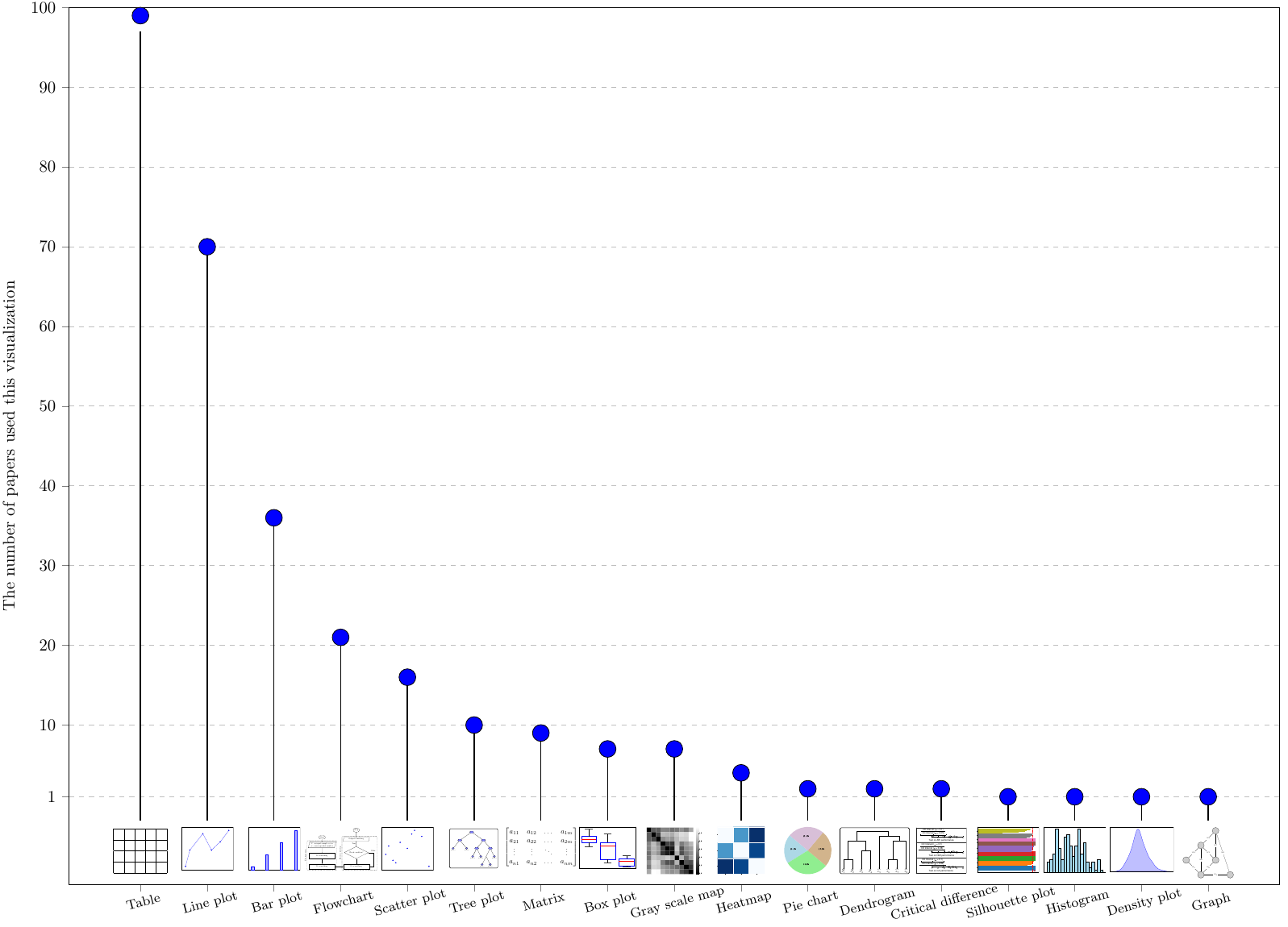}
  \vspace{-0.5cm}
  \caption{Frequency of \textsc{Visualization Techniques} used in categorical data clustering}
  \label{fig:viz_stats}
\end{figure*}

In this subsection, we explore the various data visualization techniques commonly used in the literature to present clustering results. These techniques help researchers and practitioners interpret and analyze the performance and characteristics of clustering algorithms effectively. Figure \ref{fig:viz_stats} shows frequently used visualization techniques in categorical data clustering.

The most prevalent visualization technique is the \textsc{Table}, used $99$ times, which provides a structured way to display detailed information about datasets and clustering results. \textsc{Line Plots}, employed $70$ times, are useful for showing trends and comparing clustering results over different evaluation metrics. \textsc{Bar Plots}, used $36$ times, offer a clear comparison of categorical data by representing data visually with rectangular bars. \textsc{Flow charts}, utilized $21$ times, help in illustrating the steps and processes involved in clustering algorithms.

\textsc{Scatter Plots}, used $16$ times, are effective for visualizing the relationships between two variables, revealing patterns and correlations. \textsc{Tree Plots}, appearing in $10$ paper, are particularly useful for hierarchical data and clustering trees. \textsc{Matrix Visualizations}, used $9$ times, provide a grid-like representation to highlight relationships and interactions within the data. \textsc{Box Plots}, used seven times, display the range of specific evaluation metrics, such as ARI, produced by different algorithms. They indicate the maximum value, minimum value, median value, and outliers for each algorithm.

Less frequently used techniques include \textsc{VAT (Visual Assessment of cluster Tendency)} and \textsc{Heatmaps}, both used $4$ times, which offer a visual representation of the distance or similarity between data points. \textsc{Gray Scale Maps}, used $3$ times, provide a simple way to display variations in category value frequency. \textsc{Pie Charts} and \textsc{Dendrograms}, each used twice, effectively display proportions and hierarchical relationships, and aid in selecting the number of clusters, respectively. The \textsc{Critical Difference Diagram}, also used twice, is a statistical visualization tool that highlights significant differences between methods.

Finally, several visualization techniques are rarely used in the included papers, i.e. only once each; nevertheless, they can be highly informative: \textsc{Silhouette Plots}, which display how close each point is to points in its own cluster compared to points in neighboring clusters; \textsc{Histogram Plots}, which show the distribution of a single variable; \textsc{Density Plots}, which estimate the probability density function of a variable; and \textsc{Graphs}, which illustrate the network relationships between data points.

These visualization techniques collectively enhance the understanding of categorical data clustering by providing multiple perspectives and insights into the data and the performance of clustering algorithms.

\subsection{Based on validation metrics} \label{sec:evaluation}

This subsection explores various validation metrics used to evaluate categorical data clustering algorithms, crucial for assessing their performance and effectiveness. Validation metrics are categorized into internal and external types, each serving distinct purposes in the evaluation process.

\subsubsection{Internal validation metrics}
Internal validation metrics assess the quality of clustering algorithms by evaluating their performance based solely on the data used for clustering, without external references. This section delves into different internal metrics and their significance in categorical data clustering analysis.

Figure \ref{fig:internal_metrics} illustrates the frequency of various internal validation metrics used to evaluate algorithm performance. These metrics are grouped based on similar characteristics, including cluster quality, compactness and separation, distance and variability, and information criteria.

Table \ref{tab:internal_validation_metrics} defines and explains several internal validation metrics used in at least two of the 102 surveyed papers. This table serves as a  resource for understanding frequently employed metrics and their significance in categorical clustering analysis.

\begin{figure*}[!htb]
  \centering
  \includegraphics[width=0.9\linewidth]{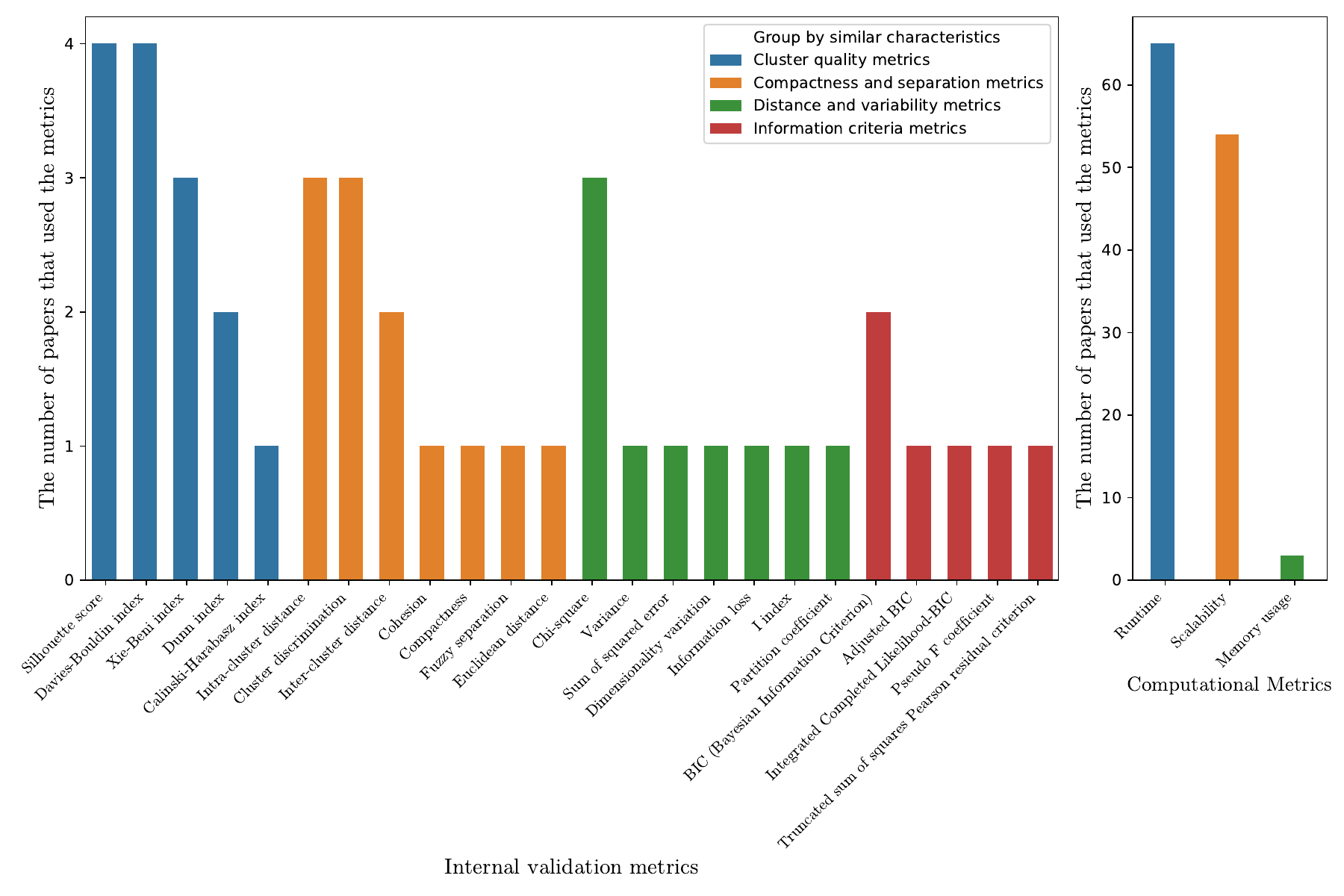}
  \caption{Frequency of \textsc{Internal Validation Metrics} used in categorical data clustering}
  \label{fig:internal_metrics}
\end{figure*}
\begin{table*}[!htb]
\vspace{-2cm}
\caption{Common \textsc{Internal Validation Metrics} for Categorical Data Clustering}
\label{tab:internal_validation_metrics}
\begin{adjustbox}{max width = \linewidth}
\renewcommand{\arraystretch}{1.3} 
\begin{tabular}{|>{\centering\arraybackslash}p{3cm}|>{\centering\arraybackslash}p{3.2cm}|>{\centering\arraybackslash}p{5.5cm}|>{\centering\arraybackslash}p{5.5cm}|>{\centering\arraybackslash}p{2.5cm}|}
\hline
\textbf{Metric} & \textbf{Formulation} & \textbf{Notation explanation} & \textbf{Meaning} & \textbf{Range and Optimal value} \\\hline

\multicolumn{1}{|>{\raggedright\arraybackslash}p{3.2cm}|}{ \vspace{0.4cm} \textsc{Silhouette Score}} & \vspace{-1cm} \[\frac{b(i) - a(i)}{\max\{a(i), b(i)\}}\] & \multicolumn{1}{>{}p{5.8cm}|} {$a(i)$: average distance from $i$ to other points in the same cluster, $b(i)$: lowest average distance from $i$ to points in a different cluster} & \multicolumn{1}{>{}p{5.5cm}|} {Measures how similar an object is to its own cluster compared to other clusters} & [-1, 1] \newline Higher is better \\\hline

\multicolumn{1}{|>{\raggedright\arraybackslash}p{3.3cm}|}{\vspace{0.4cm} \textsc{Davies-Bouldin Index}} & \vspace{-1cm} \[\frac{1}{n} \sum_{i=1}^{n} \max_{j \neq i} \left( \frac{\sigma_i + \sigma_j}{d_{ij}} \right)\] \vspace{-0.5cm} & \multicolumn{1}{>{}p{5.8cm}|}{$\sigma_i$: average distance of all points in cluster $i$ to centroid of cluster $i$, $d_{ij}$: distance between centroids of clusters $i$ and $j$} & \multicolumn{1}{>{}p{5.5cm}|} {Indicates the average similarity ratio of each cluster with the cluster most similar to it} & [0, $\infty$] \newline Lower is better \\\hline

\multicolumn{1}{|>{\raggedright\arraybackslash}p{3.2cm}|}{\vspace{0.1cm} \textsc{Dunn Index}} & \vspace{-1cm} \[\frac{\min_{i \neq j} d(C_i, C_j)}{\max_k \delta(C_k)}\] \vspace{-0.5cm} & \multicolumn{1}{>{}p{5.8cm}|}{ $d(C_i, C_j)$: inter-cluster distance between clusters $i$ and $j$, $\delta(C_k)$: intra-cluster distance of cluster $k$} & \multicolumn{1}{>{}p{5.5cm}|} {Identifies dense and well-separated clusters} & [0, $\infty$] \newline Higher is better\\\hline

\multicolumn{1}{|>{\raggedright\arraybackslash}p{3.2cm}|}{\vspace{0.4cm} \textsc{Xie-Beni Index}} & \vspace{-1cm} \[\frac{\sum_{i=1}^{n} \sum_{j=1}^{c} \mu_{ij}^2 d^2(x_i, v_j)}{n \min_{i \neq j} d^2(v_i, v_j)}\] \vspace{-0.5cm}& \multicolumn{1}{>{}p{5.8cm}|}{$\mu_{ij}$: membership degree of $x_i$ in cluster $j$, $d(x_i, v_j)$: distance between $x_i$ and cluster center $v_j$, $n$: \#instances, $c$: \#clusters } & \multicolumn{1}{>{}p{5.5cm}|} {Measures compactness and separation of clusters} & [0, $\infty$] \newline Lower is better \\\hline

\multicolumn{1}{|>{\raggedright\arraybackslash}p{3.2cm}|}{\textsc{Intra-Cluster Distance}} & \vspace{-1cm} \[\frac{\sum_{i=1}^{u} \sum_{j=1}^{u} d(x_i, x_j)}{u^2}\] \vspace{-0.5cm} & \multicolumn{1}{>{}p{5.8cm}|}{$u$: \#instances in the cluster, $d(x_i, x_j)$: distance between $x_i$ and $x_j$ in the cluster} & \multicolumn{1}{>{}p{5.5cm}|}{Quantifies the average distance between all pairs of points within the same cluster} & [0, $\infty$] \newline Lower is better \\\hline

\multicolumn{1}{|>{\raggedright\arraybackslash}p{3.2cm}|}{\vspace{0.05cm} \textsc{Inter-Cluster Distance}} & \vspace{-1cm} \[\frac{1}{n_s n_t} \sum_{x_i \in C_s} \sum_{x_j \in C_t} d(x_i, x_j)\] \vspace{-0.5cm} & \multicolumn{1}{>{}p{5.8cm}|}{$n_s$: \#instances in cluster $C_s$, $n_t$: \#instances in cluster $C_t$, $d(x_i, x_j)$: distance between $x_i$ and $x_j$} & \multicolumn{1}{>{}p{5.5cm}|}{Quantifies the average distance between points in different clusters} &  [0, $\infty$] \newline Higher is better \\\hline

\multicolumn{1}{|>{\raggedright\arraybackslash}p{3.2cm}|}{\vspace{0.2cm} \textsc{Cluster Discrimination}} & \vspace{-1cm} \[\frac{d_{\text{min}}}{d_{\text{max}}}\] & \multicolumn{1}{>{}p{5.8cm}|}{ $d_{\text{min}}$: minimum averaged inter-cluster distance, $d_{\text{max}}$: maximum averaged intra-cluster distance} & \multicolumn{1}{>{}p{5.5cm}|} {Ratio of minimum averaged inter-cluster distance to maximum averaged intra-cluster distance, indicating the separation and compactness of clusters} & [0, $\infty$] \newline Higher is better\\\hline

\multicolumn{1}{|>{\raggedright\arraybackslash}p{3.2cm}|}{\vspace{0.01cm}\textsc{Chi-Squared}} &\vspace{-1cm} \[\sum \frac{(O - \mathbb{E})^2}{\mathbb{E}}\] \vspace{-0.5cm} & \multicolumn{1}{>{}p{5.8cm}|}{$O$: Observed frequency of data points in clusters, $\mathbb{E}$: Expected frequency of data points in clusters} & \multicolumn{1}{>{}p{5.5cm}|}{Measures the goodness of fit between the observed clustering and the expected distribution} & [0, $\infty$] \newline Lower is better \\\hline

\multicolumn{1}{|>{\raggedright\arraybackslash}p{3.2cm}|}{\textsc{Bayesian Information Criterion (BIC)}} & \vspace{-1cm} \[k\ln(n)-2\ln(L)\] \vspace{-0.5cm} & \multicolumn{1}{>{}p{5.8cm}|}{$L$: Likelihood of the clustering model, $k$: \# clusters and parameters, $n$: \#instances} & \multicolumn{1}{>{}p{5.5cm}|}{Evaluates the trade-off between model fit and clustering complexity} & $(-\infty, \infty)$ \newline Lower is better \\\hline

\end{tabular}
\end{adjustbox}
\end{table*}

\subsubsection{External validation metrics}
External validation metrics compare clustering results to an external ground truth or reference standard. Figure \ref{fig:external_metrics} shows the usage frequency of different external validation metrics, categorized into cluster quality, information-theoretic, and statistical significance and error metrics.

Table \ref{tab:external_validation_metrics} provides definitions and explanations of several external validation metrics used in at least two of the 102 surveyed papers. This table serves as a reference for understanding commonly used external metrics and their relevance in categorical clustering.

\begin{figure*}[!htb]
\vspace{-0.2cm}
  \centering
  \includegraphics[width=0.9\linewidth]{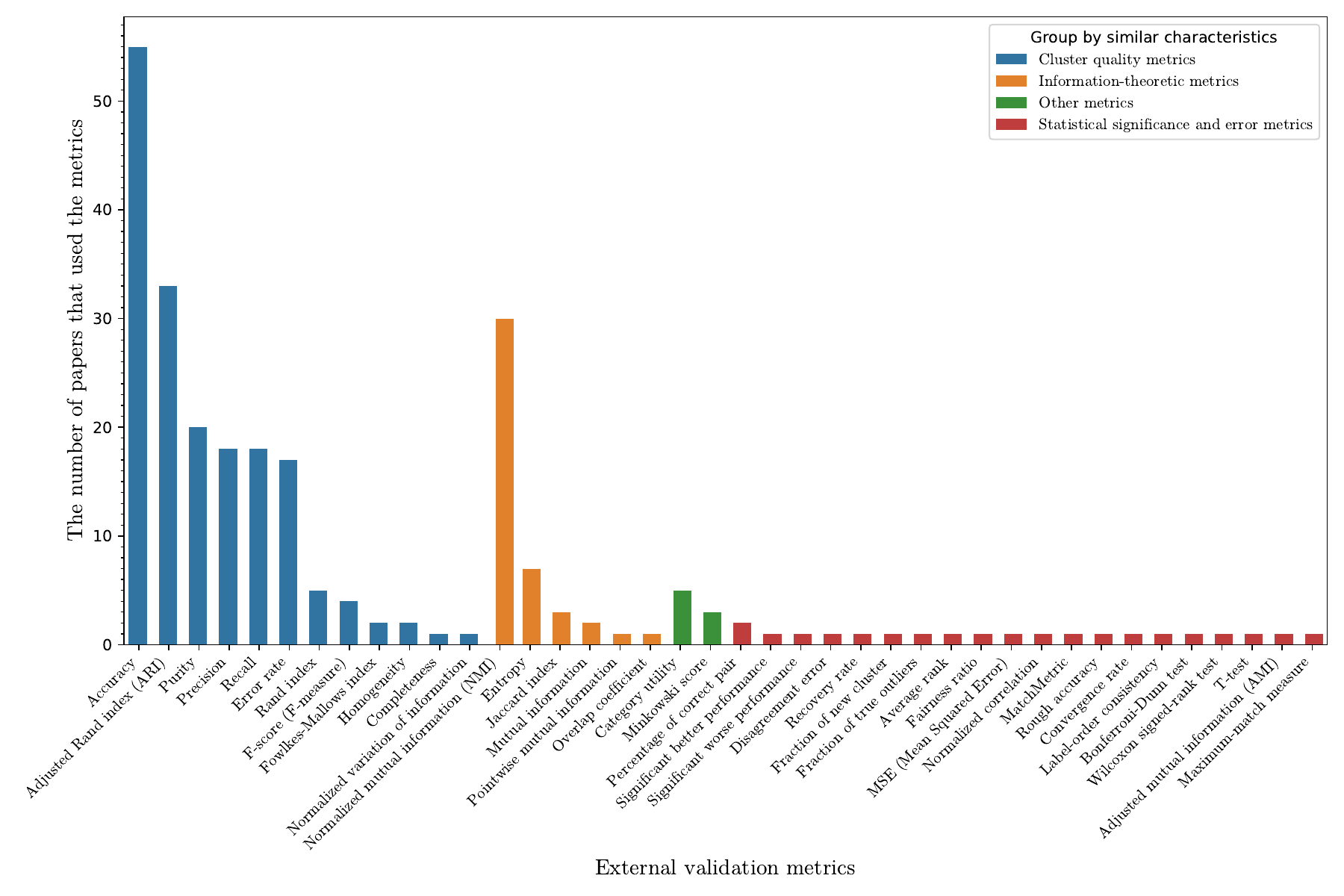}
  \caption{Frequency of \textsc{External Validation Metrics} used in categorical data clustering}
  \label{fig:external_metrics}
\end{figure*}

\begin{table*}[!htb]
\vspace{-2.5cm}
\caption{Common \textsc{External Validation Metrics} for Categorical Data Clustering}
\label{tab:external_validation_metrics}
\begin{adjustbox}{max width = \linewidth}
\begin{tabular}{|>{\centering\arraybackslash}p{3.2cm}|>{\centering\arraybackslash}p{5.3cm}|>{\centering\arraybackslash}p{5.2cm}|>{\centering\arraybackslash}p{5.5cm}|>{\centering\arraybackslash}p{2.5cm}|}
\hline
\textbf{Metric} & \textbf{Formulation} & \textbf{Notation explanation} & \textbf{Meaning} & \textbf{Range and Optimal value} \\\hline

\multicolumn{1}{|>{\raggedright}p{3.3cm}|}{\vspace{0.2cm} \textsc{Accuracy}} & \vspace{-0.8cm}\[\frac{\sum_{l=1}^{k}a_l}{n}\] \vspace{-0.5cm} & \multicolumn{1}{>{}p{5.2cm}|}{$a_l$: numbers of (\#)instances correctly assigned to class $C_l$, $n$: total \#instances} & \multicolumn{1}{>{}p{5.5cm}|}{Measure how accurately the clustering matches the ground truth.} & [0, 1] \newline Higher is better \\\hline

\multicolumn{1}{|>{\raggedright}p{3.3cm}|}{\vspace{0.05cm} \textsc{Adjusted Rand Index (ARI)}} &\vspace{-0.8cm} \[\frac{\text{RI} - \mathbb{E}[\text{RI}]}{\max(\text{RI}) - \mathbb{E}[\text{RI}]}\] \vspace{-0.5cm}& \multicolumn{1}{>{}p{5.2cm}|}{$\text{RI}$: Rand index (as defined below), $\mathbb{E}[\text{RI}]$: expected Rand index} & \multicolumn{1}{>{}p{5.5cm}|}{Adjusts the Rand index for chance, measuring clustering similarity while accounting for random agreement.} & [-1, 1] \newline Higher is better \\\hline

\multicolumn{1}{|>{\raggedright}p{3.3cm}|}{\vspace{0.2cm} \textsc{Normalized Mutual Information (NMI)}} & \vspace{-0.8cm} \[\frac{I(U;V)}{\sqrt{H(U)H(V)}}\] & \multicolumn{1}{>{}p{5.2cm}|}{$U$: set of clusters from the first clustering, $V$: set of clusters from the second clustering, $I(U;V)$: mutual information between $U$ and $V$, $H(U)$: entropy of $U$, $H(V)$: entropy of $V$} & \multicolumn{1}{>{}p{5.5cm}|}{Measure the amount of information shared between two clusterings, normalized to account for the size of the clusters.} & [0, 1] \newline Higher is better \\\hline

\multicolumn{1}{|>{\raggedright}p{3.3cm}|}{\vspace{0.05cm} \textsc{Purity}} & \vspace{-0.8cm} \[\frac{1}{n} \sum_i \max_j |C_i \cap T_j|\] \vspace{-0.5cm}& \multicolumn{1}{>{}p{5.2cm}|}{$C_i$: cluster $i$, $T_j$: class $j$, $n$: \#instances} & \multicolumn{1}{>{}p{5.5cm}|}{Measure the extent to which clusters contain a single class.} & [0, 1] \newline Higher is better \\\hline

\multicolumn{1}{|>{\raggedright}p{3.3cm}|}{\vspace{0.05cm} \textsc{Precision}} & \vspace{-0.8cm} \[\frac{\sum_{l=1}^{k}(\frac{a_l}{a_l+b_l})}{k}\] \vspace{-0.5cm} & \multicolumn{1}{>{}p{5.2cm}|}{$a_l$, $b_l$: \#instances correctly and incorrectly assigned to class $C_l$, respectively} & \multicolumn{1}{>{}p{5.5cm}|}{Measure the fraction of correctly assigned instances among all instances assigned to the same cluster.} & [0, 1] \newline Higher is better \\\hline

\multicolumn{1}{|>{\raggedright}p{3.3cm}|}{\vspace{0.05cm} \textsc{Recall}} & \vspace{-0.8cm} \[\frac{\sum_{l=1}^{k}(\frac{a_l}{a_l+c_l})}{k}\] \vspace{-0.5cm} & \multicolumn{1}{>{}p{5.2cm}|}{$a_l$: \#instances correctly assigned to class $C_l$, $c_l$: \#instances that are incorrectly rejected from class $C_l$} & \multicolumn{1}{>{}p{5.5cm}|}{Measure the fraction of correctly assigned instances among all true instances in the same cluster.} & [0, 1] \newline Higher is better \\\hline

\multicolumn{1}{|>{\raggedright}p{3.3cm}|}{\vspace{0.05cm} \textsc{Entropy}} & \vspace{-0.8cm} \[- \sum_{k} \frac{|C_k|}{n} \log \frac{|C_k|}{n} \] \vspace{-0.5cm} & \multicolumn{1}{>{}p{5.2cm}|}{$|c_k|$: \#instances in cluster $C_k$, $n$: total \#instances} & \multicolumn{1}{>{}p{5.5cm}|}{Measure the uncertainty in the clustering. A lower value indicates a better clustering in terms of purity, as it signifies less uncertainty.} & [0, $\infty$] \newline Lower is better \\\hline

\multicolumn{1}{|>{\raggedright}p{3.3cm}|}{\vspace{0.05cm} \textsc{Rand Index}} & 
\vspace{-0.8cm} \[\frac{a + d}{a + b + c + d}\] \vspace{-0.5cm} & 
\multicolumn{1}{>{\raggedright}p{5.2cm}|}{$a$, $d$: \#pairs correctly assigned to the same cluster (True Positives) and different clusters (True Negatives) $b$, $c$: \#pairs incorrectly assigned to the same cluster (False Positives) and different clusters (False Negatives)} & 
\multicolumn{1}{>{\raggedright}p{5.5cm}|}{Measure the similarity between two clusterings by considering all pairs of instances and counting the pairs that are correctly clustered together or correctly separated.} & 
[0, 1] \newline Higher is better \\\hline

\multicolumn{1}{|>{\raggedright}p{3.3cm}|}{\vspace{0.05cm} \textsc{F-score (F-measure)}} & \vspace{-0.8cm} \[\frac{2 \cdot \textsc{Precision} \cdot \textsc{Recall}}{\textsc{Precision} + \textsc{Recall}}\]  \vspace{-0.5cm} & \multicolumn{1}{>{}p{5.2cm}|}{\textsc{Precision} and \textsc{Recall} as defined above} & \multicolumn{1}{>{}p{5.5cm}|}{Combines precision and recall into a single metric.} & [0, 1] \newline Higher is better \\\hline

\multicolumn{1}{|>{\raggedright}p{3.3cm}|}{\vspace{0.05cm} \textsc{Jaccard Index}} & \vspace{-0.8cm} \[\frac{a}{a + b + c}\]\vspace{-0.5cm} & \multicolumn{1}{>{}p{5.2cm}|}{$a$: true positives, $b$: false positives, $c$: false negatives} & \multicolumn{1}{>{}p{5.5cm}|}{Measures the similarity between the clustering and the ground truth by comparing the sets of correctly and incorrectly assigned pairs of instances.} & [0, 1] \newline Higher is better \\ \hline

\multicolumn{1}{|>{\raggedright}p{3.3cm}|}{\vspace{1cm} \textsc{Category Utility}} & \vspace{0.025cm} \[\sum_{l=1}^{k} \frac{|C_l|}{n} \sum_{j=1}^{m} \sum_{q=1}^{n_j} \left[ P(a_j^{(q)}|C_l)^2 - P(a_j^{(q)})^2 \right]\] & \multicolumn{1}{>{}p{5.2cm}|}{$k$: \#clusters, $|C_l|$: \#instances in cluster $C_l$, $n$: total \#instances, $m$: total \#attributes, $n_j$: \#categories for attribute $j$, $P(a_j^{(q)} \mid C_l)$: probability of category $q$ of attribute $j$ in cluster $C_l$, $P(a_j^{(q)})$: probability of category $q$ of attribute $j$ in the entire dataset} & \multicolumn{1}{>{}p{5.5cm}|}{Measure the clustering quality by maximizing the probability that two data points in the same cluster have the same attribute values and the probability that data points from different clusters have different attributes.} & [0, 1] \newline Higher is better \\\hline

\multicolumn{1}{|>{\raggedright}p{3.3cm}|}{\vspace{1cm} \textsc{Mutual Information}} & \vspace{0.025cm} \[\sum_{u \in U} \sum_{v \in V} p(u,v) \log \frac{p(u,v)}{p(u)p(v)}\] & \multicolumn{1}{>{}p{5.2cm}|}{$U$: set of clusters from the first clustering, $V$: set of clusters from the second clustering, $p(u,v)$: joint probability of an instance being in cluster $u$ and cluster $v$, $p(u)$: probability of an instance being in cluster $u$, $p(v)$: probability of an instance being in cluster $v$} & \multicolumn{1}{>{}p{5.5cm}|}{Measure the amount of information obtained about one clustering through another.} & [0, $\infty$] \newline Higher is better \\\hline

\multicolumn{1}{|>{\raggedright}p{3.3cm}|}{\vspace{0.025cm} \textsc{Fowlkes-Mallows Index}} & \vspace{-0.8cm} \[\sqrt{\frac{a}{a + b} \cdot \frac{a}{a + c}}\] \vspace{-0.5cm} & \multicolumn{1}{>{}p{5.2cm}|}{$a$: true positives, $b$: false positives, $c$: false negatives} & \multicolumn{1}{>{}p{5.5cm}|}{Measure the similarity between the clustering and the ground truth by evaluating the geometric mean of precision and recall.} & [0, 1] \newline Higher is better \\\hline

\multicolumn{1}{|>{\raggedright}p{3.3cm}|}{\vspace{0.025cm} \textsc{Normalized Variation of Information}} & \vspace{-0.8cm} \[\frac{H(U) + H(V) - 2I(U;V)}{\log(n)}\] \vspace{-0.5cm} & \multicolumn{1}{>{}p{5.2cm}|}{$H(U)$: entropy of clustering $U$, $H(V)$: entropy of clustering $V$, $I(U;V)$: mutual information between $U$ and $V$, $n$: \#instances} & \multicolumn{1}{>{}p{5.5cm}|}{Measure the amount of information lost and gained between two clusterings, normalized to the range [0, 1].} & [0, 1] \newline Lower is better \\\hline

\multicolumn{1}{|>{\raggedright}p{3.3cm}|}{\vspace{0.0125cm} \textsc{Homogeneity}} & \vspace{-1cm} \[H(C|K)\] \vspace{-0.5cm} & \multicolumn{1}{>{}p{5.2cm}|}{$H(C|K)$: Conditional entropy of the classes given the cluster} & \multicolumn{1}{>{}p{5.5cm}|}{Measure if each cluster contains only members of a single class.} & [0, 1] \newline Higher is better \\\hline

\multicolumn{1}{|>{\raggedright\arraybackslash}p{3.3cm}|}{\vspace{0.5cm} \textsc{Minkowski Score}} & \vspace{-0.5cm} \[\frac{\sqrt{n_{01} + n_{10}}}{\sqrt{n_{11} + n_{10}}}\] \vspace{-0.5cm} & \multicolumn{1}{>{}p{5.2cm}|}{$n_{11}$: pairs in the same cluster in both the ground truth (T) and the clustering result (S), $n_{01}$: pairs in the same cluster only in S, $n_{10}$: pairs in the same cluster only in T} & \multicolumn{1}{>{}p{5.5cm}|}{Normalized distance between pairs in the clustering result and the ground truth} & [0, $\infty$] \newline Lower is better \\\hline

\multicolumn{1}{|>{\raggedright}p{3.3cm}|}{\vspace{0.25cm} \textsc{Percentage of Correct Pair (\%CP)}} & \vspace{-0.5cm} \[\frac{|S \cap C|}{|S|} \times 100\] \vspace{-0.5cm} & \multicolumn{1}{>{}p{5.2cm}|}{$S$: Set of pairs in the same cluster according to the ground truth, $C$: Set of pairs in the same cluster according to the clustering result} & \multicolumn{1}{>{}p{5.5cm}|}{Measure the percentage of correctly clustered pairs compared to the total number of actual pairs in the same cluster.} & [0, 100] \newline Higher is better \\\hline

\end{tabular}
\end{adjustbox}
\end{table*}
\subsubsection{Computational metrics}
In addition to quality-based metrics, the computational performance of clustering algorithms is also a critical consideration. The right panel in Figure \ref{fig:internal_metrics} showcases the frequency of usage for three key computational performance metrics:
\begin{itemize}
    \item Runtime: measures the time required for an algorithm to process the input dataset and produce outputs. A lower runtime indicates better computational efficiency, allowing for faster analysis of larger datasets.
    \item Scalability: this aspect assesses how the performance of clustering algorithms changes with varying dataset sizes. Good scalability implies that an algorithm maintains consistent performance across different data volumes, making it suitable for a wide range of applications.
    \item Memory usage: also referred to as memory consumption, this metric quantifies the amount of memory required to execute a specific algorithm. Lower memory usage indicates better resource efficiency, which is particularly important when working with limited computational resources or large-scale datasets.
\end{itemize}

While lower runtime and memory usage generally indicate a more efficient clustering algorithm in terms of computational complexity, these metrics should be balanced with the quality of clustering results obtained through internal and external validation metrics.

Generally, most papers in the literature use external validation metrics to verify clustering results and algorithm performance. Among all metrics, \textsc{Accuracy}, \textsc{ARI}, \textsc{NMI}, \textsc{Purity}, \textsc{Precision}, and \textsc{Recall} are the most frequently used. Despite clustering is an unsupervised learning technique dealing with unlabeled datasets, researchers often use labeled data for experiments. In practice, label columns are omitted during clustering, and the resulting clusters are compared to the original labels to measure how well the algorithms capture the underlying ground truth of the datasets.       
\section{Categorical data clustering in different fields} \label{sec:categorical_clustering_in_other_fields}
Categorical data extend far beyond the realms of social and biomedical sciences. They are commonly found in behavioral sciences, epidemiology and public health, genetics, botany and zoology, education, marketing, engineering sciences, and industrial quality control \citep{agresti2012categorical}. In this section, we review several applications of categorical data clustering as shown in Figure \ref{fig:categorical_clustering_applications}.

\begin{figure}[!htb]
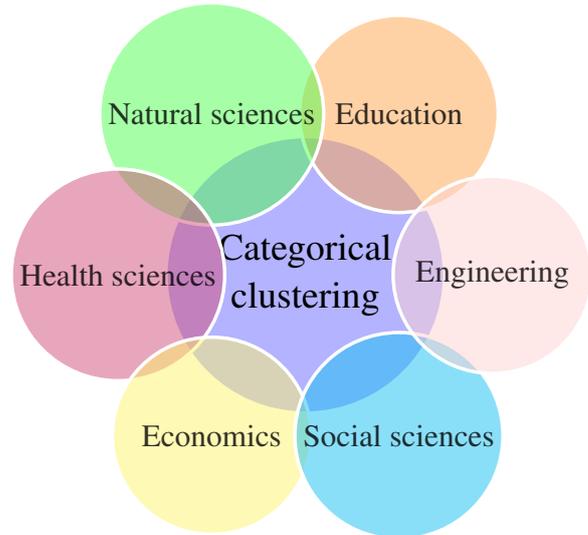

    \centering
    \smartdiagramset{
        bubble center node color=blue!30,
        set color list={orange!70!white, green!70!white, purple!70!white, yellow!70!white, cyan!70!white, pink!70!white},
        bubble center node font=\LARGE,
        bubble node font=\Large,
        bubble node size=3.2cm
    }
    \resizebox{0.9\linewidth}{!}{
    \smartdiagram[bubble diagram]{
      Categorical \\ clustering,
      Education, 
      Natural sciences, 
      Health sciences, 
      Economics, 
      Social sciences,
      Engineering}
    }
    \caption{Applications of categorical data clustering in various fields}
    \label{fig:categorical_clustering_applications}
\end{figure}
\tikzset{
    basic/.style  = {draw, rectangle},
    root/.style   = {basic, rounded corners=2pt, thin, align=center},
    onode/.style  = {basic, thin, rounded corners=2pt, align=center},
    tnode/.style  = {basic, thin, rounded corners=2pt, align=center},
    rotated tnode/.style = {tnode, rotate=90}
}

\begin{figure*}[!htb]
\vspace{-2.5cm}
\centering
\begin{adjustbox}{width=1\linewidth}
\begin{forest}
for tree={
    grow=south,
    s sep+=1pt, 
    align=left, 
    l sep+=1cm, 
    calign=center,
    tier/.wrap pgfmath arg={tier #1}{level()},
    inner sep=5pt, fill=none,
    edge path={
        \noexpand\path [\forestoption{edge}]
        (!u.parent anchor) |- +(0,-30pt) -| (.child anchor)\forestoption{edge label};
    }
}
[\fontsize{50}{60}\selectfont {Categorical data clustering}, root
    [\fontsize{40}{60}\selectfont {Natural sciences}, onode
        [\fontsize{30}{60}\selectfont  {Agricultural science} \\ \hline \\
           \fontsize{30}{60}\selectfont  {\cite{akos2021evaluation}} \\
           \fontsize{30}{60}\selectfont {\cite{zhang2021identifying}}
           ,tnode
        ]
        [\fontsize{30}{60}\selectfont {Biology} \\ \hline \\
           \fontsize{30}{60}\selectfont{\cite{dettling2002supervised}} \\ 
           \fontsize{30}{60}\selectfont{\cite{yang2003cluseq}} \\ 
           \fontsize{30}{60}\selectfont{\cite{zhao2005data}} \\ 
           \fontsize{30}{60}\selectfont{\cite{guo2016cluster}} \\ 
           \fontsize{30}{60}\selectfont{\cite{yan2022bayesian}} \\ 
           \fontsize{30}{60}\selectfont{\cite{alexandre2022disa}} \\ 
           \fontsize{30}{60}\selectfont{\cite{riva2023toward}}, tnode
        ]
        [\fontsize{30}{60}\selectfont {Chemistry} \\ \hline \\
           \fontsize{30}{60}\selectfont{\cite{xiao2005supervised}} \\
           \fontsize{30}{60}\selectfont{\cite{livera2015statistical}} \\
           \fontsize{30}{60}\selectfont{\cite{doncheva2018cytoscape}} \\
           \fontsize{30}{60}\selectfont{\cite{sakamuru2021predictive}} \\
           \fontsize{30}{60}\selectfont{\cite{morishita2022initial}}, tnode    
        ]
        [\fontsize{30}{60} \selectfont{Physics} \\ \hline \\
            \fontsize{30}{60}\selectfont{\cite{tamarit2020hierarchical}} \\
           \fontsize{30}{60}\selectfont{\cite{visani2024holographic}}, tnode
        ]
        [\fontsize{30}{60}\selectfont {Materials science} \\ \hline \\
           \fontsize{30}{60}\selectfont{\cite{rahnama2019application}} \\ 
           \fontsize{30}{60}\selectfont{\cite{huang2020structure}} \\ 
           \fontsize{30}{60}\selectfont{\cite{hase2021gryffin}}, tnode
        ]
        [\fontsize{30}{60}\selectfont {Neuroscience} \\ \hline \\
           \fontsize{30}{60}\selectfont{\cite{kriegeskorte2008matching}} \\ 
           \fontsize{30}{60}\selectfont{\cite{borghesani2016word}} \\
           \fontsize{30}{60}\selectfont{\cite{nevado2021preserved}} \\
           \fontsize{30}{60}\selectfont{\cite{baenas2024cluster}}, tnode
        ]
    ]
    [\fontsize{40}{60}\selectfont  {Health sciences} \\ \hline \\
        \fontsize{30}{60}\selectfont{\cite{burgel2012two}} \\
        \fontsize{30}{60}\selectfont{\cite{vanfleteren2013clusters}} \\
        \fontsize{30}{60}\selectfont{\cite{ravishankar2013association}} \\
        \fontsize{30}{60}\selectfont{\cite{mamykina2016revealing}} \\
        \fontsize{30}{60}\selectfont{\cite{hose2017latent}} \\
        \fontsize{30}{60}\selectfont{\cite{papachristou2018congruence}} \\
        \fontsize{30}{60}\selectfont{\cite{adegunsoye2018phenotypic}} \\
        \fontsize{30}{60}\selectfont{\cite{davoodi2018mortality}} \\
        \fontsize{30}{60}\selectfont{\cite{zahraei2020comprehensive}} \\
        \fontsize{30}{60}\selectfont{\cite{elmer2020unsupervised}} \\
        \fontsize{30}{60}\selectfont{\cite{gonzalez2021phenotypes}} \\
        \fontsize{30}{60}\selectfont{\cite{vagts2021unsupervised}} \\
        \fontsize{30}{60}\selectfont{\cite{webster2021characterisation}} \\
        \fontsize{30}{60}\selectfont{\cite{mulick2021four}} \\
        \fontsize{30}{60}\selectfont{\cite{morandini2022artificial}} \\
        \fontsize{30}{60}\selectfont{\cite{ding2023uncovering}} \\
        \fontsize{30}{60}\selectfont{\cite{saito2023phenotyping}} \\
        \fontsize{30}{60}\selectfont{\cite{shpigelman2023clustering}} \\
        \fontsize{30}{60}\selectfont{\cite{abbasi2023clinical}} \\
        \fontsize{30}{60}\selectfont{\cite{ferreira2023obstructive}} \\
        \fontsize{30}{60}\selectfont{\cite{gabarrell2023variations}} \\
        \fontsize{30}{60}\selectfont{\cite{solomon2023assessing}} \\
        \fontsize{30}{60}\selectfont{\cite{wang2024meals}} \\
        \fontsize{30}{60}\selectfont{\cite{ji2024identifying}}
        , onode
    ]
    [\fontsize{40}{60}\selectfont  {Economics} \\ \hline \\
        \fontsize{30}{60}\selectfont{\cite{de2014mining}} \\
       \fontsize{30}{60}\selectfont{\cite{marchal2019detecting}} \\
       \fontsize{30}{60}\selectfont{\cite{kundu2021cloud}} \\
       \fontsize{30}{60}\selectfont{\cite{huang2024imbalanced}} \\
       \fontsize{30}{60}\selectfont{\cite{andonovikj2024survival}}
       , onode
    ]
    [\fontsize{40}{60}\selectfont  {Education}  \\ \hline \\
    \fontsize{30}{60}\selectfont{\cite{baumann2022research}} \\
    \fontsize{30}{60}\selectfont{\cite{sugumaran2022rough}}
    , onode]
     [\fontsize{40}{60}\selectfont  {Engineering} \\ \hline \\
       \fontsize{30}{60}\selectfont{\cite{andreopoulos2007clustering}} \\ 
       \fontsize{30}{60}\selectfont{\cite{singh2019chain}} \\
       \fontsize{30}{60}\selectfont{\cite{perea2021carbon_in_waterdss}} \\
       \fontsize{30}{60}\selectfont{\cite{sadeghi2023apache}} \\
       \fontsize{30}{60}\selectfont{\cite{tamakloe2024critical}} \\
       \fontsize{30}{60}\selectfont{\cite{wei2024analysis}},
       onode
    ]
    [\fontsize{40}{60}\selectfont {Social sciences} \\ \hline \\
       \fontsize{30}{60}\selectfont{\cite{okada2015personality}} \\
       \fontsize{30}{60}\selectfont{\cite{froemelt2018using}} \\
       \fontsize{30}{60}\selectfont{\cite{okada2019modeling}} \\
       \fontsize{30}{60}\selectfont{\cite{sgroi2024analyzing}} 
       , onode
    ]
]
\end{forest}
\end{adjustbox}
\caption{A taxonomy of categorical data clustering in different fields}
\label{Fig:categorical_data_clustering_applications}
\end{figure*}
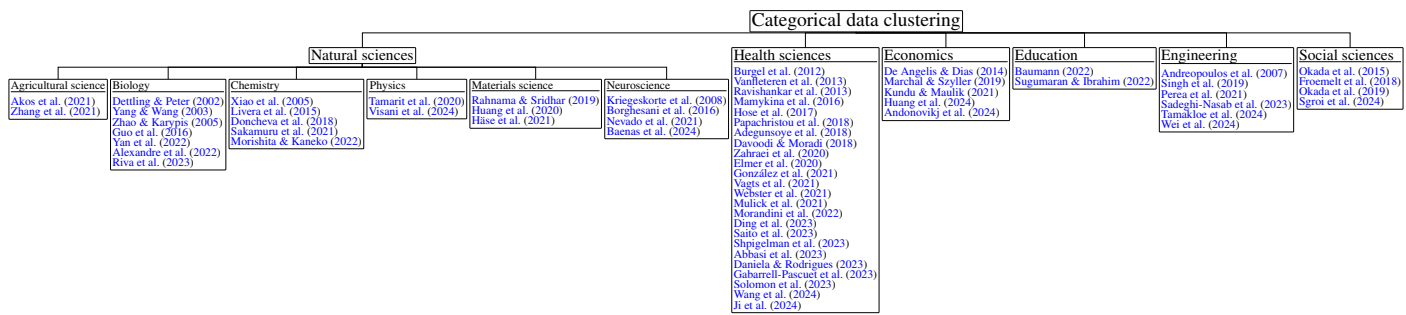
\begin{figure*}[!htb]
\vspace{-0.3cm}
  \centering
  \includegraphics[width=\linewidth]{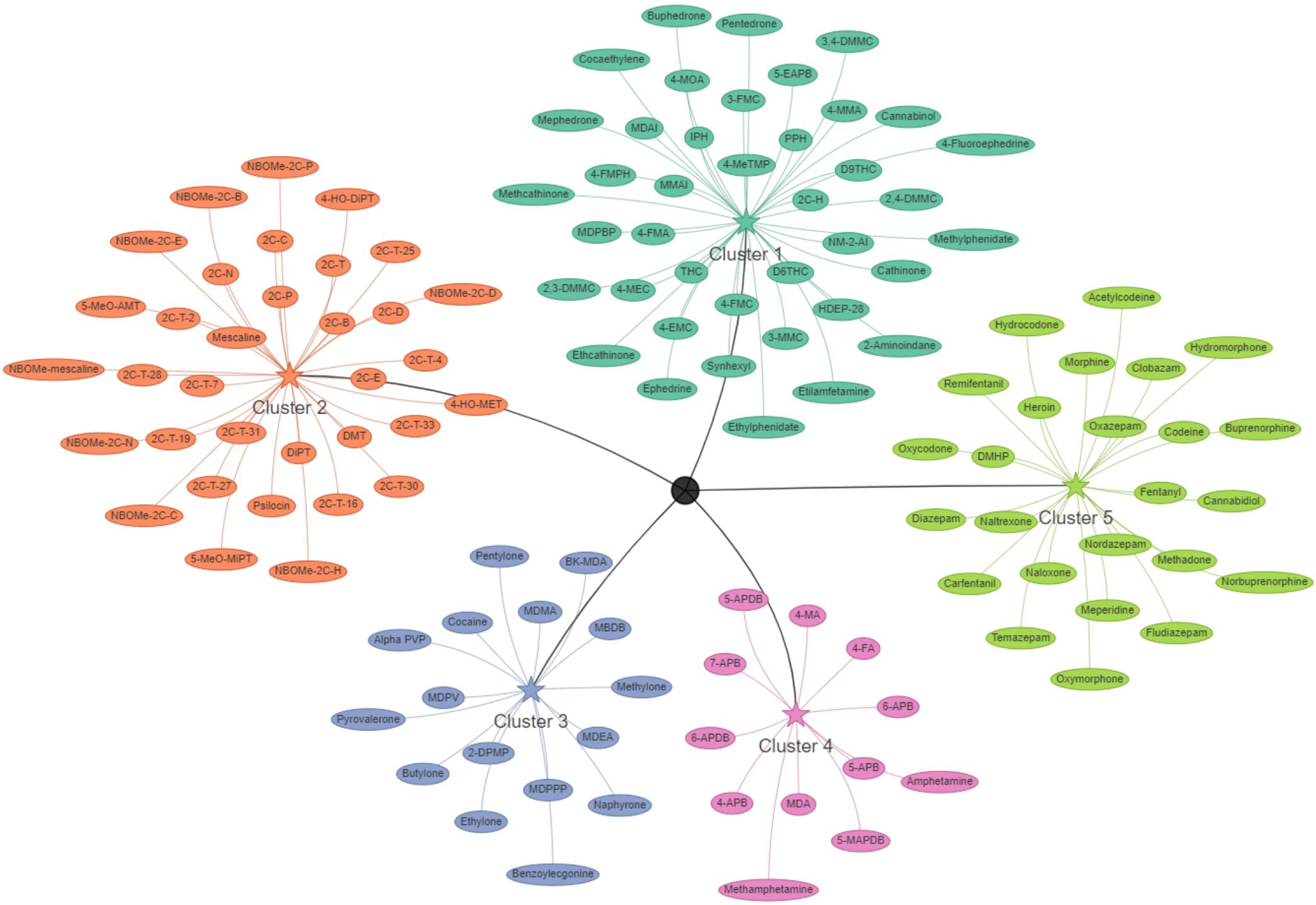}
  \caption{Clustering results on the Novel Psychoactive Substances (NPS) dataset \citep{he2023examining} into five clusters by \textsc{K-modes}}
  \label{fig:NPS_data}
\end{figure*}
\subsection{Health sciences}
Although visual, audio, and text data are increasingly utilized within medical research, categorical data maintains a substantial proportion in the field of healthcare, not least in register records and interview/survey responses. After decades of a largely ``one size fits all" perspective on disease followed by hypothetical-deductive reasoning to derive subgroups of patients and subtypes of disorders, it has been made apparent that there is a vast heterogeneity in pathophysiology which is often too complex to be overviewed manually \citep{loftus2022phenotype}.

Cluster analysis is a promising tool for accelerating the discovery of clinically meaningful disease subgroups, given the ability of unsupervised machine learning techniques to traverse high dimensional input data and within it discern latent patterns in a more data-driven manner to provide as hypothesis-generating output. However, computational methodology is often lacking, even in works published in high-impact journals, as few researchers in this field receive formal training or collaborate/consult with relevant methodologists. For this reason, novel iterations of classical algorithms such as \textsc{K-modes} are seldom utilized, and likewise, validation and model selection is often suboptimally performed and reported, although comprehensive reviews/guidelines and end-to-end software packages have been produced in recent years \citep{esnault2023qluster}. Nevertheless, categorical/mixed data cluster analysis has enabled discovery of impactful clusters of disorder subtypes \citep{papachristou2018congruence}, risk strata \citep{ding2023uncovering}, and comorbidity patterns \citep{vanfleteren2013clusters,morandini2022artificial}.

In the cross-sectional setting with categorical or mixed data, algorithms such as partitioning around medoids (\textsc{PAM}) \citep{adegunsoye2018phenotypic,vagts2021unsupervised} and \textsc{K-Prototypes} \citep{elmer2020unsupervised,saito2023phenotyping,shpigelman2023clustering,abbasi2023clinical} are commonplace, with attributes to stability to outliers and ability to accommodate mixed data (typically after preceding construction of an appropriate distance matrix or embedding of categorical data to a continuous subspace).
\cite{davoodi2018mortality} proposed a Deep Rule-Based Fuzzy System to generate interpretable fuzzy rules from heterogeneous ICU patient data, including both numeric and categorical variables. The algorithm adapts k-means clustering by combining Euclidean distance for numeric attributes with a matching dissimilarity measure for categorical attributes.

\textsc{K-modes} \citep{papachristou2018congruence,morandini2022artificial,solomon2023assessing} and \textsc{Hierarchical Cluster Analysis (HCA)} \citep{burgel2012two,zahraei2020comprehensive,gonzalez2021phenotypes, webster2021characterisation} are also relatively often used. 
\cite{mamykina2016revealing} proposed a mixed-methods approach to studying interdisciplinary handoff in critical care. Verbal handoff discussions were coded to create categorical datasets. \textsc{Hierarchical Clustering} and \textsc{K-means} were used to analyze these datasets, revealing systematic patterns in clinical communication. The findings can inform the design of electronic tools to support interdisciplinary team handoffs.
\cite{ferreira2023obstructive} clustered obstructive sleep apnea patients using 13 categorical variables, reduced from an initial 51 through preprocessing and feature selection. The \textsc{K-modes} algorithm identifies three distinct patient groups based on these similarities.

Longitudinal patterns of how diseases develop over time are typically also of interest to discern. In this context, \textsc{Latent Class Analysis (LCA)} \cite{hose2017latent,mulick2021four} is arguably the most popular algorithm, utilized in numerous high-impact works. 
Recently, \cite{gabarrell2023variations} used model-based clustering to categorical responses from 2,900 individuals on the nine PHQ-9 depression screening questions, which were on a 4-level Likert scale. The algorithm identified six latent clusters with distinct patterns of depressive symptoms.
\cite{wang2024meals} employed a cluster randomized controlled trial design, with secondary schools assigned to three groups: full intervention, partial intervention, and control. They used categorical data on adolescent and parent characteristics to evaluate the impact of the nutrition intervention on various nutrition-related outcomes.
\cite{ji2024identifying} used a multi-kernel manifold method for multi-site clustering to identify psychosis subtypes from aggregated individualized covariance structural differential networks. Gaussian kernels measure inter-patient distance, and the SIMLR method learns the similarity matrix. 

Other methods have been applied for clustering categorical healthcare datasets.
\cite{ravishankar2013association} employed group-based trajectory modeling to cluster subjects by their growth patterns and BNP levels over time. This method reveals population heterogeneity in attributes like height z-score and their changes over time. Clustering simplifies regression by converting multiple correlated continuous predictors into a single categorical cluster indicator.
\cite{patterson2023emergency} used clustering algorithms to manage hierarchical categorical data, such as EMS agencies and their clinicians. By randomly assigning agencies to intervention or control groups, the algorithm accounts for within-agency similarities while comparing outcomes like sleep quality and fatigue. This method helps analyze agency characteristics and survey responses to evaluate the impact of sleep health education.

Figure \ref{fig:NPS_data} illustrates the application of \textsc{K-modes} on the Novel Psychoactive Substances (NPS) dataset, which includes molecular structures and formulas of synthetic or designer drugs. They are designed to mimic the effects of traditional controlled substances, such as cannabis and cocaine, often circumventing existing drug laws, posing challenges for law enforcement and public health agencies. We used the dataset from \cite{he2023examining}, which contains 127 unique NPS compounds classified into 16 major chemical structure categories, including opioids, stimulants, hallucinogens, sedatives, and cannabinoids. For this experiment, we excluded all numerical attributes and employed the Elbow method to identify the optimal number of clusters. The results demonstrate that \textsc{K-modes} effectively groups similar substances into five distinct clusters.

\subsection{Natural sciences}
Categorical data is widely used in various natural science disciplines, including agricultural science, biology, chemistry, physics, materials science, and neuroscience.

In \textsc{Agricultural science}, \cite{akos2021evaluation} examines improved rice lines from three parent cultivars resistant to bacterial leaf blight, blast, and drought stress. Categorical datasets detail the genotyping, showing specific resistance genes and drought tolerance quantitative trait loci. Clustering algorithms, including UPGMA and PCA, analyze genetic variability and relationships among the genotypes.
\cite{zhang2021identifying} employed clustering algorithms to classify manure and sewage management practices among dairy farms in China. By analyzing categorical data, it identifies key pathways and simplifies diverse strategies, offering insights for policy-making and sustainability assessment.

In \textsc{Biology}, categorical data is used to classify organisms, their genetic traits, and various biological processes, supporting the organization and understanding of complex biological systems. In genetics, categorical data involve the types of alleles inherited by offspring. In botany and zoology, they are used to determine whether a particular organism is observed in a sampled quadrat \citep{agresti2012categorical}.
Many clustering algorithms in section \ref{sec:taxonomy} used biological datasets from UCI such as \textsc{Splice-junction Gene Sequences}, \textsc{Molecular Biology (Promoter Gene Sequences)}, \textsc{Mushroom}, and \textsc{Soybean} in the experiments. 

In addition to the algorithms listed above, \cite{dettling2002supervised} proposed a supervised clustering algorithm that uses response variable information (tissue types) to guide the clustering of gene expression data from microarrays. The algorithm identifies gene groups that differentiate tissue types or cancer classes by employing a greedy forward search and optimizing an objective function based on \emph{Wilcoxon's test statistic} and a margin criterion. It incrementally builds clusters by adding genes that enhance class discrimination. The algorithm was tested on datasets for leukemia, breast cancer, prostate cancer, colon cancer, small round blue cell tumors, lymphoma, brain tumors, and NCI cancer cell lines, representing categorical biological data.
\cite{yang2003cluseq} proposed \textsc{CLUSEQ} for a database of 8000 protein sequences from 30 biological families. Protein sequences are categorical data, as they consist of ordered lists of amino acids. \textsc{CLUSEQ} employs a probabilistic suffix tree to identify significant patterns and uses conditional probability distributions to measure sequence similarity.
\cite{zhao2005data} discussed clustering approaches including \textsc{Group-average Agglomerative Clustering}, \textsc{K-means}, \textsc{Self-Organizing Maps}, \textsc{Subspace Clustering}, \textsc{Dynamic Time-Warping}, and \textsc{Model-Based Clustering} for life science datasets. These datasets include biological sequences, clinical information, gene functional categorization, protein families and structures, phylogenetic data, microarray experiments, and protein motifs. The algorithms aim to identify co-regulated genes, distinctive tissue types, common inducers, and organisms with similar responses.

\cite{guo2016cluster} proposed the \textsc{Robust K-means for Sequences} algorithm for biological sequences like glycoside hydrolases family 2, bacterial genomes, and homologous gene families from microbes. This algorithm improves \textsc{K-means} by optimizing initial centroid selection and eliminating noise clusters. \cite{yan2022bayesian} outlines a general Bayesian approach to bi-clustering in moderate to high dimensions and introduces three Bayesian bi-clustering models for categorical data. These methods handle feature distribution complexities and are applied to genetic datasets, including human single nucleotide polymorphisms from the HapMap project and single-cell RNA sequencing data. \cite{alexandre2022disa} introduced DISA, a Python software for evaluating patterns in biological datasets with categorical and numerical outcomes. DISA enhances pattern discovery and subspace clustering by assessing patterns' discriminative power in gene expression analysis, biomarker discovery, and metabolic engineering.
\cite{riva2023toward} used the \textsc{Louvain} clustering algorithm to identify clusters in a co-citation network constructed from the reference lists of articles on ecological complexity.

In \textsc{Chemistry}, categorical data classifies compounds, elements and their properties, aiding in the systematic study and understanding of chemical reactions and interactions. \cite{xiao2005supervised} proposed a \textsc{Supervised Self-Organizing Map (sSOM)} that uses nonlinear, topology-preserving mapping to classify compounds into categorical bins based on biological activity. It combines vector quantization and clustering to map high-dimensional data onto a lower-dimensional grid of neuronal nodes, each with a prototype vector. During training, the algorithm identifies the best matching unit (BMU) for each input and updates the BMU and its neighbors using the Kohonen rule, preserving distance and proximity relationships. By incorporating descriptor data and class information, \textsc{sSOM} enables supervised clustering that links molecular features to biological activity categories.
\cite{livera2015statistical} used \textsc{Hierarchical Clustering} on both samples and metabolites to evaluate how well the data identifies the correct groupings. It uses Manhattan distance and Ward's minimum variance agglomerative clustering for this purpose. This clustering approach assesses how different normalization methods preserve biological variation of interest while removing unwanted variation.
\cite{doncheva2018cytoscape} used the Markov clustering algorithm via the \textsc{clusterMaker2 Cytoscape} app to group proteins based on STRING interactions. This simplified network visualization by retaining only intra-cluster interactions, facilitating data interpretation. The clustered network was then color-coded to visualize categorical data, such as phosphorylation cluster assignments.
\cite{sakamuru2021predictive} used the \textsc{K-modes} algorithm to cluster compounds based on PubChem fingerprints, identifying similar groups for further analysis. \cite{morishita2022initial} converted categorical variables (ligands, solvents, bases) into numerical descriptors using Mordred software. They then applied clustering algorithms (\textsc{K-means}, \textsc{K-Medoids}, \textsc{DBSCAN}, \textsc{PAM}) to ensure uniformly selected samples from the chemical space, avoiding similar experimental conditions.

In \textsc{Physics}, categorical data helps categorize different physical phenomena, particles, and their characteristics, facilitating the exploration of the fundamental laws of nature. \cite{tamarit2020hierarchical} proposed a hierarchical clustering algorithm called \textsc{clusterBip} for analyzing complex networks and detecting community structures. The algorithm first performs Fisher's exact test (FET) on every pair of entities in the bipartite network. It then uses the SLINK, a single-linkage clustering algorithm, to generate a hierarchical clustering based on the results of these pairwise FETs. \textsc{clusterBip} applies to various physical systems, such as plant-pollinator interactions or gene-virus associations. It incorporates concepts from statistical physics, including susceptibility from percolation theory, to identify optimal clustering partitions.
\cite{visani2024holographic} used \textsc{K-means} to evaluate the learned latent space representations for categorical datasets like MNIST digits and 3D object classes from Shrec17. The clustering results demonstrated that the \textsc{Holographic-(V)AE} model learns meaningful, separable representations. The \textsc{Holographic-(V)AE} model emphasizes SO(3) equivariance (rotational symmetry in 3D space) and utilizes spherical harmonics and Fourier space representations, concepts deeply rooted in physics.

In \textsc{Materials Science}, materials are typically described using a combination of numerical and categorical features, collectively referred to as material descriptors.
Numerical features, such as \emph{atomic mass}, \emph{ionization energy}, and \emph{electronegativity}, are continuous values that characterize the constituent atoms of a material.
In contrast, categorical features are crucial in material design. These include discrete properties like the types of atoms (\emph{species}), their \emph{atomic numbers}, positions in the periodic table (\emph{periods}, \emph{groups}, and \emph{blocks}), and the number of \emph{valence electrons}.
Categorical variables also appear in experimental contexts, such as synthesis actions, processing techniques, and elemental combinations, making them integral to various aspects of materials research. Since material discovery often involves selecting and combining discrete options—such as specific elements, processing methods, and synthesis conditions—categorical data is fundamental in enabling systematic exploration within materials science.

\cite{rahnama2019application} used \textsc{K-means} to identify patterns and similarities across categorical material classes in the Hydrogen Storage Materials Database, which contains numerical and categorical features on various metal hydride materials classified into different material classes such as Mg alloys, complex hydrides, solid solution alloys, and A2B intermetallic compounds.
\cite{huang2020structure} used \textsc{Self-Organizing Maps (SOM)} to cluster and visualize high-dimensional data on organic photovoltaic (OPV) polymer materials. The SOM creates 2D maps grouping materials with similar properties. An ``information projection" function overlays categorical data, such as chemical elements and structural information, onto these maps, enabling researchers to visually identify correlations between categorical features (e.g., certain atoms or structural characteristics) and OPV performance metrics, especially power conversion efficiency.
\cite{hase2021gryffin} propsosed an algorithm named \textsc{Gryffin} for clustering and optimizing datasets with categorical features. \textsc{Gryffin} uses Bayesian optimization with kernel density estimation directly in the categorical space, leveraging one-hot encoding and domain-specific descriptors. This method efficiently handles categorical parameters, enhancing optimization in fields like materials science and chemistry.

\begin{figure}[!htb]
  \centering
  \includegraphics[width=\linewidth]{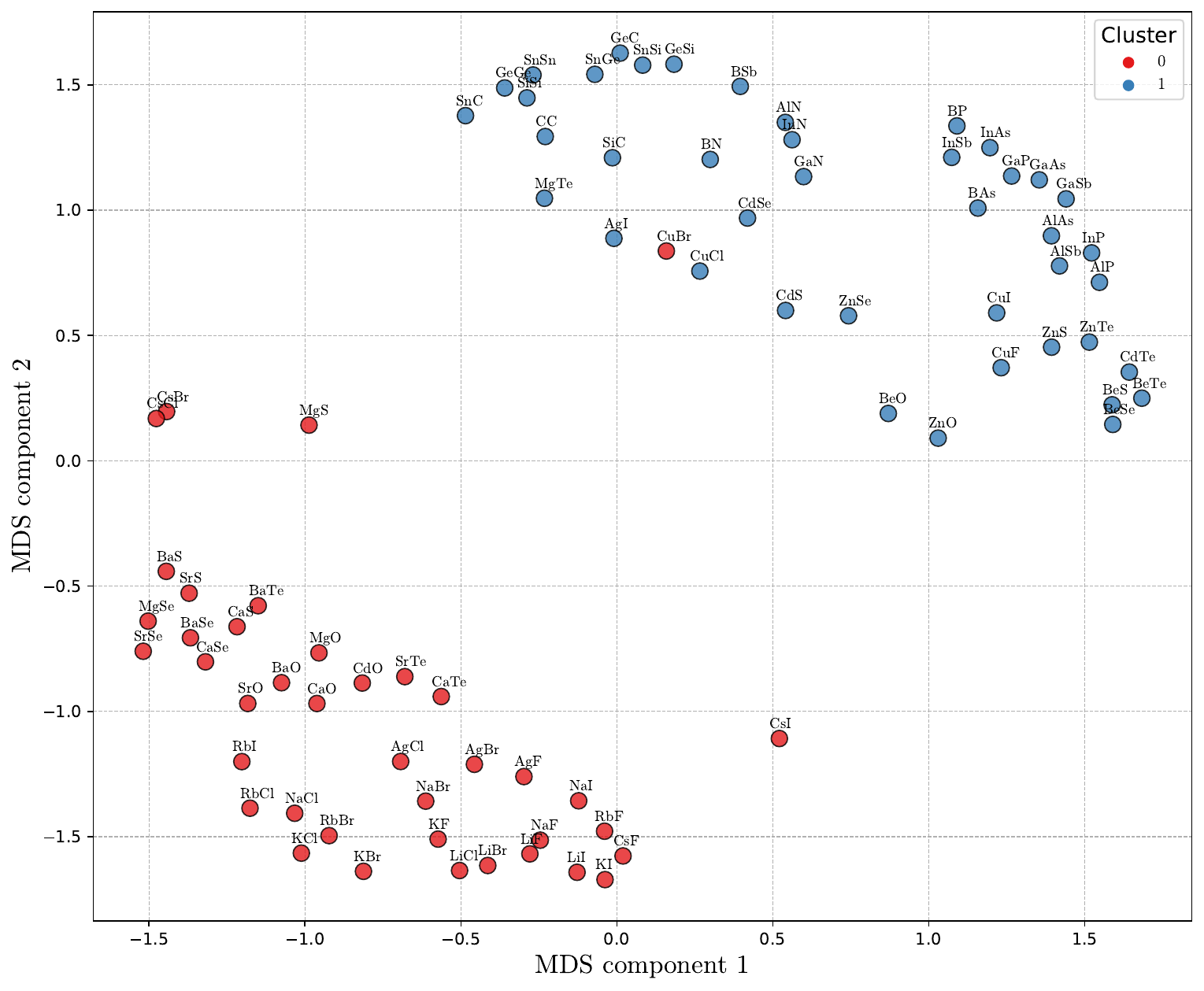}
  \caption{\textsc{K-modes} results on the binary octet dataset \citep{pilania2015classification}}
  \label{fig:binary_octet}
\end{figure}

Figure \ref{fig:binary_octet} illustrates the application of \textsc{K-modes} clustering on a dataset comprising 82 sp-bonded binary octet compounds \citep{pilania2015classification}. These compounds are AB-type materials, with elements from groups I-VII of the periodic table. They crystallize in various structures, including zincblende (ZB), rocksalt (RS), wurtzite (WZ), cesium chloride (CsCl), and diamond cubic (DC) crystal structures. The dataset includes compounds with both four-fold coordination (ZB/WZ) and six-fold coordination (RS) environments. For this experiment, we considered ZA, ZB, classification labels, observed structure, Born effective charges in ZB, $\alpha_p$, and delta-EWZ as categorical features, given the presence of many repeated values among the 82 compounds for these features. The clustering results were generated using \textsc{K-modes} with two clusters. A pairwise distance matrix was then constructed using the Hamming distance for all instances in the dataset. To visualize the clustering results, \textsc{Multidimensional Scaling (MDS)} was applied to reduce the distance matrix into a 2D space.

In \textsc{Neuroscience}, \cite{kriegeskorte2008matching} analyzed 92 images of real-world objects (animate and inanimate) and measured inferior temporal (IT) response patterns using single-cell recordings in monkeys and high-resolution fMRI in humans. \textsc{Hierarchical Clustering} revealed similar categorical structures in both species, with distinct clusters for animate and inanimate objects, and subclusters for faces and bodies. The findings suggest a common code in primate IT response patterns emphasizing behaviorally important categorical distinctions.
\cite{borghesani2016word} utilized multivariate pattern analysis techniques, including decoding and representational similarity analysis, to analyze fMRI data from participants reading words related to animals and tools. The study identified a gradient of semantic coding along the ventral visual stream: early visual areas mainly encoded perceptual features, such as implied real-world size, while more anterior temporal regions encoded conceptual features, such as taxonomic categories. Even after controlling for perceptual features, the researchers found that the anterior temporal lobe could still distinguish between semantic categories and sub-categorical clusters.
\cite{nevado2021preserved} used clustering algorithms to analyze the semantic organization of words from a verbal fluency test. A hierarchical clustering algorithm identified semantic categories, comparing the structure of semantic memory in healthy older adults and those with mild cognitive impairment.
\cite{baenas2024cluster} used a two-step clustering method on a sample of 297 treatment-seeking adults with gambling disorder (GD) to identify subtypes of GD patients based on sociodemographic, neuropsychological, and neuroendocrine features. 

\subsection{Education}

\cite{baumann2022research} employed a two-step clustering method that first preclustered cases, followed by hierarchical clustering of the preclusters. The optimal number of clusters was determined automatically using the \emph{Schwarz Bayesian Criterion (BIC)}. The analysis incorporated both categorical and continuous variables related to research output, motivation, qualification, knowledge, self-efficacy, experience, and working time in Swiss non-traditional higher education institutions.
\cite{sugumaran2022rough} proposed \textsc{RS-LDNI} algorithm, which clusters e-learners based on their learning styles using the VAK (Visual, Auditory, Kinesthetic) model. It used the equivalence class concept from rough set theory and the coefficient of variation for categorical data. The algorithm selects attributes for splitting using the least dissimilarity index and measures variability with the coefficient of unlikability.
\subsection{Engineering}
\cite{andreopoulos2007clustering} proposed the \textsc{MULICsoft} algorithm, which incorporates both static (structural) and dynamic (runtime) information into the clustering process. It is applied to categorical datasets representing dependencies between source files, organized into a matrix where each entry shows if one file calls or references another, with weights indicating call frequency.
\cite{singh2019chain} used \textsc{K-modes} and \textsc{Expectation-Maximization (EM)} to identify hazardous elements, initiating events, pivotal events, accident scenarios, and their consequences to improve workplace safety. The datasets include both proactive data (workplace observations and high-risk control program data) and reactive data (incident records) collected from an integrated steel plant.
\cite{perea2021carbon_in_waterdss} designed the Carbon\_in\_WaterDSS tool to cluster hourly energy transformation rates into High, Medium, and Low categories based on categorical data such as electricity tariff periods (P1, P2, P3) and days of the week. This clustering identifies patterns in energy usage and emissions, aiding in irrigation scheduling to potentially lower carbon footprints.
\cite{sadeghi2023apache} proposed an IoT data anonymization framework using Apache Flink and clustering. It assigns cluster-IDs, treats categories as superclusters, and uses a histogram-like approach for numerical data. The algorithm processes stream data by clustering, checking cluster sizes, and handling expired data.
\cite{tamakloe2024critical} used Cluster Correspondence Analysis (CCA) to segment PMD rider-at-fault crash data into distinct groups. CCA integrates dimension reduction, cluster analysis, and correspondence analysis to reveal unique risk factor patterns within each cluster, providing insights for targeted safety interventions.
\cite{wei2024analysis} proposed a framework to identify typical bicycle crash scenarios from categorical data. It uses \textsc{Pearson's Chi-Squared} test and \textsc{Categorical Principal Component Analysis} to select relevant variables and reduce dimensionality, respectively. The \textsc{CLARA} algorithm, an extension of \textsc{K-medoids} clustering, is then used to efficiently handle the large dataset and identify seven typical crash scenarios.
\begin{figure*}[!htb]
\vspace{-2.5cm}
  \centering
  \includegraphics[width=0.48\linewidth]{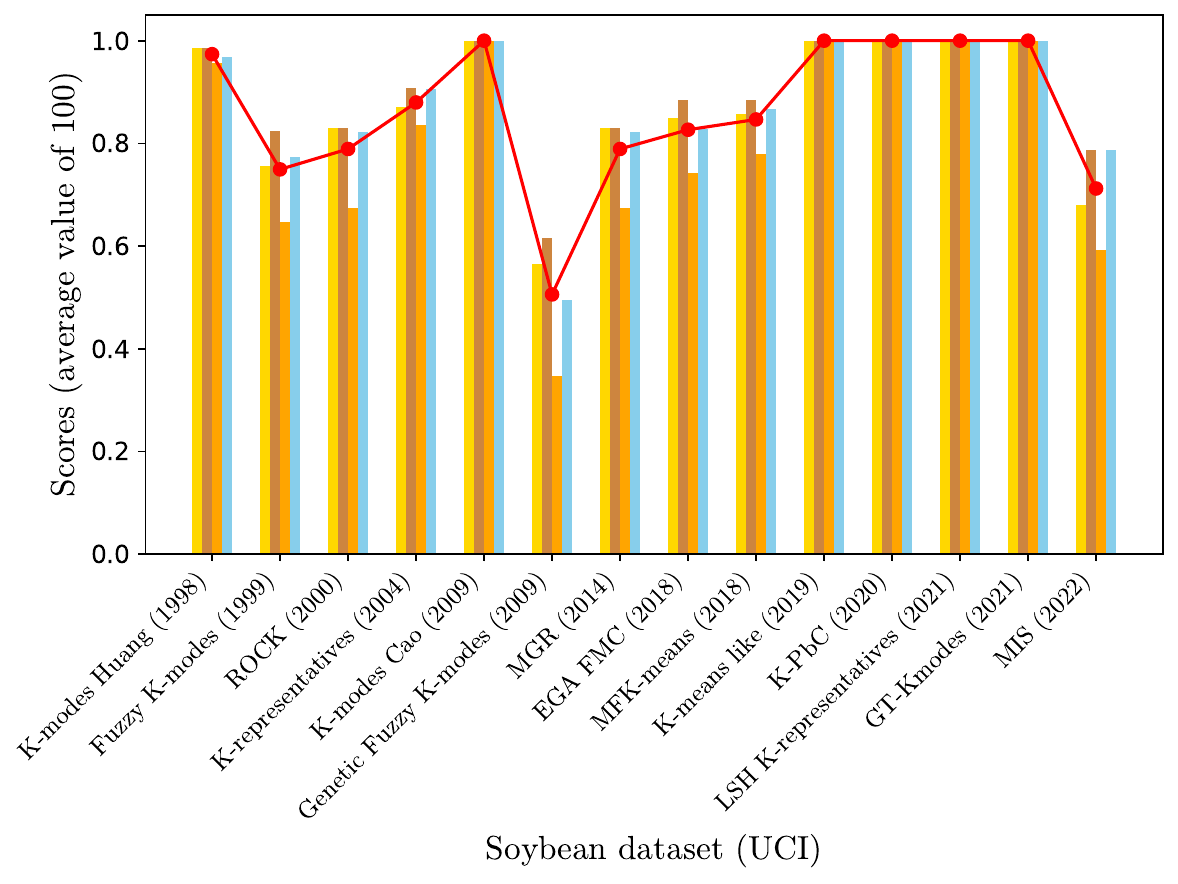}
  \includegraphics[width=0.48\linewidth]{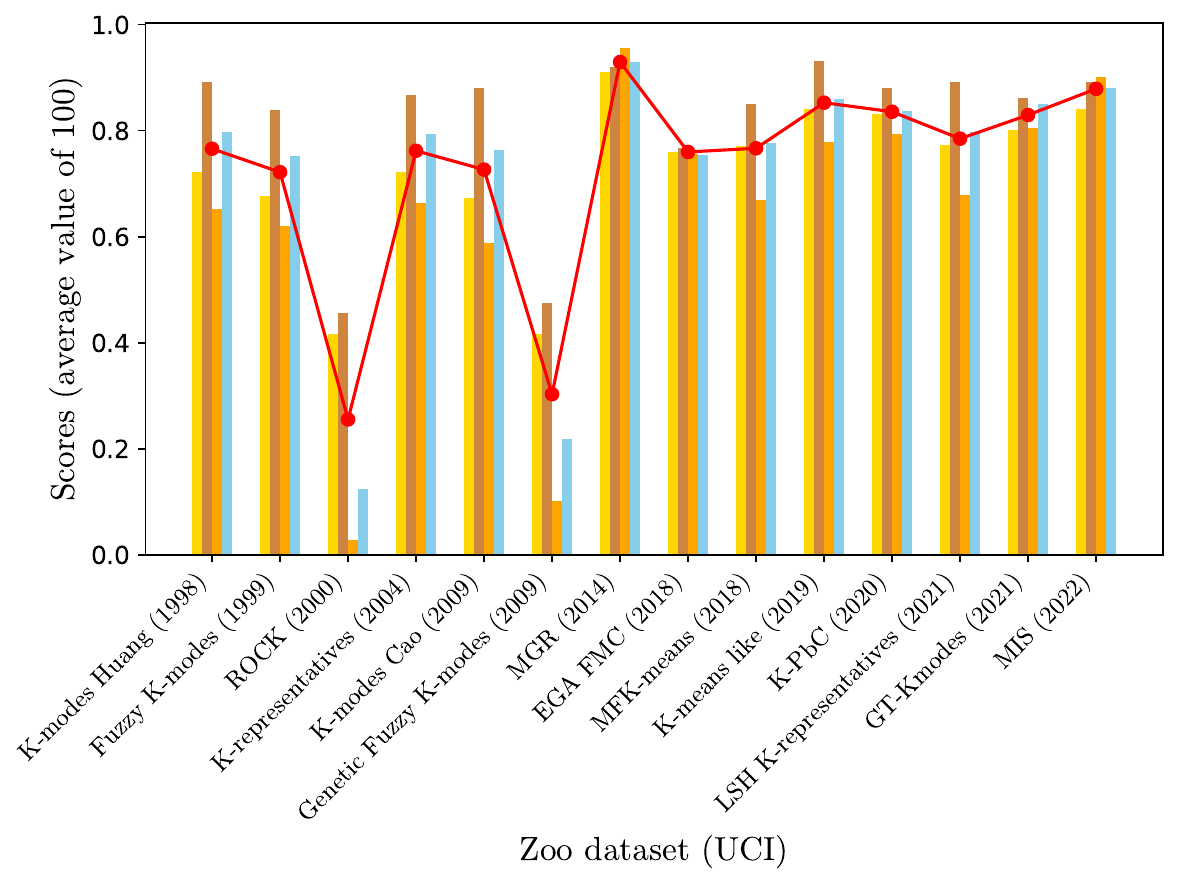}
  \includegraphics[width=0.48\linewidth]{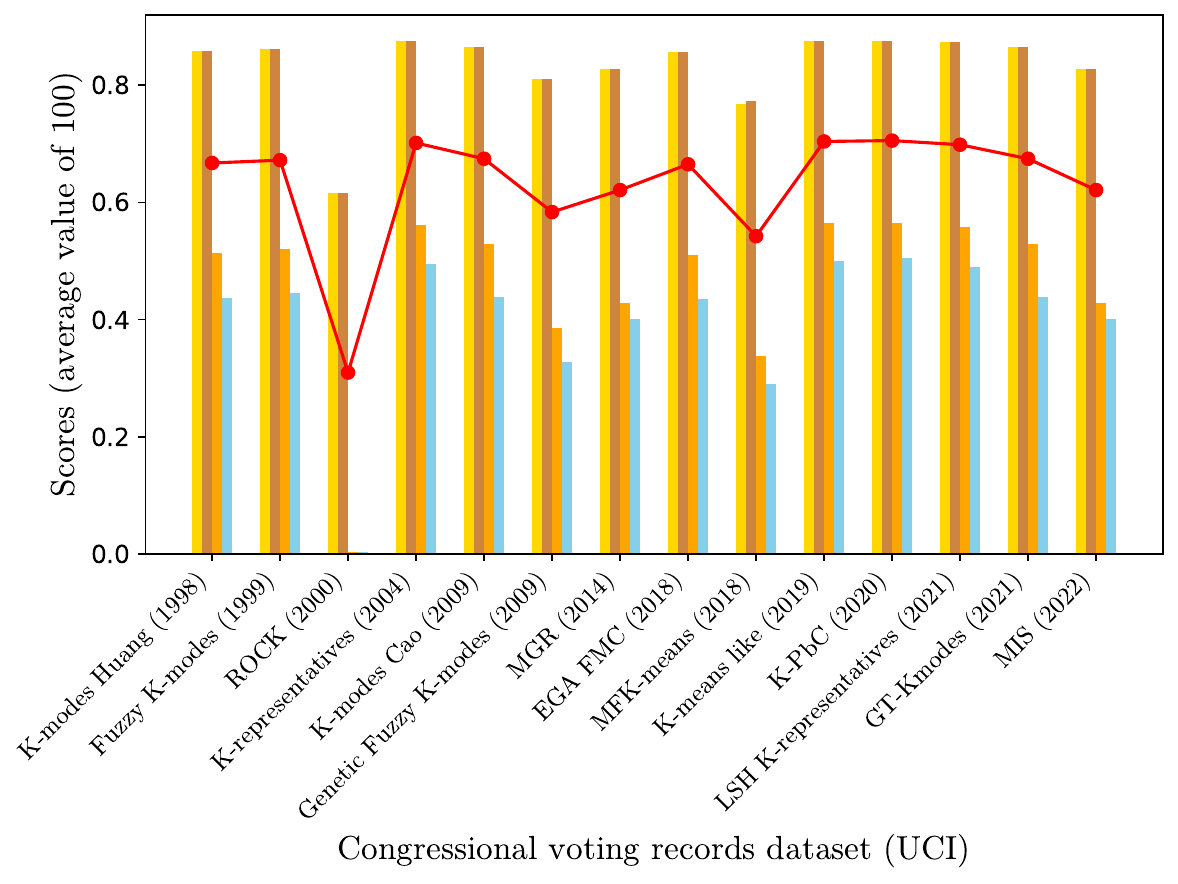}
  \includegraphics[width=0.48\linewidth]{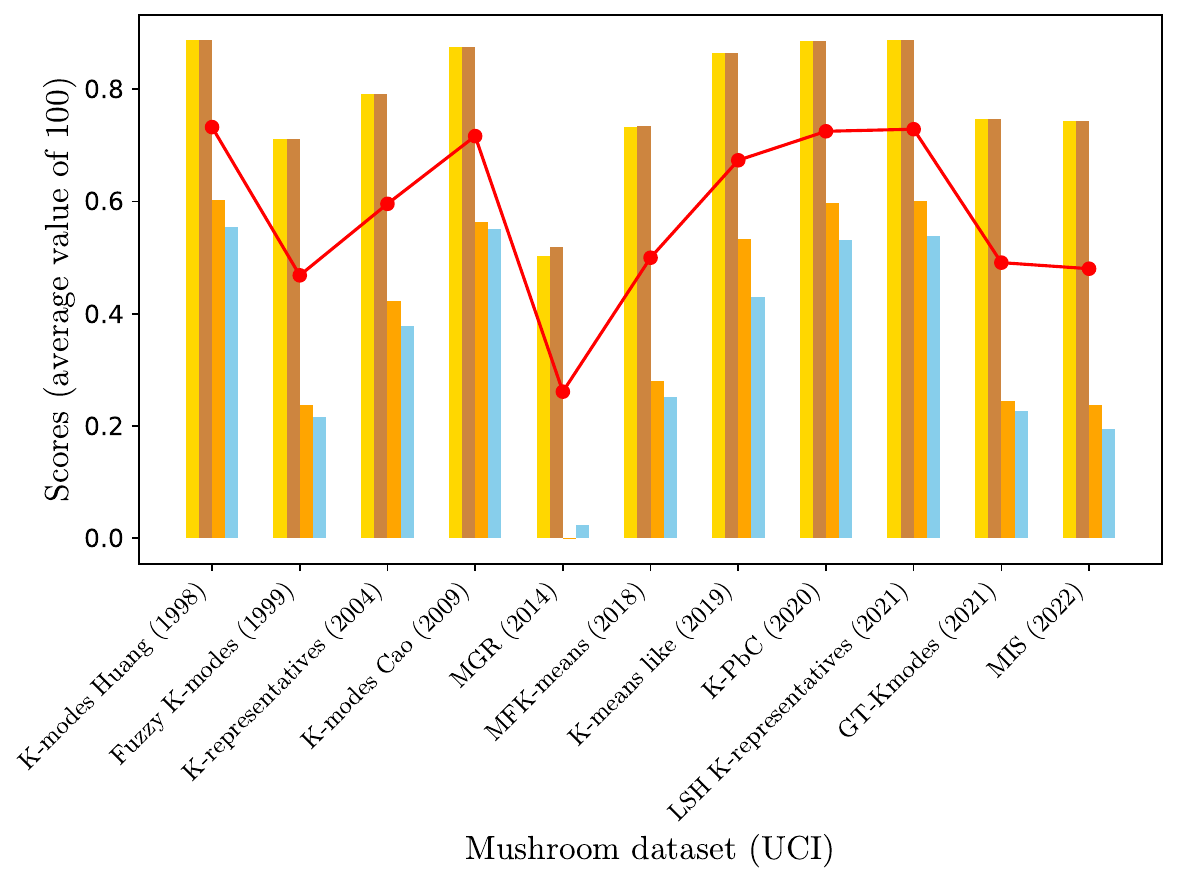}
  \includegraphics[width=0.4\linewidth]{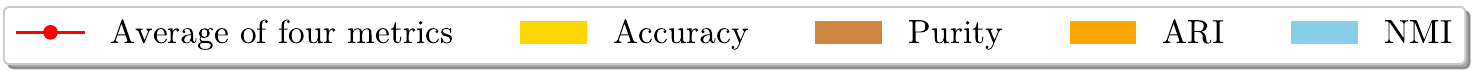}
  \caption{Performance comparison of algorithms on the top four most commonly used datasets}
  \label{fig:github_comparison}
\end{figure*}

\begin{figure*}[!htb]
\vspace{-2.5cm}
  \centering
  \includegraphics[width=0.92\linewidth]{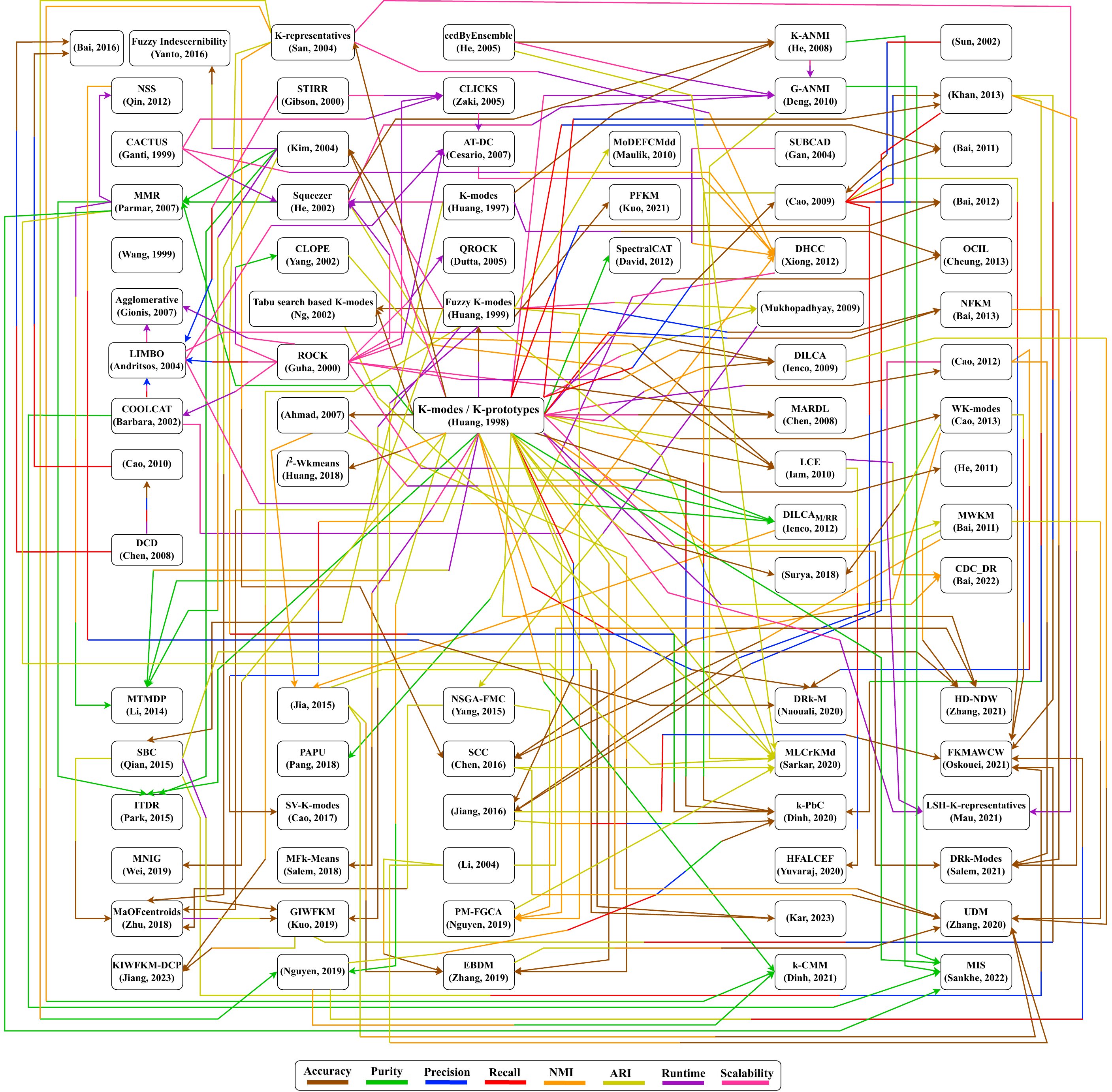}
  \caption{Performance tracking of categorical data clustering algorithms}
  \label{fig:tracking_performance}
\end{figure*}

\subsection{Social sciences}

\cite{okada2015personality} used K-means to convert continuous prosodic features (e.g., pitch, energy) into categorical levels (low, medium, high) as discrete events. A graph clustering algorithm then identified frequently co-occurring patterns in multimodal categorical data, extracting interaction patterns (speech, gestures, gaze) for personality trait classification.
\cite{froemelt2018using} proposed a two-tiered clustering approach to analyze household consumption patterns. They first used a \textsc{Self Organizing Map (SOM)} to reduce the dimensionality, then applied \textsc{hierarchical agglomerative clustering} with \textsc{Ward Linkage} to the SOM, resulting in 34 clusters. The dataset included 157 attributes, covering 85 household consumption expenditures and income categories.

\cite{okada2019modeling} used clustering to analyze multimodal interaction data, focusing on human behavior, personality traits, and social interactions. First, \textsc{K-means} discretized continuous prosodic signals into categorical levels. Then, a custom similarity based algorithm iteratively clustered multimodal events based on temporal co-occurrence.
\cite{sgroi2024analyzing} used \textsc{K-means} clustering to group participants based on their behaviors and preferences for functional foods. This highlights again that categorical variables must be converted to numerical forms through encoding for subsequent analysis with \textsc{K-means}.
\subsection{Economics}
\cite{de2014mining} proposed a two-step clustering method for categorical sequence data. First, a Panel Hidden Markov model transforms sequences into a probabilistic space to capture temporal dependencies. Then, hierarchical clustering groups similar patterns, regardless of sequence length. This method identified clusters in web browsing and employment status sequences, revealing similar behavioral patterns or career trajectories.
\cite{marchal2019detecting} introduced the \textsc{RecAgglo} algorithm for detecting organized eCommerce fraud. It is designed to process large volumes of categorical online order data to group fraudulent orders from the same campaign together while keeping legitimate orders separate.
\cite{kundu2021cloud} incorporated \textsc{Shapley} values from cooperative game theory into categorical clustering algorithms like \textsc{K-modes}, \textsc{Fuzzy K-modes}, and \textsc{Fuzzy K-medoids} to enhance their performance. These algorithms were implemented using Apache Spark for cloud deployment and were applied  to car recommendation.

\cite{huang2024imbalanced} addressed credit card fraud detection by utilizing a hybrid neural network combined with clustering-based undersampling to manage imbalanced data. For categorical variables, the centroid data in each cluster is populated with the most frequently occurring value within that cluster. These clusters and all fraudulent transactions form a balanced training set.
\cite{andonovikj2024survival} used semi-supervised oblique predictive clustering trees (SPYCTs) to analyze a jobseeker dataset with categorical and continuous features. SPYCTs partition data using oblique splits, handling categorical data and non-linear dependencies, and treating censored data as missing values. The resulting clusters represent jobseekers with similar characteristics and predicted employment outcomes.

\section{Comparative experiments} \label{sec:comparative_experiment}
In this section, we have gathered available Python source codes for 14 categorical clustering algorithms from GitHub. We then conducted an experimental comparison of these algorithms using the four most commonly used datasets: \textsc{Mushroom}, \textsc{Soybean}, \textsc{Zoo} and \textsc{Congressional Voting Records} (section \ref{sec:datasets}). This comparison was performed using the four leading external validation metrics: \textsc{Accuracy}, \textsc{Purity}, \textsc{NMI}, and \textsc{ARI} (section \ref{sec:evaluation}). The objective of this experiment is to provide an impartial assessment of the performance of these algorithms. 

The original implementations from GitHub were used with minimal modifications to the source code. Projects with significant errors that could not be resolved were excluded from our analysis. Additionally, any code originally written in Python 2 was updated to Python 3 to ensure compatibility. On our GitHub page\footnote{\href{https://github.com/ClarkDinh/Categorical-data-clustering-25-years-beyond-K-modes}{Categorical data clustering: 25 years beyond K-modes}}, we provide the datasets, source codes for the 14 algorithms, and a Jupyter notebook that contains the clustering results from our experiment.
Figure \ref{fig:github_comparison} illustrates the performance of the 14 algorithms across the four datasets. The height of each bar represents the average performance of each algorithm over 100 runs. For each metric, the tallest bar indicates the best performance. For the \textsc{Mushroom} dataset, three algorithms were excluded from the evaluation due to exceeding the runtime limit. It is important to note that some algorithms may require parameter tuning to achieve optimal results.

Additionally, we provide a performance tracking of the algorithms mentioned in Section \ref{sec:taxonomy}. This tracking allows readers to quickly compare the performance of an algorithm with other algorithms discussed in each paper. In Figure \ref{fig:tracking_performance}, nodes represent papers, and a link connecting two nodes, such as $A \rightarrow B$, indicates that the algorithm proposed in paper B is compared with the algorithm proposed in paper A (with A published before B). The direction of the link signifies that the algorithm in paper B performs better than the one in paper A according to specific evaluation metrics. To capture the variety of metrics used across different comparisons, we represent each metric with distinct colors within the same arrow in the figure. It is worth noting that the experimental results in Figure 24 are sourced directly from the studies cited in each referenced paper. We did not perform independent experiments to assess the performance of these algorithms.
\section{Major challenges} \label{sec:major_challenges}
Despite the promising and, in many contexts, transformative results achieved, several significant challenges remain before fully realizing the potential of categorical data clustering. This section outlines key challenges in categorical data clustering that have been highlighted and discussed in previous studies.
\subsection{Distance metrics}
Distance metrics are essential in data mining and clustering, influencing the structure and quality of clusters. In categorical clustering, many algorithms, including partitional, hierarchical, ensemble, and graph-based clustering, rely on these metrics to group similar objects. The choice of distance metric significantly impacts the final clustering results. The use of an inappropriate metric or transformation method can mislead the clustering process, resulting in poor performance and inaccurate clusters, obscuring true patterns and relationships within the data \citep{reddy2018survey}.

Typically, objects in clustering are represented as vectors in $\mathbb{R}^m$, where $m$ is the number of attributes. For numeric attributes, distance measures like Manhattan, Euclidean, Minkowski, and Mahalanobis are used effectively. However, these measures are often unsuitable for categorical data, where values lack natural ordering and inherent geometric properties, making it difficult to define meaningful (dis)similarity measures \citep{jia2015new}. Therefore, using a measure that captures the intrinsic nature of categorical data is challenging. An effective distance metric for categorical data should account for both the distinctiveness of categories and their frequency of occurrence, ensuring that it meaningfully captures the relationships between different data points. Specialized measures such as Hamming distance, probabilistic measures, information-theoretic measures, and context-based similarity measures have been developed \citep{boriah2008similarity, andreopoulos2018clustering, brzezinska2023similarity}, but finding a universally applicable metric remains an ongoing challenge.
\subsection{High-dimensional data} \label{subsection:high-dimensional}
High-dimensional data often suffer from the \emph{curse of dimensionality}, where the data becomes sparse in the attribute space. This sparsity makes it challenging to identify meaningful clusters across all dimensions \citep{ganti1999cactus}. Instead, clusters often exist in different subspaces of the attribute space. This means that not all attributes are equally relevant for all clusters, and different clusters may be defined by different subsets of attributes. Traditional clustering algorithms, which consider all attributes equally, struggle to identify these subspace clusters effectively. Moreover, the computational cost of clustering algorithms often increases exponentially with the number of dimensions. This is particularly problematic for categorical data, where traditional distance measures may not be applicable \citep{cao2013weighting}.

To address these challenges, researchers have developed subspace clustering techniques tailored for categorical data. Subspace clustering focuses on identifying clusters within different subsets of dimensions, rather than treating all dimensions equally. This approach includes \emph{hard subspace} clustering, which determines specific subsets of dimensions for each cluster \citep{gan2004subspace}, and \emph{soft subspace} clustering, which assigns weights to dimensions based on their relevance in cluster discovery \citep{bai2011novel, he2011attribute, cao2013weighting, zhang2021learnable}. Additionally, algorithms designed to handle high-dimensional categorical data efficiently have been proposed, such as those utilizing parallel subspace methods \citep{pang2018parallel}.

Furthermore, \emph{collinearity} can exacerbate issues in high dimensional space by introducing redundancy and affecting clustering performance. Addressing collinearity involves employing dimensionality reduction techniques and detecting correlation among attributes to enhance clustering effectiveness \citep{carbonera2019subspace}.
\subsection{Large-scale data}
The explosive increase in big data, including categorical data such as market basket and census datasets, has presented significant challenges for clustering algorithms, particularly in terms of computational efficiency when dealing with massive numbers of objects and clusters. Traditional clustering methods often struggle with scalability when applied to very large datasets, as they were initially developed for smaller datasets \citep{huang1997fast}. This scalability issue has become a focal point in data mining research, with efforts directed towards developing algorithms that can efficiently handle the volume and complexity of large-scale categorical datasets while maintaining clustering quality \citep{mau2021lsh}.

Partitional clustering methods generally scale well with data size, making them suitable for large datasets \citep{huang1998extensions, nguyen2023method}. However, various strategies have been proposed to enhance computational efficiency and scalability for large-scale datasets. These include parallel methods \citep{dutta2005qrock, pang2018parallel} and locality-sensitive hashing-based approaches \citep{mau2021lsh}. For high dimensional data, additional approaches such as subspace clustering, discussed in subsection \ref{subsection:high-dimensional}, and feature selection before clustering have also been explored to improve performance and manageability \citep{fomin2023parameterized}.
\subsection{Cluster initialization}
As illustrated in Figure \ref{fig:tree_taxonomy}, partitional clustering has received the most attention from researchers compared to other categorical data clustering techniques. Algorithms like \textsc{K-modes} commonly use \emph{random initialization} to start the clustering process. However, random initialization can lead to non-repeatable and suboptimal results, with different runs potentially yielding varying outcomes \citep{bai2011initialization, khan2013cluster, dinh2020k}.

To address this issue, several initialization methods have been proposed, including frequency-based approaches \citep{huang1998extensions, san2004alternative, khan2013cluster}, density and distance measures \citep{cao2009new, bai2012cluster, sajidha2021initial}, outlier detection techniques \citep{jiang2016initialization}, and pattern mining methods \citep{dinh2020k}. These approaches aim to enhance clustering stability and accuracy by providing more deterministic and representative initial cluster centers. However, each method comes with its own set of limitations, such as computational complexity, sensitivity to outliers, and the need for predefined parameters, searching for an optimal initialization method an ongoing research challenge.

\subsection{Optimal number of clusters}
Determining the optimal number of clusters is a significant challenge in partitional clustering, alongside cluster initialization. Many clustering algorithms require the number of clusters to be specified in advance, which can complicate the prediction of the actual number of clusters and affect the interpretation of the results. An inaccurate estimate, whether over- or under-estimated, can substantially impact the quality of clustering outcomes. Thus, identifying the correct number of clusters in a dataset is a crucial aspect of clustering analysis \citep{dinh2019estimating}.

To address this challenge, researchers have developed techniques to assess both intra-cluster compactness and inter-cluster separation, ensuring that clusters are internally cohesive and externally distinct \cite{guo2016cluster}. Another common method involves using cluster validation techniques in a trial-and-error approach. This entails generating clusterings with varying numbers of clusters and evaluating them using internal validation metrics, such as the silhouette coefficient \citep{dinh2019estimating}, to determine the most appropriate number of clusters. In addition, hierarchical clustering offers an alternative approach as it does not require pre-specifying the number of clusters. The challenge is further intensified by the need for efficient algorithms able to manage large datasets, identify true cluster structures, and deliver robust clustering results.

\subsection{Stream and dynamic data}
Data streams produce large volumes of data at high speeds, making it infeasible to store and process the entire dataset. Consequently, algorithms need to be able to process data in a single pass and keep up with the incoming data rate. Users may want to analyze clusters over different time horizons or with different parameters. Algorithms need to provide this flexibility without reprocessing the entire data stream. A significant challenge in clustering categorical data streams is the detection and adaptation to concept drift. \textsc{Concept Drift} refers to the changes in the underlying data distribution over time, which can result in the formation of new clusters and the dissolution of old ones. Effective clustering algorithms must continuously update their models to reflect these changes, ensuring that the clustering remains relevant and accurate \citep{aggarwal2010clustering}.

Several methods have been proposed to address the challenges of clustering categorical data streams. One approach uses condensation to summarize the data stream into a set of fine-grained cluster droplets. These droplets serve as a condensed representation of the data, capturing essential information while reducing the data volume \citep{aggarwal2010clustering}. Another method employs a sliding window technique, detecting changes or drifts in the underlying data concept within the current window compared to previous clustering results \citep{chen2008catching, bai2016optimization, cao2010framework}.

The dynamic nature of data streams poses significant challenges for clustering, requiring algorithms to process data in real-time or near-real-time while maintaining computational efficiency, accuracy, and relevance of the clustering results.

\subsection{Interpretability}
The last challenge we discuss in this section is the interpretability of categorical data clustering algorithm output. In many fields, such as health sciences, there is a need for more granular insights into the decision-making processes of clustering algorithms. Advanced clustering approaches often produce results that are too complex to be easily applied in clinical practice. This complexity versus interpretability trade-off means that more sophisticated models may not be practical for end-users without specialized training \citep{budiarto2023machine}.

Model explainability is crucial for the general public, as it helps instill trust and cooperation. This is particularly important in contexts such as transfer learning or when there is a risk of bias or unfairness due to data or modeling decisions \citep{van2023current}. Transparency and ability to correct complex clustering models will also be needed to refine output. Improved interpretability can be achieved through end-to-end pipelines and enhanced model feedback, including detailed log messages and graphical output representations.

Another critical issue is interpreting certainty in label assignment for individual cases. While fuzzy clustering addresses this by allowing degrees of membership to multiple clusters, a gap remains in handling uncertain cases, especially in hard clustering. Deciding what constitutes a \emph{certain enough} assignment is crucial, as misclassifications can have significant consequences, particularly in sensitive applications like healthcare. Developing methods to quantify and communicate the certainty of label assignments can help users make more informed decisions and trust the clustering results.

Furthermore, the lack of training in computational methods among non-specialists underscores the need for user-friendly frameworks and guidelines. These tools would facilitate the systematic reporting of critical modeling decisions, such as parameter tuning and selecting the optimal number of clusters, thereby improving the overall usability of clustering algorithms.
\section{Trends and opportunities} \label{sec:trends_opportunities}
\subsection{Hybrid models}
In recent years, there has been a growing trend towards hybrid models that combine clustering and regression techniques to enhance the analysis of categorical data.
\cite{carrizosa2021clustering} introduced a method that integrates the clustering of categorical predictors directly into generalized linear models. This approach simplifies the model by reducing the number of parameters while maintaining or improving predictive accuracy, offering a more interpretable solution for high-cardinality categorical data.
\cite{richman2024high} used clustering as a preprocessing step to structure high-cardinality hierarchical categorical covariates, which are then integrated into advanced regression models, such as neural networks, through techniques like entity embedding.
\cite{campo2024clustering} proposed a hybrid model that combines clustering and regression to analyze complex categorical survey data. This approach employs categorical clustering to consolidate multiple item responses into a single meta-item through joint data reduction. This meta-item is then analyzed using regression on ordered polytomous variables, with recursive partitioning for variable selection.
\cite{campo2024clustering} proposed the \textsc{PHiRAT} algorithm, which clusters categories within a hierarchical risk structure using features derived from regression models like random effects models for damage rates and claim frequencies. 

These models leverage the strengths of clustering to group similar categorical levels, thereby reducing the complexity and dimensionality of the data before applying regression models. This approach not only improves the interpretability and efficiency of the regression models but also helps to mitigate issues like overfitting, which are common with high-dimensional categorical data.
\subsection{Graph data mining}
Graph data mining is becoming a key trend in categorical data clustering due to the need to manage complex relationships in categorical datasets \citep{bai2022categorical, bandyapadhyay2023parameterized, fomin2023parameterized}, as discussed in section \ref{sec:graph_clustering}.

\cite{amburg2020clustering} proposed a framework for clustering nodes in hypergraphs with categorical edge labels. The objective function groups nodes based on common interaction types. For two categories, they use a minimum s-t cut problem, while for more than two categories, they apply approximation algorithms involving linear programming and multiway cut problems. This method extends to hypergraphs, enabling analysis of higher-order interactions in categorical data. 
\cite{soemitro2024spectral} proposed a spectral clustering for categorical and mixed-type data by constructing a graph where nodes represent data points and edges represent numerical similarities. To incorporate categorical variables, additional nodes are added for each category and connected to the corresponding data points with fixed-weight edges. This approach embeds categorical information directly into the graph, avoiding preprocessing steps like discretization, and enables more effective clustering of data into homogeneous groups. 

As graph data mining techniques continue to evolve, they present opportunities for more sophisticated and scalable clustering algorithms that can better handle diverse and large-scale categorical datasets.

\subsection{Parallel clustering}
High-dimensional and large-scale data pose significant challenges in categorical data clustering, necessitating the development of frameworks and algorithms capable of addressing these issues \cite{pang2018parallel}.

\cite{pang2019puma} proposed the \textsc{PUMA} algorithm, which utilizes \textsc{Hadoop} to distribute computational tasks across multiple nodes, efficiently managing large-scale, high-dimensional categorical datasets. In its first stage, \textsc{PUMA} performs local clustering on individual nodes to identify sub-clusters. These sub-clusters are then merged in the second stage to form global clusters. This parallel processing strategy effectively reduces processing time and enhances scalability.
\cite{li2022mics} proposed the \textsc{MiCS-P} algorithm, which harnesses the \textsc{Spark} platform to parallelize mutual information computations, thereby improving the efficiency of clustering large-scale categorical datasets. It employs a column-wise transformation approach to decompose the dataset into feature subsets, allowing for parallel mutual information calculations between feature pairs.

By utilizing frameworks such as \textsc{Hadoop} and \textsc{Spark}, researchers can optimize clustering algorithms to handle larger and more complex datasets. Emerging parallel clustering techniques have the potential to significantly improve scalability, reduce processing times, and facilitate real-time data analysis.

\subsection{Large language models}
An emerging trend in categorical data clustering is the application of large language models (LLMs). This innovative approach involves a multi-step process: first, embedding LLMs, such as OpenAI's text-embedding models, are used to transform text-based categorical variables into high-dimensional numeric vectors, capturing rich semantic information. Then \textsc{K-means} or \textsc{hierarchical clustering} is applied to these embeddings rather than the original categorical data. Finally, text-generating LLMs, such as GPT-4-Turbo, are employed to interpret and label the resulting clusters, providing human-readable insights. This method offers a sophisticated way to handle text-based categorical variables, potentially extracting more nuanced information than traditional encoding techniques. This trend represents a promising intersection of natural language processing and clustering methodologies, opening new avenues for analyzing complex categorical data in various domains, including healthcare and beyond \citep{sharabiani2024predictive}.
\section{Conclusion} \label{sec:conclusion}
This paper provides a comprehensive review of the advancements in categorical data clustering over the past 25 years, starting from the introduction of the \textsc{K-modes} algorithm. The paper delves into the historical development of clustering algorithms, categorizing them into hierarchical, partitional, ensemble, subspace, graph-based, and genetic-based clustering methods. Each category is thoroughly examined, with detailed comparisons provided in dedicated tables, capturing the essential aspects of each algorithm, including their computational complexity, evaluation metrics, and visualization techniques.

A significant contribution of this paper is the taxonomy of clustering algorithms based on their underlying methodologies, datasets used, and validation metrics. This taxonomy offers a snapshot overview of the theoretical frameworks and experimental information related to each algorithm, enabling readers to grasp the main ideas and understand the evolution of clustering techniques in the research community.

Furthermore, the paper explores the applications of categorical data clustering across different domains, including health sciences, natural sciences, education, engineering, social sciences, and economics. This exploration demonstrates how these techniques are adapted and utilized in various contexts, highlighting the broad impact and relevance of the research. Practical comparisons of algorithms accessible on GitHub are conducted, revealing the performance of recent algorithms on several benchmark categorical datasets. This practical aspect underscores the importance of reproducibility and the need for efficient clustering techniques in real-world applications.

The paper also addresses the major challenges and opportunities in categorical data clustering. Key challenges include the development of effective distance metrics, handling high-dimensional and large-scale data, cluster initialization, and the interpretability of clustering results. Lastly, the paper discusses emerging trends and potential avenues for future research, such as hybrid models, graph data mining, parallel clustering, and the application of large language models.

In conclusion, this review provides a valuable resource for researchers and practitioners working with categorical data clustering. As the volume and complexity of categorical data continue to grow, the need for efficient and effective clustering techniques becomes increasingly critical. This underscores the importance of continued research and development in this vital area of data science.
\section*{Declaration of competing interest}
The authors declare that they have no known competing financial interests or personal relationships that could have appeared to influence the work reported in this paper.

\section*{CRediT authorship contribution statement}
Tai Dinh: conceptualization, data curation, methodology, experiment, visualization, writing, editing; Wong Hauchi: data curation, writing, visualization; Philippe Fournier-Viger: writing, editing; Daniil Lisik: writing, editing; Minh-Quyet Ha: writing; Hieu-Chi Dam: editing, validation; Van-Nam Huynh: editing, validation.

\section*{Acknowledgments}
The authors sincerely thank the senior editor and the two referees for their valuable remarks and insightful comments on the manuscript.

\bibliographystyle{model5-names}
\biboptions{authoryear}
\bibliography{main}
\end{document}